\def\year{2021}\relax
\documentclass[letterpaper]{article} 
\usepackage{aaai21}  
\usepackage{times}  
\usepackage{helvet} 
\usepackage{courier}  
\usepackage[hyphens]{url}  
\usepackage{graphicx} 
\usepackage{natbib}  
\usepackage{caption} 
\usepackage{comment}
\usepackage{amsmath,amssymb} 
\usepackage{multirow}
\usepackage{multicol}
\usepackage{subcaption}
\usepackage{pifont}
\usepackage{bbm}
\usepackage{bm}
\usepackage{array}
\usepackage{color}
\usepackage[switch]{lineno}
\usepackage{xspace}

\newcommand{\PreserveBackslash}[1]{\let\temp=\\#1\let\\=\temp}
\newcolumntype{C}[1]{>{\PreserveBackslash\centering}p{#1}}
\newcolumntype{R}[1]{>{\PreserveBackslash\raggedleft}p{#1}}
\newcolumntype{L}[1]{>{\PreserveBackslash\raggedright}p{#1}}

\makeatletter
\DeclareRobustCommand\onedot{\futurelet\@let@token\@onedot}
\def\@onedot{\ifx\@let@token.\else.\null\fi\xspace}

\def\eg{\emph{e.g}\onedot} 
\def\ie{\emph{i.e}\onedot}

\def\etal{\emph{et al}\onedot}
\makeatother

\title{High-Resolution Deep Image Matting}
\author{
    Haichao Yu$^1$, Ning Xu$^2$, Zilong Huang$^1$, Yuqian Zhou$^1$, Humphrey Shi$^{1,3}$\\
}
\affiliations{
\textbf{$^1$UIUC, $^2$Adobe Research, $^3$University of Oregon}\\
\{haichao3, yuqian2, hshi10\}@illinois.edu, nxu@adobe.com, zilong.huang2020@gmail.com
}

\begin{document}
\maketitle

\begin{abstract}
Image matting is a key technique for image and video editing and composition.
Conventionally, deep learning approaches take the whole input image and an associated trimap to infer the alpha matte using convolutional neural networks. 
Such approaches set state-of-the-arts in image matting; however, they may fail in real-world matting applications due to hardware limitations, since real-world input images for matting are mostly of very high resolution.
In this paper, we propose HDMatt, a first deep learning based image matting approach for high-resolution inputs.
More concretely, HDMatt runs matting in a patch-based crop-and-stitch manner for high-resolution inputs with a novel module design to address the contextual dependency and consistency issues between different patches.
Compared with vanilla patch-based inference which computes each patch independently, we explicitly model the cross-patch contextual dependency with a newly-proposed Cross-Patch Contextual module (CPC) guided by the given trimap.
Extensive experiments demonstrate the effectiveness of the proposed method and its necessity for high-resolution inputs. Our HDMatt approach also sets new state-of-the-art performance on Adobe Image Matting and AlphaMatting benchmarks and produce impressive visual results on more real-world high-resolution images.
\end{abstract}
\section{Introduction}
Image matting is a key technique in image and video editing and composition. Given an input image and a trimap indicating the background, forground and unknown regions, image matting is applied to estimate the alpha matte inside the unknown region to clearly separate the foreground from the background. Recently, many deep-learning-based methods~\cite{xu2017deep,lu2019indices,hou2019context,cai2019disentangled} have achieved significant improvements over traditional methods~\cite{wang2007optimized,gastal2010shared,sun2004poisson,levin2007closed,grady2005random}. These deep learning methods~\cite{xu2017deep,lu2019indices,hou2019context} mostly take \textit{the whole images} and \textit{the associated whole trimaps} as the inputs, and employ deep neural networks such as VGG~\cite{simonyan2014very} and Xception~\cite{chollet2017xception} as their network backbones.

\begin{figure}[t]
    \centering
    \begin{subfigure}{.32\linewidth}
      \centering
      \includegraphics[width=.99\linewidth]{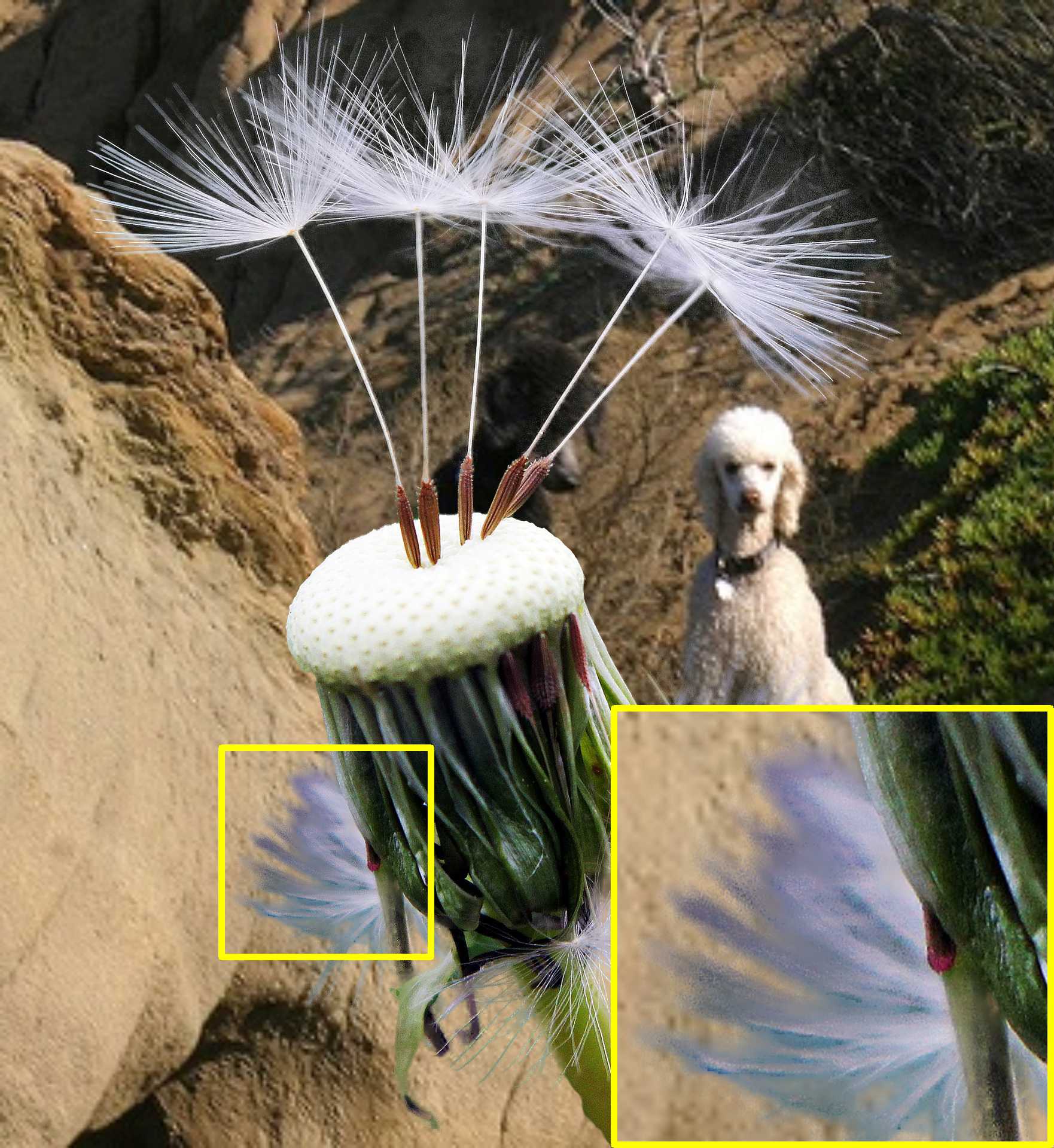}
      \subcaption{HR Image}
    \end{subfigure}
    \begin{subfigure}{.32\linewidth}
      \centering
      \includegraphics[width=.99\linewidth]{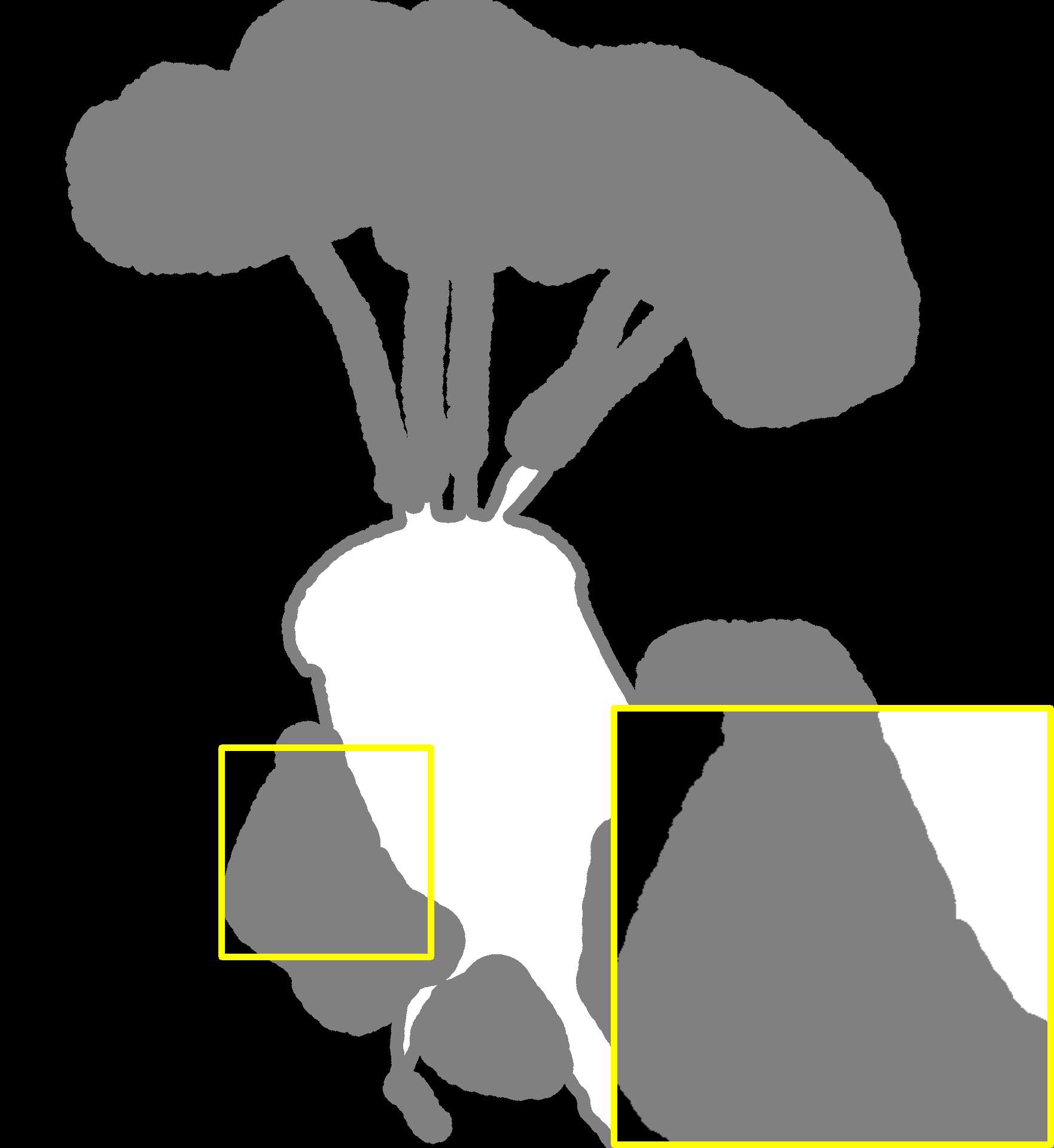}
      \subcaption{Trimap}
    \end{subfigure}
    \begin{subfigure}{.32\linewidth}
     \centering
     \includegraphics[width=.99\linewidth]{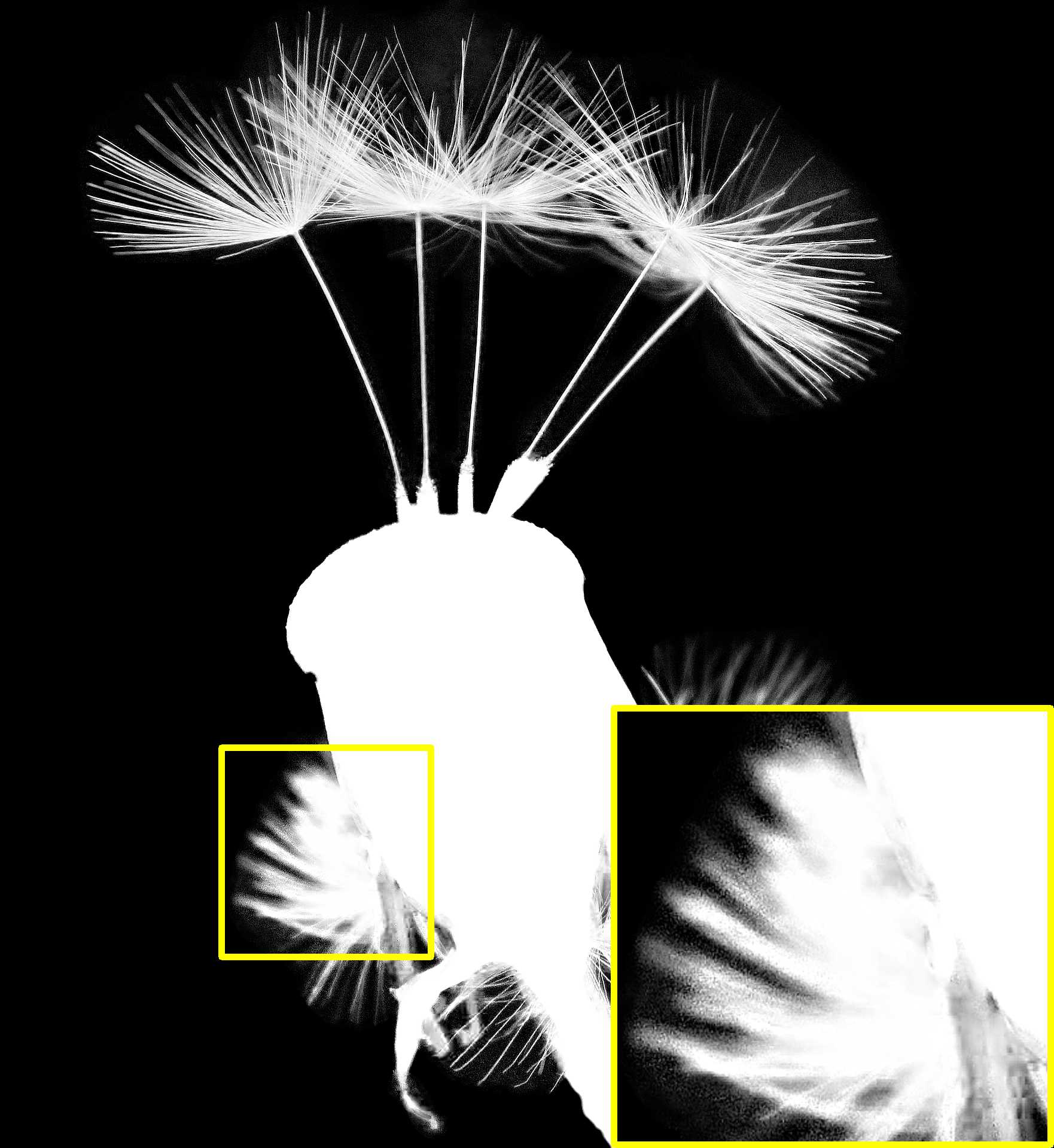}
     \subcaption{Ground Truth}
    \end{subfigure}
    \begin{subfigure}{.32\linewidth}
      \centering
      \includegraphics[width=.99\linewidth]{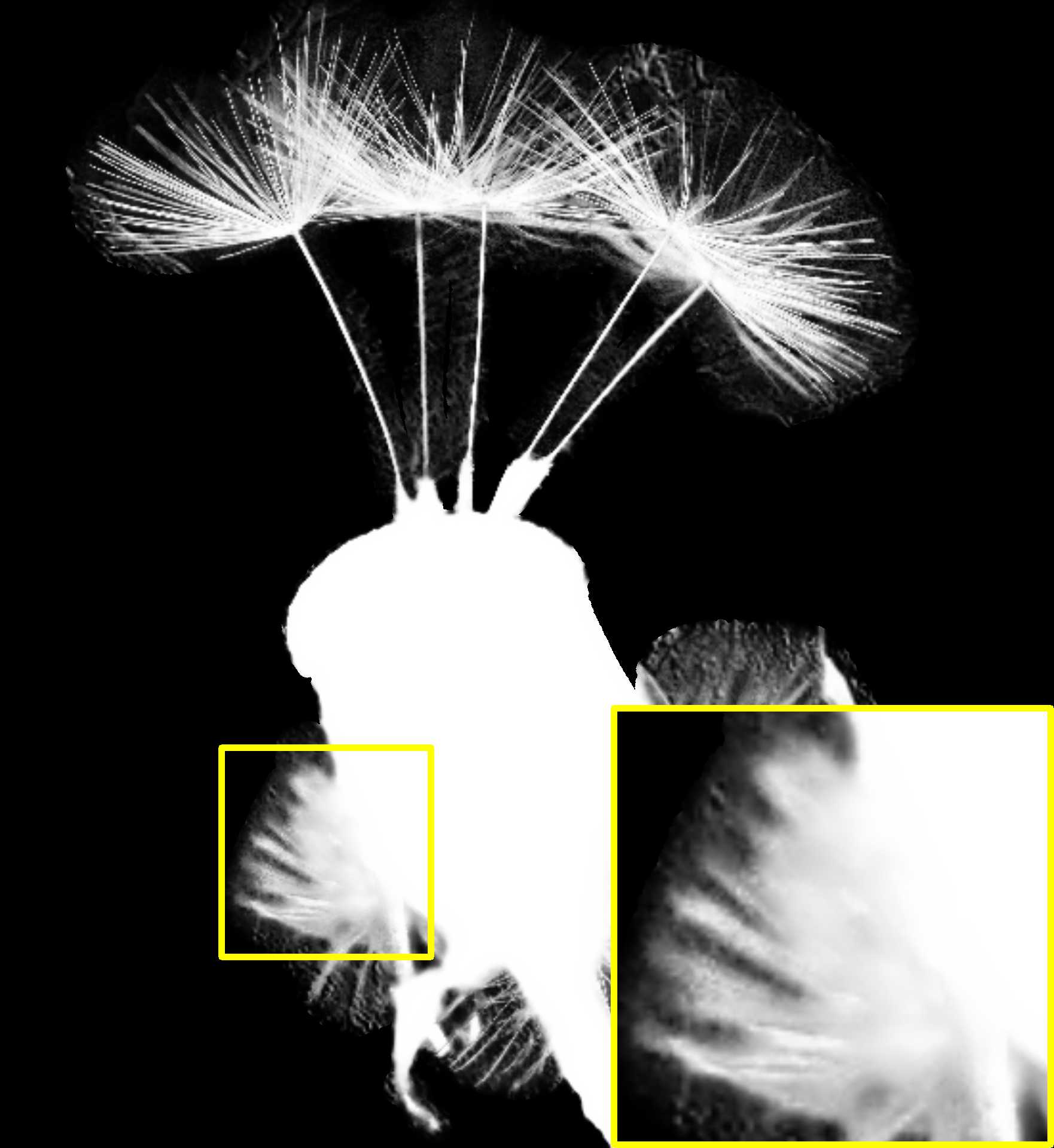}
      \subcaption{ContextNet-DS}
      \label{fig.down_fail}
    \end{subfigure}
    \begin{subfigure}{.32\linewidth}
      \centering
      \includegraphics[width=.99\linewidth]{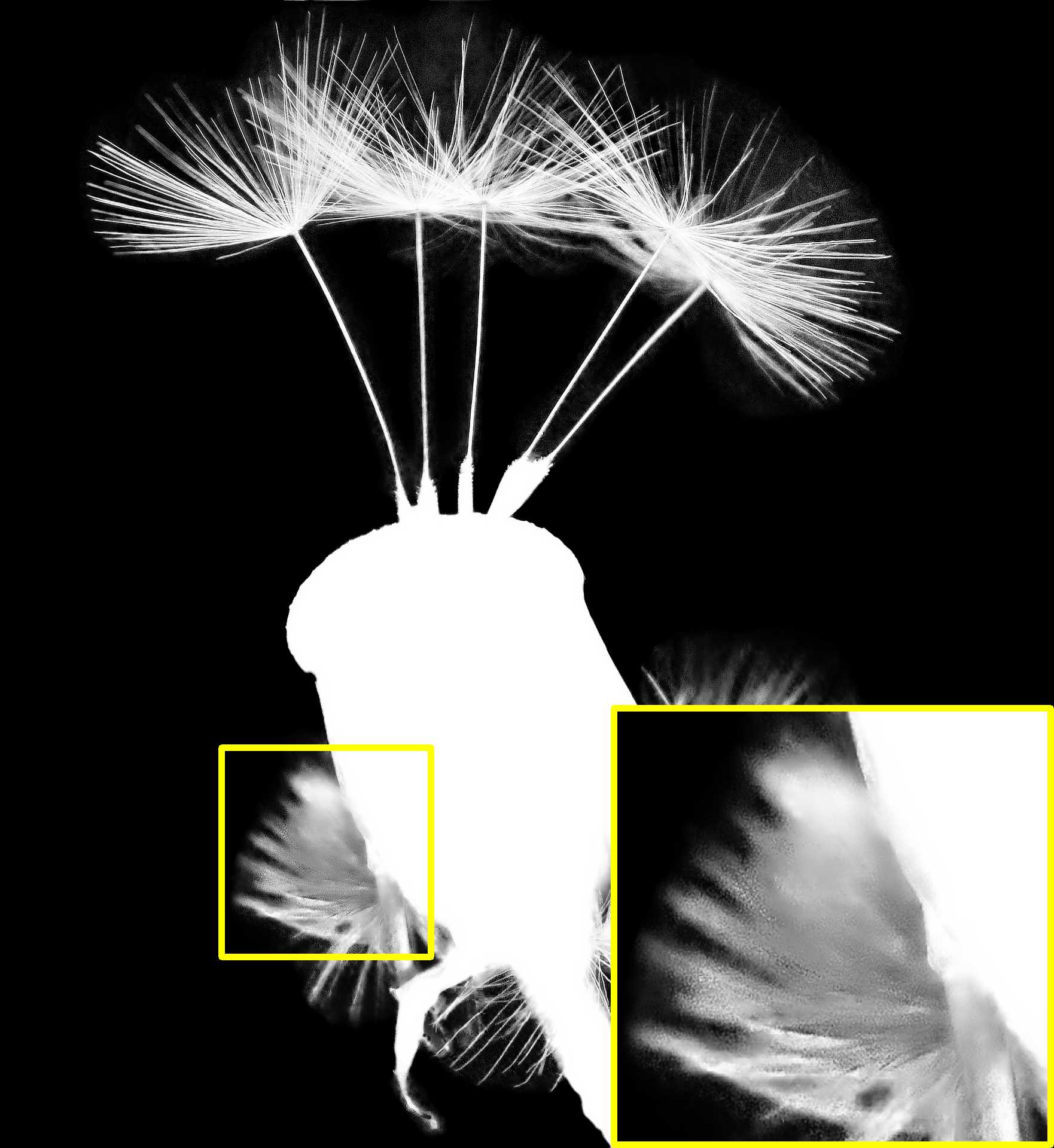}
      \subcaption{ContextNet-C}
      \label{fig.crop_fail}
    \end{subfigure}
    \begin{subfigure}{.32\linewidth}
      \centering
      \includegraphics[width=.99\linewidth]{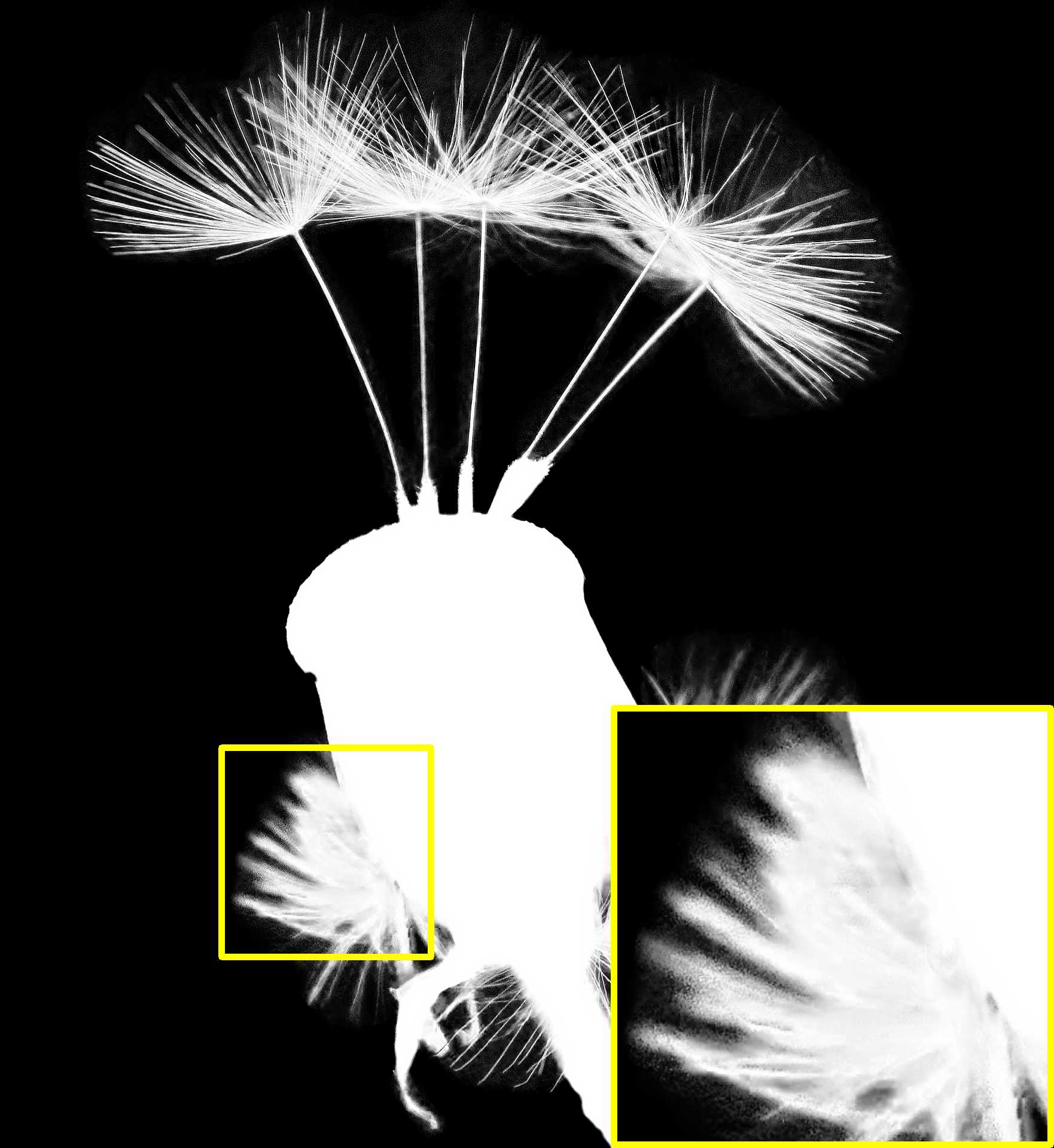}
      \subcaption{HDMatt (Ours)}
      \label{fig.patch_good}
    \end{subfigure}
    \caption{Down-sampling (DS) and cropping (C) strategies applied to ContextNet~\cite{hou2019context} on an HR image. DS results in blurry details, and trivial cropping causes cross-patch inconsistency. Our HDMatt resolves the above drawbacks. Best viewed when zoomed in with colors.}
    \label{fig:teaser}
    \vspace{-1.5em}
\end{figure}

However, these methods may fail when dealing with high-resolution (HR) inputs. Image matting is frequently applied to HR images of size such as $5000\times5000$ or even higher in real-world applications. Due to hardware limitations like GPU memory, HR images cannot be directly handled by previous deep learning methods. Two common strategies of adapting those methods are down-sampling the inputs~\cite{he2010guided} or trivial patch-based inference. The former strategy results in losing most fine details, and the latter causes patch-wise inconsistency. Besides, HR images may have larger or even fully unknown regions within a patch. This further requires the models to understand contextual information from long-range patches for successful matting. A comparison run with ContextNet~\cite{hou2019context} and our proposed method is shown in Fig. \ref{fig:teaser} to demonstrate these drawbacks.

In this paper, we propose HDMatt, a novel patch-based deep learning approach for high-resolution image matting. Specifically, we crop an input image into patches, and propose a Cross-Patch Contextual module (CPC) to explicitly capture cross-patch long-range contextual dependency. For each given patch to be estimated (\ie query patch), CPC samples other patches (\ie reference patches) which are highly correlated with it within the image. Then CPC ensembles those correlated features towards a more faithful estimation.

To measure the correlation and ensemble the information effectively, inspired by traditional propagation-based methods~\cite{levin2007closed,sun2004poisson,levin2007closed}, Trimap-Guided Non-Local operation (TGNL) is specifically designed for matting and embedded into the CPC. In particular, compared with the original non-local operation~\cite{wang2018non} applied to the whole patch, we leverage the pixel labels in the trimap to guide the correlation computing. Pixels in unknown regions in the query patch will be compared with three regions (\ie foreground, background and unknown) in the reference patches separately, allowing an efficient information propagation across different pixel types.

The above mentioned designs are intended for cross-patch long-range dependency modeling. As a patch-based method, it is intrinsically indispensable for HR image matting. In summary, the contributions of this paper are three-folds:

\begin{itemize}
    \item To our best knowledge, we are the first to propose a deep learning based approach to HR image matting, and makes high-quality HR matting practical in the real-world under hardware resources constraints.
    \item We propose a novel Cross-Patch Contextual module (CPC) to capture long-range contextual dependency between patches in our HDMatt approach. Inside the CPC, a newly-proposed Trimap-Guided Non-Local (TGNL) operation is designed to effectively propagate information from different regions in the reference patches.
    \item Both quantitatively and qualitatively, our method achieves new state-of-the-art performance in image matting on the Adobe Image Matting (AIM)~\cite{xu2017deep}, the AlphaMatting~\cite{rhemann2009perceptually} benchmark, and our newly collected real-world HR image dataset.
\end{itemize}

\section{Related Work}
\label{sec:related}
\subsection{Image Matting}
Before deep learning methods, there are two types of classic methods for matting task. One is sampling-based methods. Given an unknown pixel, these methods sample matched pixels from foreground and background regions and then find a proper combination of these pixels to predict alpha value of the unknown pixel. These methods include boundary sampling~\cite{wang2007optimized}, ray casting sampling~\cite{gastal2010shared}, etc.
Another interesting sampling-based method is Divide-and-Conquer~\cite{cao2016divide}. In this paper, the authors proposed an adaptive patched-based method for HR image matting. To capture global information, they sample as context the pixels that are close to current pixel in RGBXY feature space in other patches. Although our method shares a similar sampling spirit with their method, ours is intrinsically different from theirs in many aspects. First, to our best knowledge, we are the first to use deep learning models to capture long-range contexts among patches for image matting. Second, we sample context patches in high-level feature space instead of pixels in RGBXY space. Thus, our method can better capture long-range context in semantic level.

Another type is propagation-based methods. These methods include Poisson equation based method~\cite{sun2004poisson}, random walks for interactive matting~\cite{grady2005random} and closed-form matting\cite{levin2007closed}, which, based on local smoothness, formulates a cost function and then find the globally optimized alpha matte by solving linear equation system. Another popular propagation-based method is non-local image matting~\cite{lee2011nonlocal,chen2013knn}. For an unknown pixel to predict alpha value, this method sample pixels that match with current pixel in some feature space and make prediction with the the sampled pixels as context. Our method shares some spirit with this method in that our method make prediction by sampling context patches to capture long-range context.

Deep learning-based methods is another branch which has been widely explored. Cho~\etal in \cite{cho2016natural} proposed a novel deep learning method to combine alpha mattes from KNN matting~\cite{chen2013knn} and Closed-form matting. However, these methods are still restricted to specific type of images due to limited training set. The first large-scale image matting dataset is collected by Xu~\etal~\cite{xu2017deep}. Building on this, they proposed a novel Deep Image Matting (DIM) model with refinement module. They achieved state-of-the-art performance on their collected test dataset. Since the availability of the large-scale dataset, deep learning methods for matting have been extensively explored. Lutz~\etal proposed a generative adversarial network AlphaGan for image matting~\cite{lutz2018alphagan}. Hou~\etal~\cite{hou2019context} proposed ContextNet, which used dual-encoder structure to capture contextual and detail information and dual-decoder structure for foreground and alpha prediction. Among these methods, Unpooling is usually preferred to other upsampling methods like transposed convolution and bilinear upsampling. This is studied by Lu~\etal~\cite{lu2019indices}. They further proposed IndexNet to dynamically determine the indices for unpooling operation. Recently, Li~\etal~\cite{li2020natural} proposed GCAMatting, which utilizes pixel-wise contextual attention to capture long-range contexts. Though having impressive performance, these models will potentially fail on ultra-high-resolution image inference due to hardware limitation, thus not practical enough. Our proposed patch-based method works well on ultra-high-resolution images, and additional modeling of cross-patch dependency address the issues caused by crop-and-stitch manner.

\subsection{Non-local Operations}
Non-local operations are widely used for various tasks such as video classification~\cite{wang2018non}, object detection~\cite{wang2018non,huang2018ccnet}, semantic segmentation~\cite{fu2019dual,huang2018ccnet} and machine translation~\cite{vaswani2017attention}. Wang~\etal~\cite{wang2018non} proposed a group of non-local operations to capture long-range context. Their method achieved impressive results on video classification task. Based on that, to reduce memory consumption inside the non-local operations, Huang~\etal~\cite{huang2018ccnet} used stacked criss-corss attention to mimic the non-local operations. DANet~\cite{fu2019dual} used channel-wise and spatial attention to capture long-range dependency along both channel and spatial dimensions. In this paper, we are aware that long-range context dependency is potentially necessary for high-resolution images, especially those with large unknown areas. Therefore, we further develop the non-local module from \cite{wang2018non} to make it adaptive to cross-patch modeling (\ie, CPC) and trimap guidance (\ie, TGNL). Our new state-of-the-art experimental results indicate the promising directions of adapting non-local operations to image matting. 

\begin{figure}[t]
    \centering
    \includegraphics[width=0.9\linewidth]{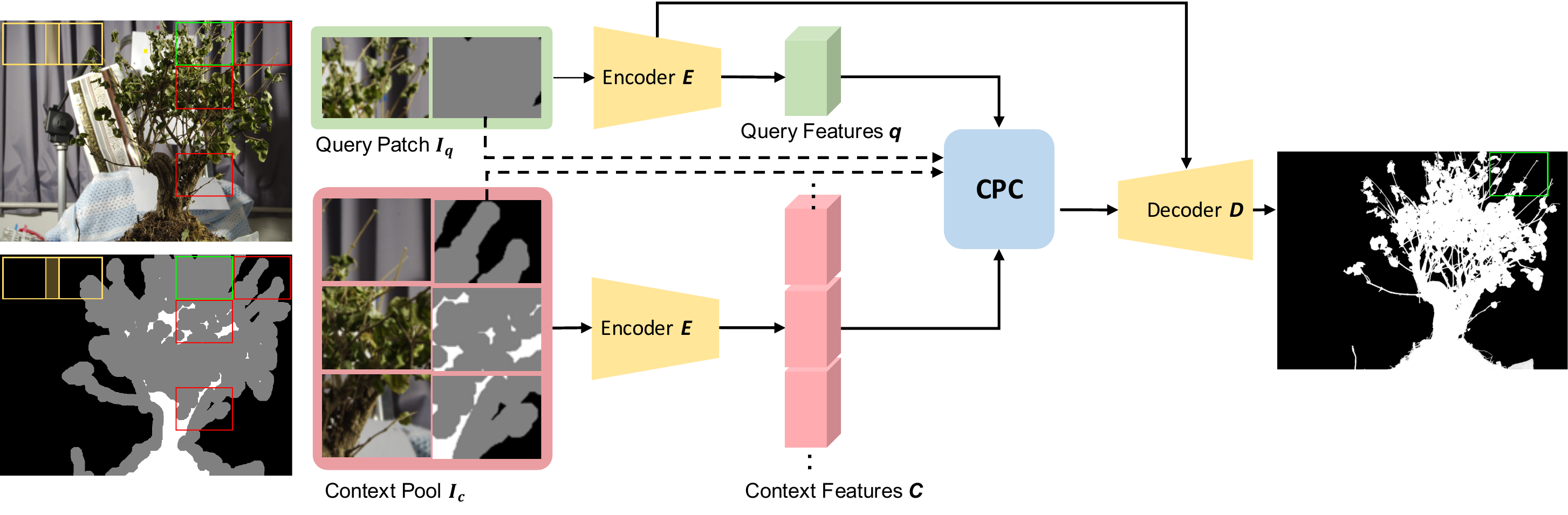}
    \caption{An overview of our proposed HDMatt approach. It works on patches and is basically an encode-decoder structure. Query patch concatenated with its associated trimap is fed into the encoder. The patches in the context pool and their trimaps are also fed into the encoder shared weights with $E$. The extracted features go through the Cross-Patch Context (CPC) module. Afterwards, the output feature of CPC is fed into the decoder for alpha estimation of the query patch. The green and red boxes are query and context patches during training. The yellow boxes are two consecutive query patches during test.}
    \label{fig:archi}
    \vspace{-1em}
\end{figure}

\section{The Proposed HDMatt Approach}\label{sec:method}
To handle high-resolution image matting, our method first crops an input image and trimap into patches (Sec.~\ref{sec:patch}) and then estimates the alpha values of each patch. Only using information from a single patch will cause information loss and prediction inconsistency between different patches. Therefore we propose a novel Cross-Patch Context Module (CPC) (Sec.~\ref{sec:cpc}) to leverage cross-patch information for each query (current) patch effectively. Finally, the estimated alpha value of each patch is stitched together to output the final alpha matte of the whole image. The network structure and loss function are described in Sec.~\ref{sec:network}. Fig.~\ref{fig:archi} illustrates the framework of our method.

\subsection{Patch Cropping and Stitching}
\label{sec:patch}
Given a training image and trimap, our method randomly samples image patches and their corresponding trimaps of different sizes (\eg~$320\times320,480\times480,640\times640$) at different locations and then they are resized to a fixed size $320\times320$. During inference,
the whole test image $I$ and trimap $T$ are first cropped into overlapping patches (See the two yellow patches in Fig.~\ref{fig:archi} as an example). For those patches exceeding the image boundary, we utilize reflective padding to fill up the pixels. The small overlapping region is helpful to avoid boundary artifacts when stitching the alpha mattes of nearby patches together. In particular, we design some blending function to merge the estimated alphas of overlapping regions between  nearby patches for a smooth transition. See Appendix~\ref{app:stitch} for the detailed blending function.

\begin{figure}[t]
    \centering
    \includegraphics[width=0.9\linewidth]{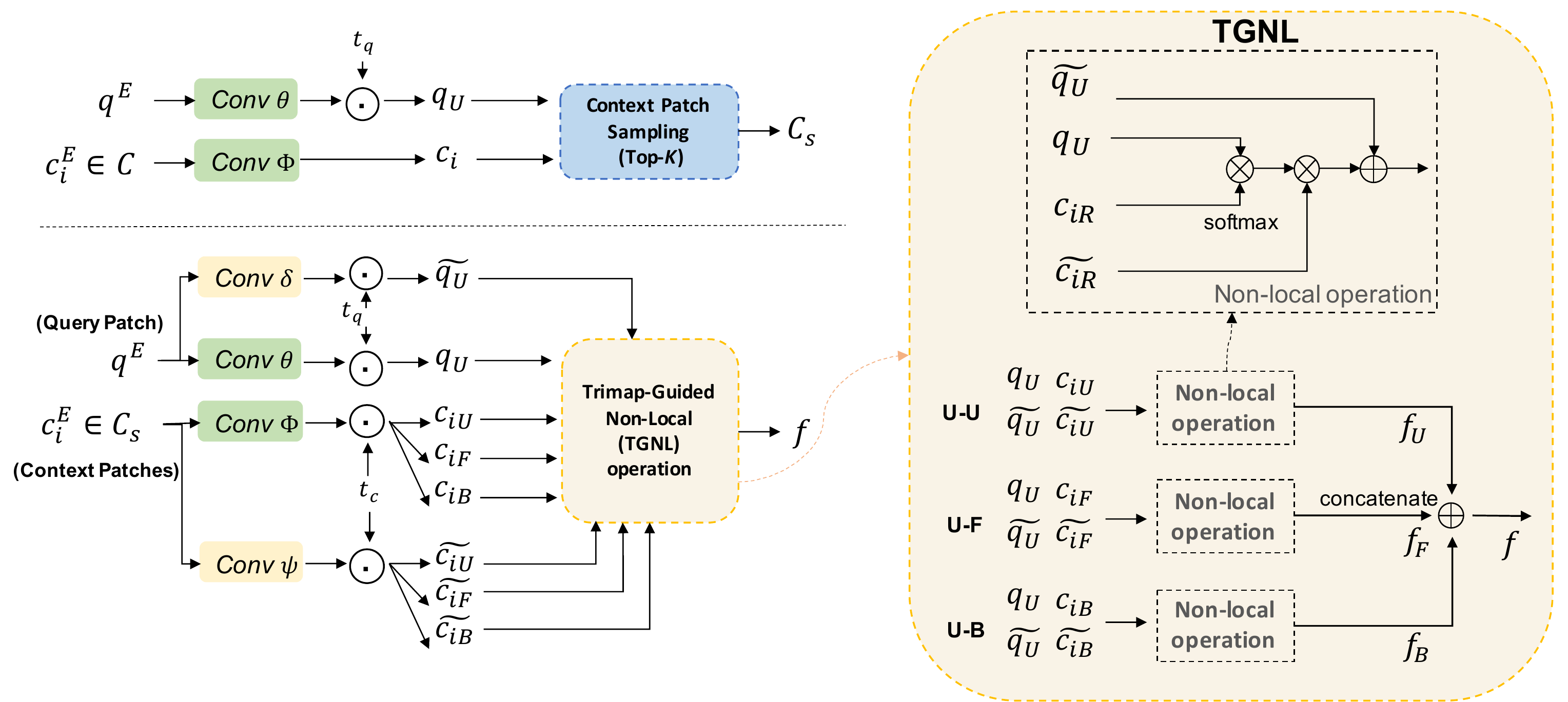}
    \caption{The workflow of the Cross-Patch Context (CPC) module. It consists of a context patch sampling, and a  Trimap-guided Non-Local (TGNL) operation. $\otimes$: matrix multiplication, $\oplus$: feature map concatenation, $\odot$: element-wise multiplication.}
    \label{fig:cpc}
    \vspace{-1em}
\end{figure}

\subsection{Cross-Patch Context Module}\label{sec:cpc}
Our method  leverages cross-patch information for high-resolution image matting. For each query patch, Instead of using all the information from the other patches, we propose an effective sampling strategy to only sample top-$K$ patches which are most relevant and useful to the query patch, and thus save computation greatly without decreasing the accuracy. In addition, in contrast to most prior works that only concatenate the trimap with image as a network input, our method proposes a more effective and explicit way to leverage trimap as a guidance to propagate information from different regions.
\subsubsection{Context Patch Sampling}
Given a query patch $I_q$, to select top-$K$ patches from $N$ context patches $I_{c_i}, c_i \leq N$, our method first computes the correlation between the \textit{unknown regions} of $I_q$ and the \textit{whole regions} of each $I_{c_i}$.
Specifically, as shown in Fig.~\ref{fig:archi}, both $I_q$ and $I_{c_i}$ along with their trimaps are fed into an encoder to extract higher-level feature maps (For simplicity, let $\bm{q}^E$ and $\bm{c}_i^E$ denote their corresponding feature maps). Then  $\bm{q}^E$ and $\bm{c}_i^E$ are further embedded by two convolutional layers $\theta$ and $\phi$ into $\bm{q}$ and $\bm{c}_i$, as shown in Fig.~\ref{fig:cpc}. To get the unknown regions (U) of the new feature $\bm{q}$, we use the downsampled trimap to zero out the foreground (F) and background (B) regions of $\bm{q}$, \ie, $\bm{q}_U = \bm{q} \odot \mathbbm{1}_{s \in U}$, where $s$ is the pixel index. Then the correlation between the two feature maps $\bm{q}_U$ and $\bm{c}_i$ can be computed by  summing over the dot product of their features at each location, \ie
\begin{equation}
    \label{eq:h}
   h(\bm{q}_U, \bm{c}_i) = \sum_{s, s^\prime}{\bm{q}_{U,s}\cdot \bm{c}_{i,s^\prime}},
\end{equation}
where $s$ and $s^\prime$ are the pixel positions in $\bm{q}_U$ and $\bm{c}_i$ respectively. The correlations between query patch and all $N$ context patches are normalized via the softmax operation, which results in a similarity score for each context patch, \ie
\begin{equation}
    \label{eq:d}
    d_{c_i} =
    \frac{e^{h(\bm{q}_U, \bm{c}_i})}{\sum_{\hat{i}}e^{{h(\bm{q}_U, \bm{c}_{\hat{i}})}}},
\end{equation}
A higher score indicates that the candidate context patch is more relevant to the unknown regions of the query patch, and thus should play a more important role in information propagation. During inference, we rank all the context patches according to their similarity scores $d_{c_i}$ and only select the top-$K$ context patches for feature propagation. Empirically we find that $K=3$ can already achieve comparable accuracy compared to utilizing all N context patches.

\subsubsection{Trimap-Guided Non-Local (TGNL) Operation}
To propagates the useful information of context patches to the query patch, we leverage non-local operation~\cite{wang2018non,oh2019video} which were proposed for different tasks. In addition, for the matting problem, trimap provides very useful information about the foreground, background and unknown regions. A unknown pixel which is similar to a foreground pixel is more likely to be foreground pixel than background pixel, and vice versa. Therefore, it is important to propagate the context information from different regions indicated by the trimaps. While recent deep-learning-based matting methods usually concatenate the trimaps as input, which makes it difficult for their methods to leverage such information precisely.

To remedy this issue, our method incorporates the trimap information into the non-local operation. Specifically, our method compares the unknown  region (U) of the query patch with the unknown (U), foreground (F) and background (B) regions of the context patches separately. Then the correlation features from the three different relationships (\ie~U-U, U-F, and U-B) are concatenated together and used as the decoder input. 

As shown in Fig.~\ref{fig:cpc}, the query feature $\bm{q}^E$ from the encoder output is further embedded by two convolutional layers $\theta$ and $\delta$ into a key feature map $\bm{q}$ and a value feature map $\bm{\widetilde{q}}$. Similarly, every sampled context patch feature $\bm{c}_i^E$ is embedded by two convolutional layers $\theta$ and $\phi$ into a key feature map $\bm{c}_i$ and a value feature map $\widetilde{\bm{c}_i}$. We then use the downsampled query trimap to extract the feature maps of the unknown region, \ie~$\bm{q}_U = \bm{q} \odot \mathbbm{1}_{s \in U}$, and $\widetilde{\bm{q}_U} = \widetilde{\bm{q}} \odot \mathbbm{1}_{s \in U}$.
Similarly, we use the context trimap to extract the feature maps of the three regions separately, \ie~$\bm{c}_{i,R} = \bm{c}_i \odot \mathbbm{1}_{s \in R}$ and $\widetilde{\bm{c}_{i,R}} = \widetilde{\bm{c}_i} \odot \mathbbm{1}_{s \in R}$, where $R \in \{\text{U, F, B}\}$. Then the propagated features by comparing the U region of the query patch with the R region of all sampled context patches can be computed as follows,
\begin{equation}
    \label{eq:c}
    \bm{f}_{R,s} = \widetilde{\bm{q}_{U,s}} + \sum_{i, s^\prime}
    \frac{e^{(\bm{q}_{U,s}\cdot {\bm{c}_{i,R,s^\prime}})}}{\sum_{\hat{i}, \hat{s^\prime}}e^{(\bm{q}_{U,s}\cdot {\bm{c}_{\hat{i},R,\hat{s^\prime}}})}} \widetilde{\bm{c}_{i,R,s^\prime}},
\end{equation}
where $s$ is a pixel location of the aggregated feature map $\bm{f}_{R,s}$. 
Finally, the aggregated feature maps of all three regions $\bm{f}_U$, $\bm{f}_F$ and $\bm{f}_B$ are concatenated together as the module output and are used for the decoder. It is possible that some context trimaps may not contain all the three regions, but our Eqn.~\ref{eq:c} still works, and empirically we find that our model have robust results for such cases.

\subsection{Network Structure and Losses}
\label{sec:network}
Fig.~\ref{fig:archi} illustrates the overall framework of our method.
The encoder $E$ consists of a backbone feature extractor ResNet-34~\cite{he2016deep} and Atrous Spatial Pyramid Pooling (ASPP)~\cite{chen2017rethinking}. Pooling outputs from encoder blocks are skip-connected to the corresponding decoder layers. Following \cite{xu2017deep}, we use unpooling operation for feature map upsampling in decoder, which is verified to be more effective~\cite{lu2019indices} for matting-related tasks. 

For fair comparison, we use the same loss function as in \cite{xu2017deep} to train the whole network end-to-end. It is an average of alpha loss $\mathcal{L}_{\alpha}$ and composite loss $\mathcal{L}_{c}$. Formally, for each pixel, the losses are defined as
\begin{equation}
\begin{split}
    &\mathcal{L}_{overall} = 0.5 \mathcal{L}_{\alpha} + 0.5\mathcal{L}_{c}, \\
    &\mathcal{L}_{\alpha}= \sqrt{||\alpha_{gt} - \alpha_{q}||^{2} + \epsilon}, \\
    &\mathcal{L}_{c} = \sqrt{||I_{q} - (\alpha_{q} I_{q}^F + (1 - \alpha_{q}) I_{q}^B )||^{2} + \epsilon},
\end{split}
\end{equation}
where $\alpha_{gt}$ is the ground truth alpha matte, $I_q^F$ and $I_q^B$ are the ground truth background and foreground images to composite $I_q$, and $\epsilon$ is the slacking factor to be set as $10^{-12}$. As mentioned earlier, smooth blending is employed on the overlapping region between neighboring patches during test, and thus  pixels along the boundary regions of each training patch should be weighted accordingly. Therefore, we employ the same blending function as a weighted mask to the loss $\mathcal{L}_{overall}$.
\section{Experiments}
\label{sec:exp}

\begin{table}[t]
    \begin{center}
    \caption{The quantitative results on AIM testset~\cite{xu2017deep}. Methods in the section ``Whole" take as input the whole image, which are also the proposed testing strategy for these methods. We also test several methods on overlapping patches with the same patch size as our method.}
    \begin{tabular}{c|l|cccc}
        \hline
        \multicolumn{2}{c|}{Methods} & SAD & MSE  & Grad & Conn \\
        \hline
        \multirow{6}{*}{\rotatebox{90}{\footnotesize Whole}}
        & AlphaGAN           & 52.4 & 30 & 38 & 53 \\
        & DIM                & 50.4 & 14 & 31.0 & 50.8 \\
        & IndexNet           & 45.8 & 13 & 25.9 & 43.7 \\
        & AdaMatting         & 41.7 & 10 & 16.8 & - \\
        & ContextNet         & 35.8 & 8.2 & 17.3 & 33.2 \\
        & GCAMatting         & 35.3 & 9.1 & 16.9 & 32.5 \\
        \hline
        \multirow{4}{*}{\rotatebox{90}{{\footnotesize Patch}}} &  IndexNet & 54.5 & 16.8 & 31.8 & 54.0 \\
        & ContextNet & 37.6 & 8.7 & 18.7 & 34.8 \\
        & GCAMatting & 36.9 & 8.9 & 17.1 & 34.1 \\
        & HDMatt (Ours) & \bf{33.5} & \bf{7.3} & \bf{14.5} & \bf{29.9} \\
        \hline
    \end{tabular}\label{tab:aim}
    \end{center}
    \vspace{-1em}
\end{table}

\begin{figure*}[th]
\centering
\begin{subfigure}{.16\linewidth}
  \centering
  \includegraphics[width=.99\linewidth]{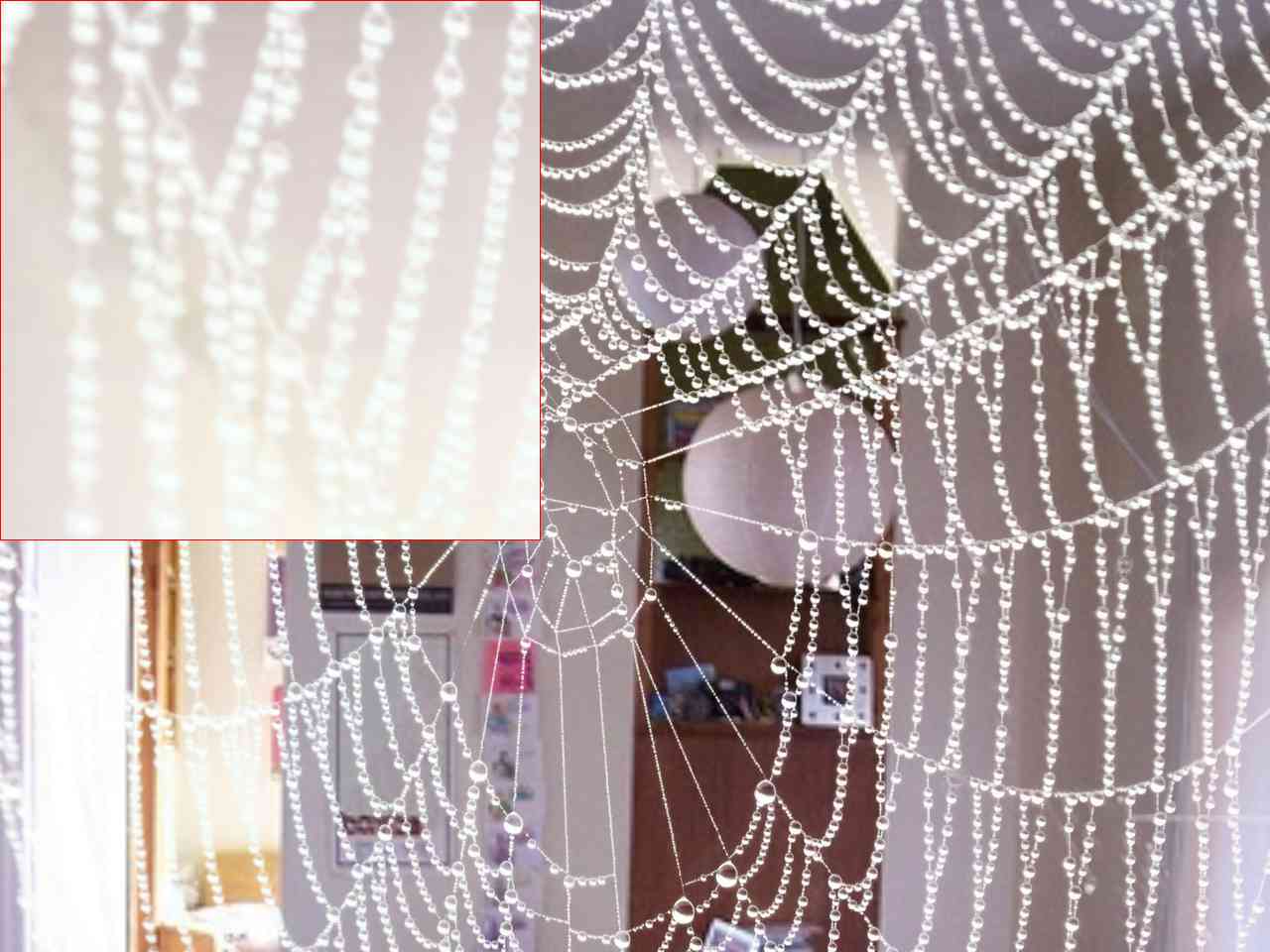}
\end{subfigure}
\begin{subfigure}{.16\linewidth}
  \centering
  \includegraphics[width=.99\linewidth]{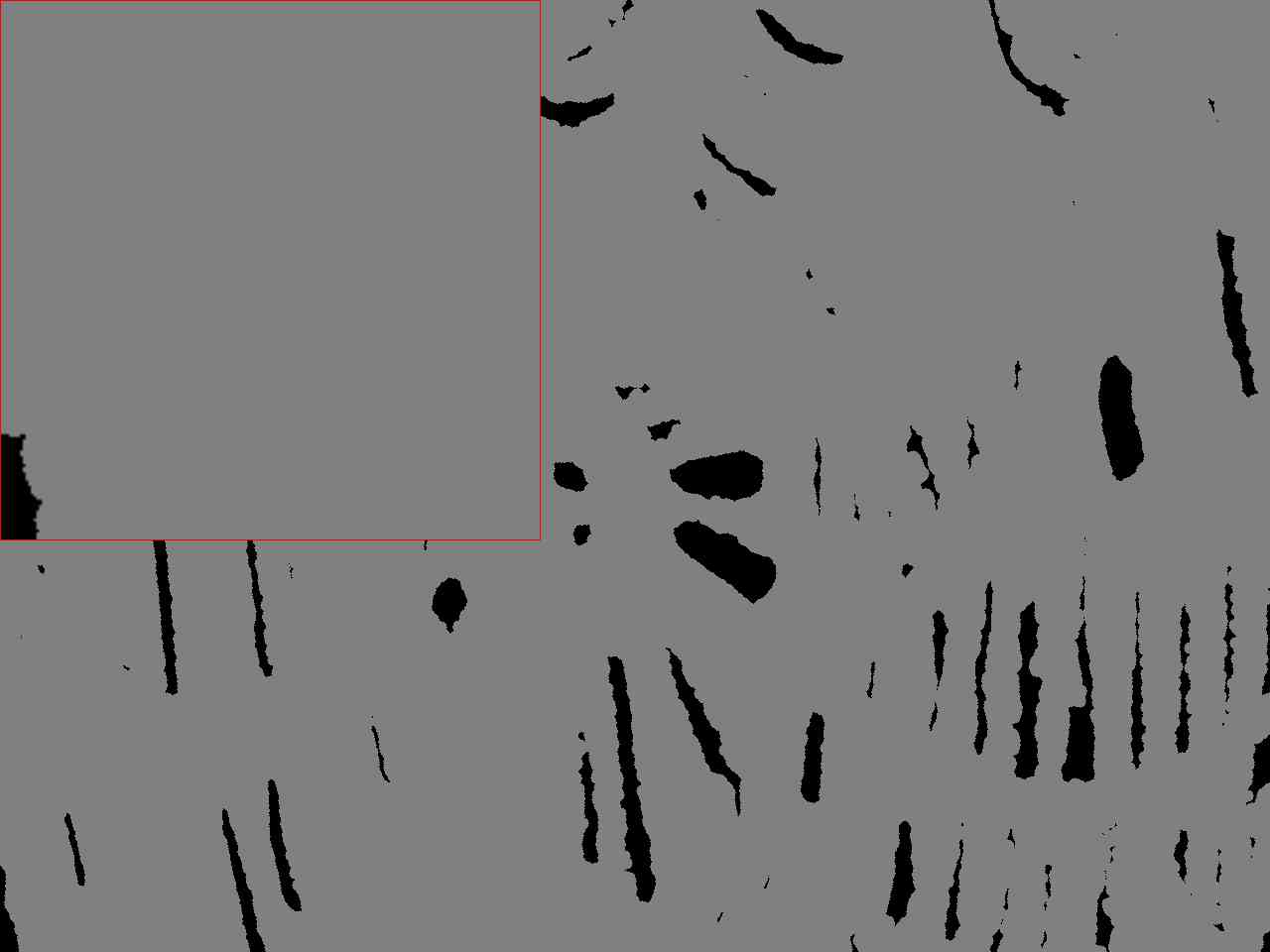}
\end{subfigure}
\begin{subfigure}{.16\linewidth}
  \centering
  \includegraphics[width=.99\linewidth]{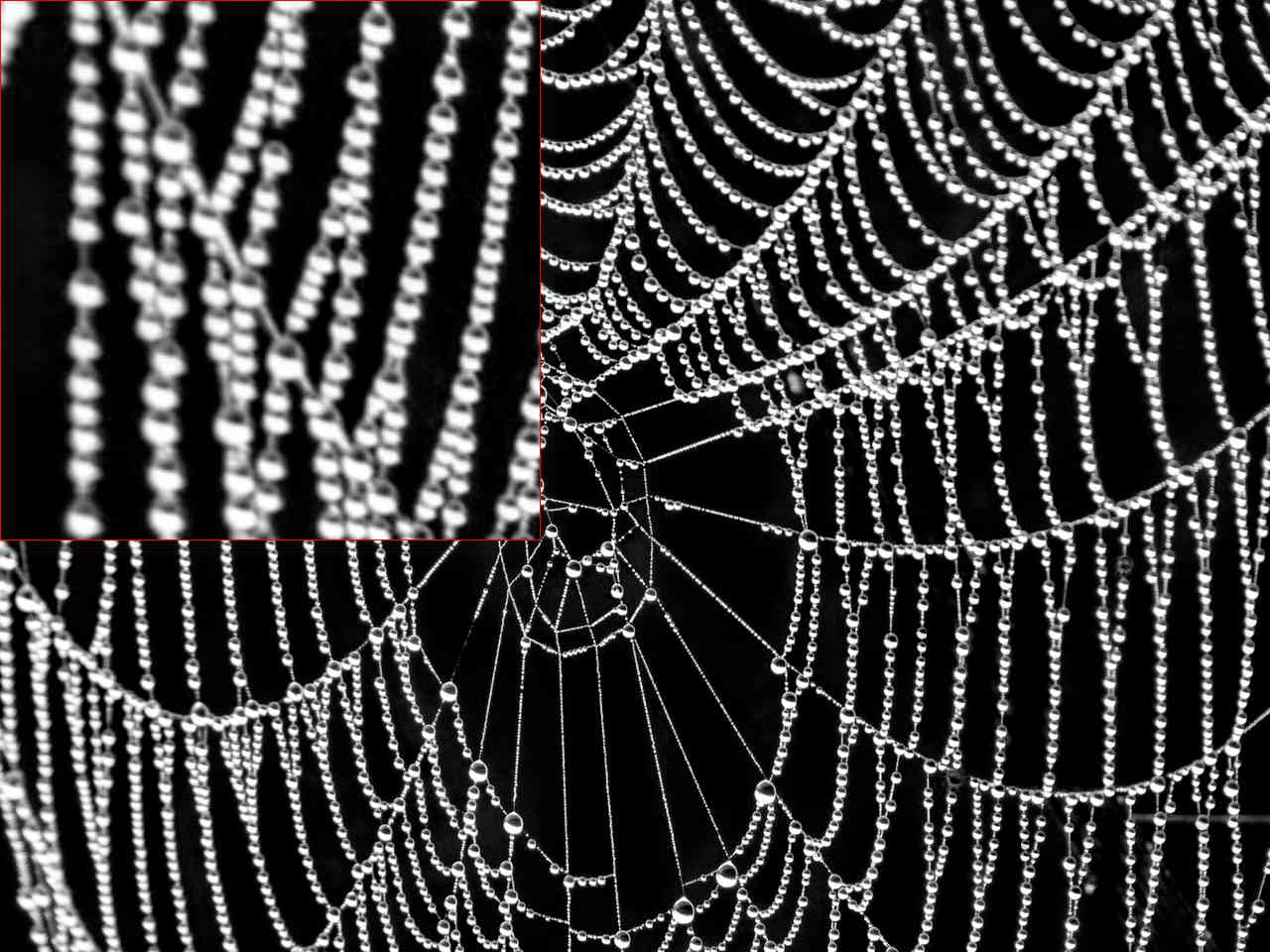}
\end{subfigure}
\begin{subfigure}{.16\linewidth}
  \centering
  \includegraphics[width=.99\linewidth]{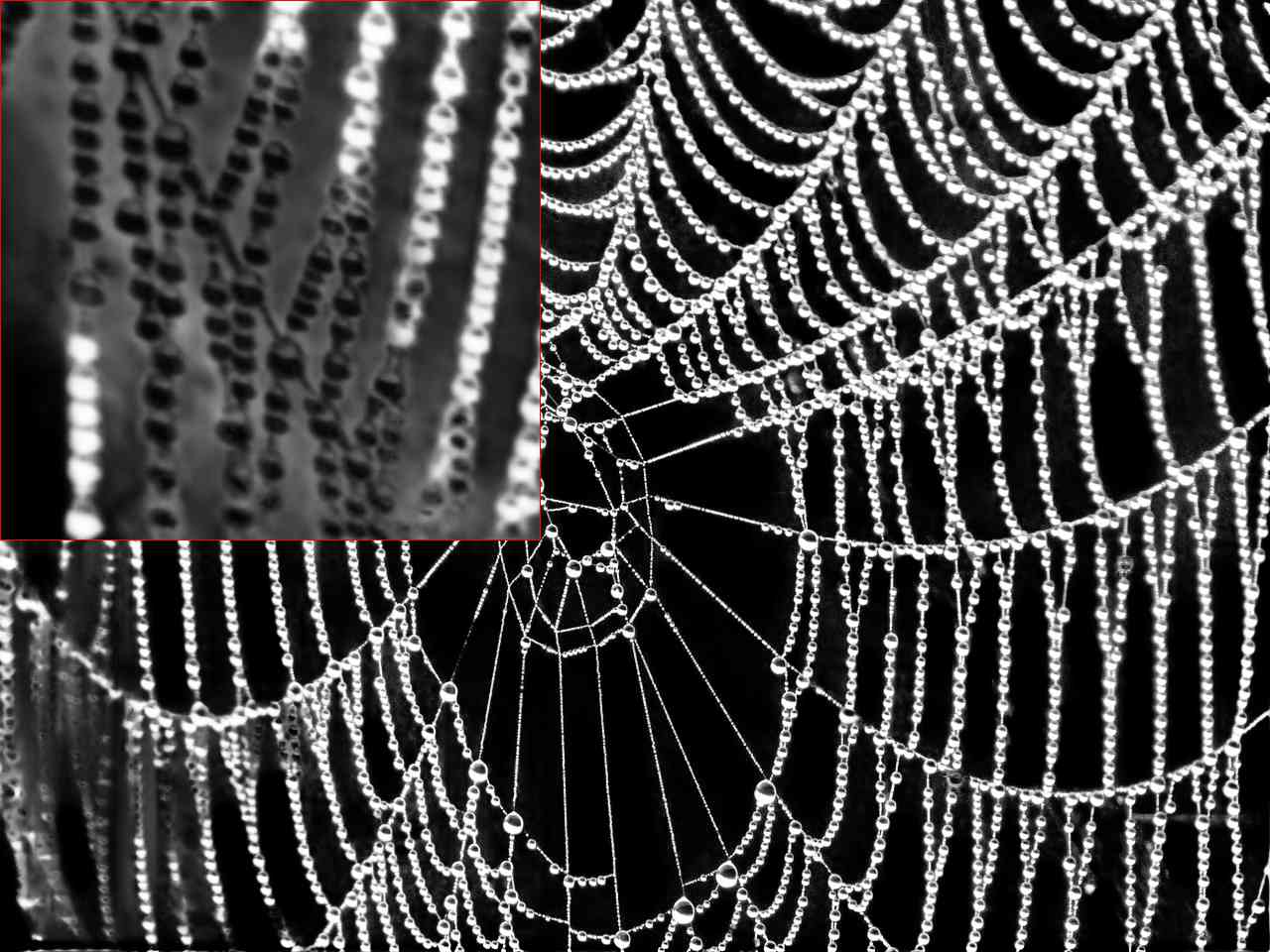}
\end{subfigure}
\begin{subfigure}{.16\linewidth}
  \centering
  \includegraphics[width=.99\linewidth]{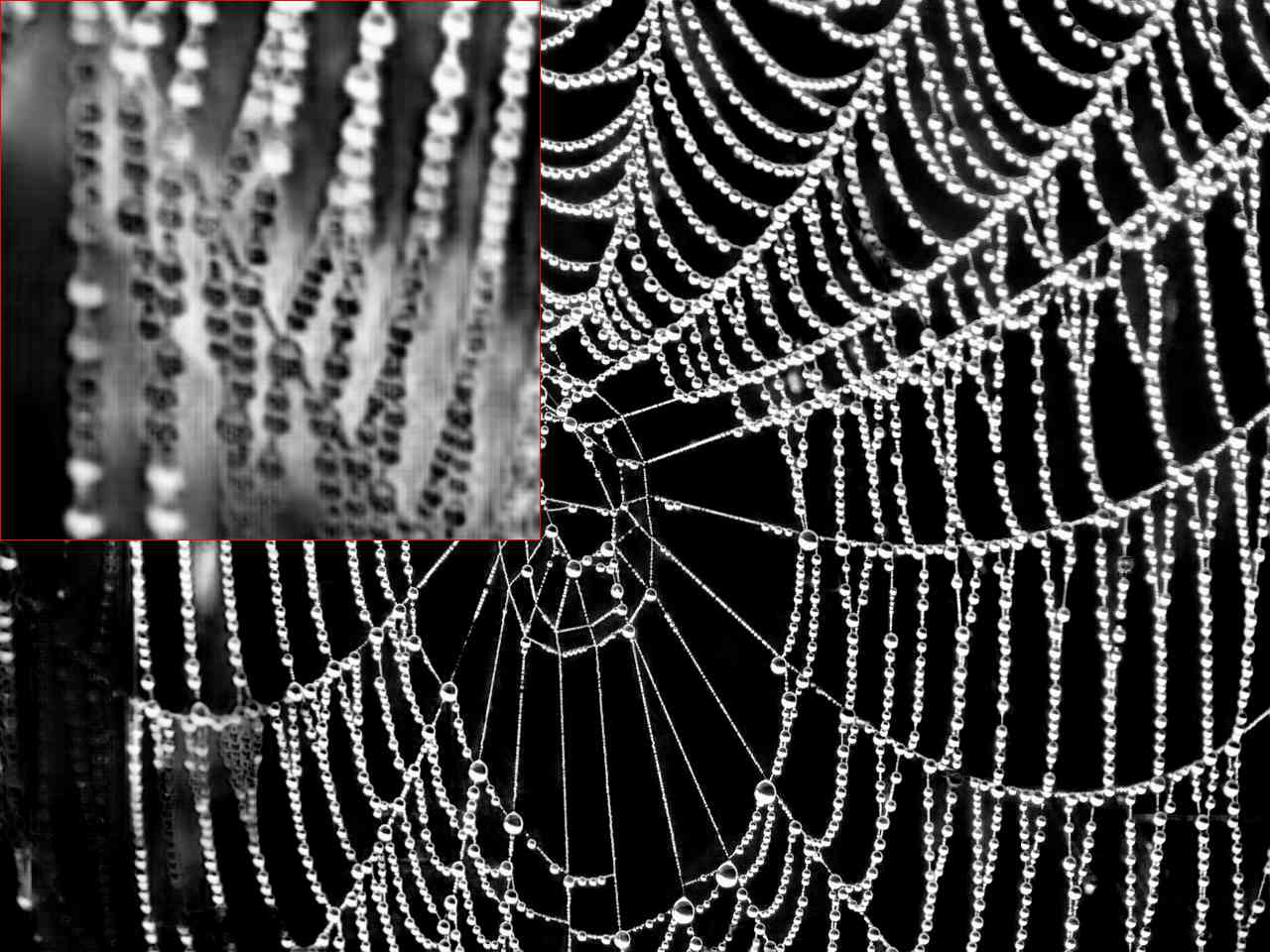}
\end{subfigure}
\begin{subfigure}{.16\linewidth}
  \centering
  \includegraphics[width=.99\linewidth]{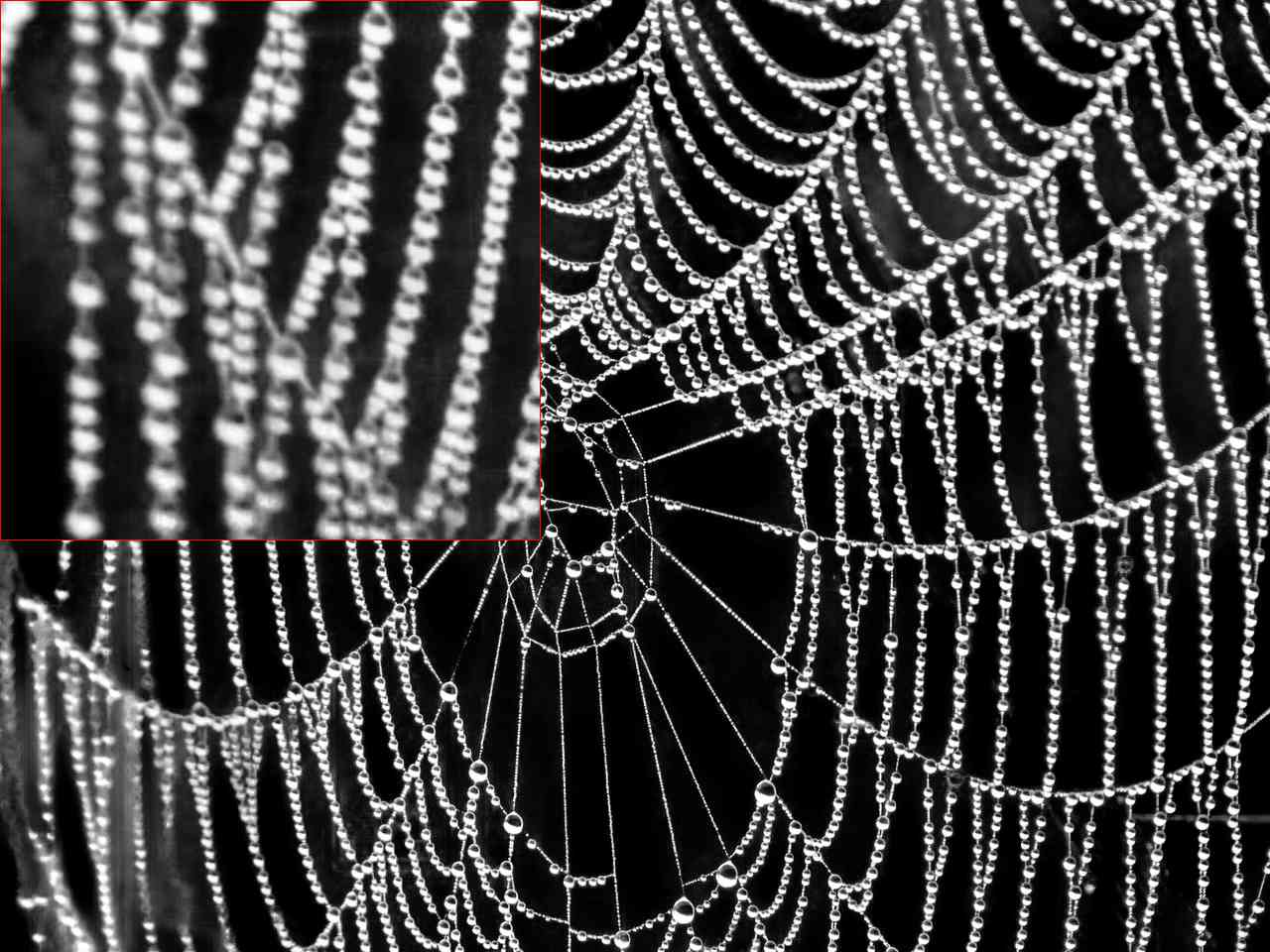}
\end{subfigure}

\begin{subfigure}{.16\linewidth}
  \centering
  \includegraphics[width=.99\linewidth]{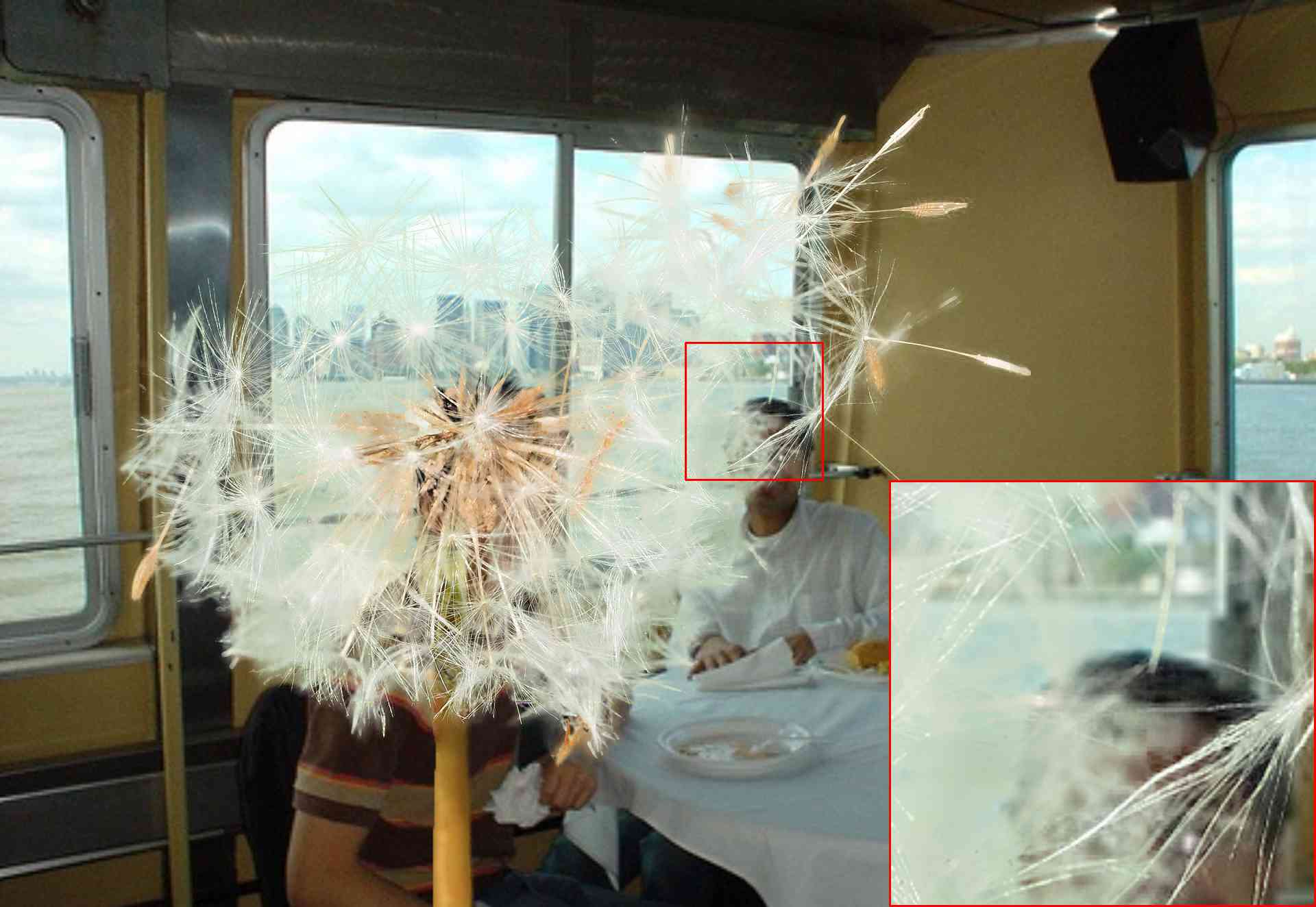}
\end{subfigure}
\begin{subfigure}{.16\linewidth}
  \centering
  \includegraphics[width=.99\linewidth]{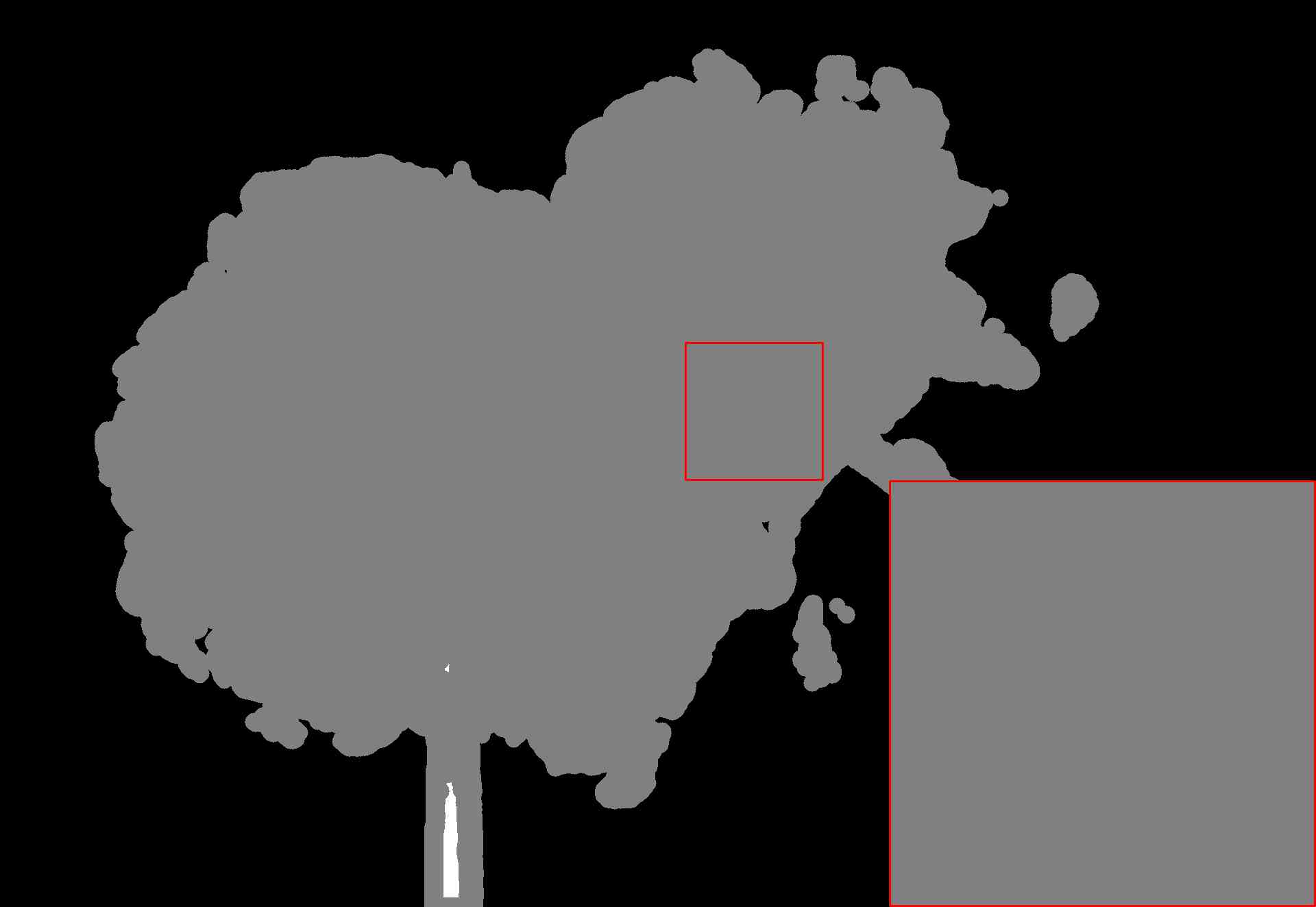}
\end{subfigure}
\begin{subfigure}{.16\linewidth}
  \centering
  \includegraphics[width=.99\linewidth]{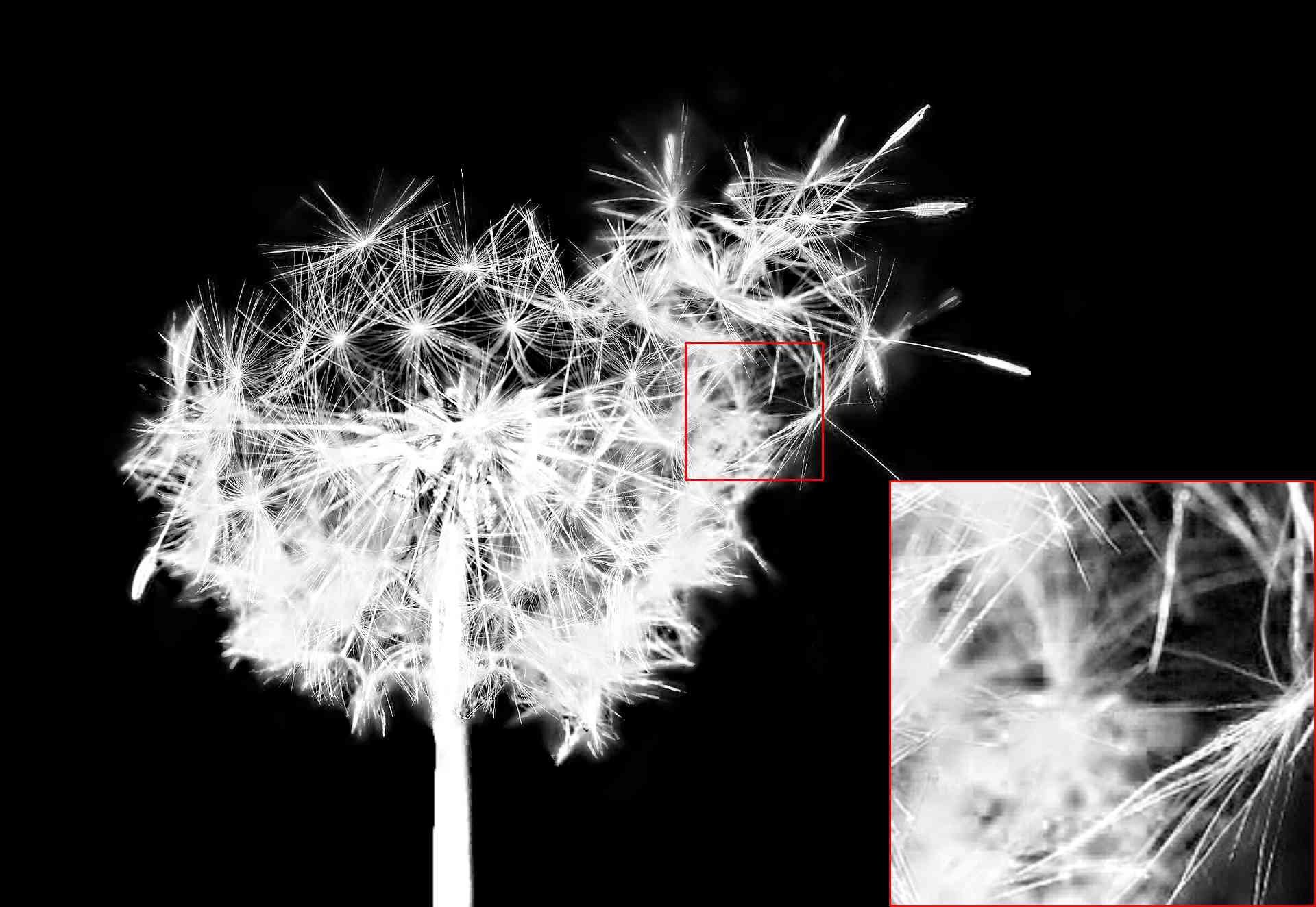}
\end{subfigure}
\begin{subfigure}{.16\linewidth}
  \centering
  \includegraphics[width=.99\linewidth]{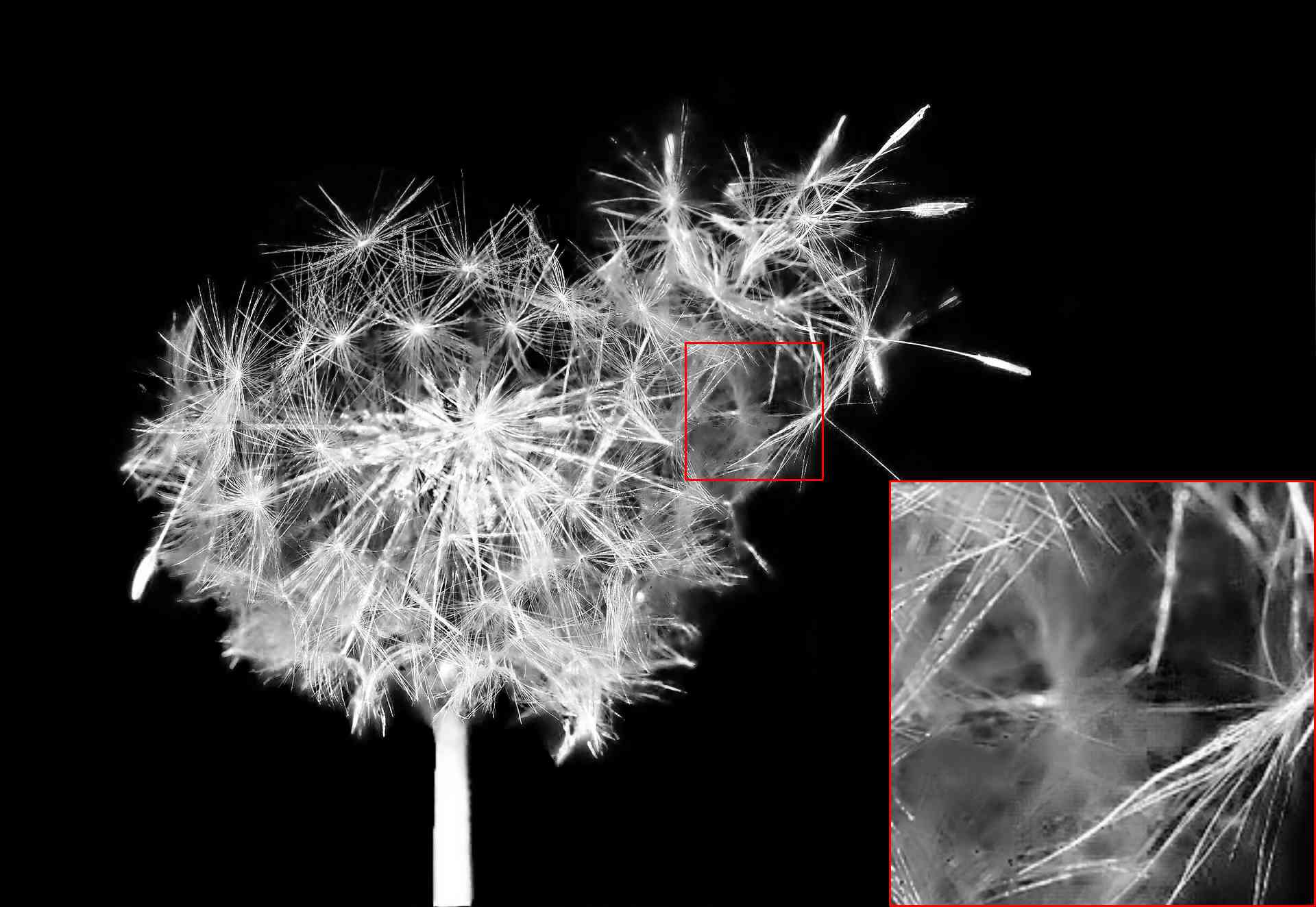}
\end{subfigure}
\begin{subfigure}{.16\linewidth}
  \centering
  \includegraphics[width=.99\linewidth]{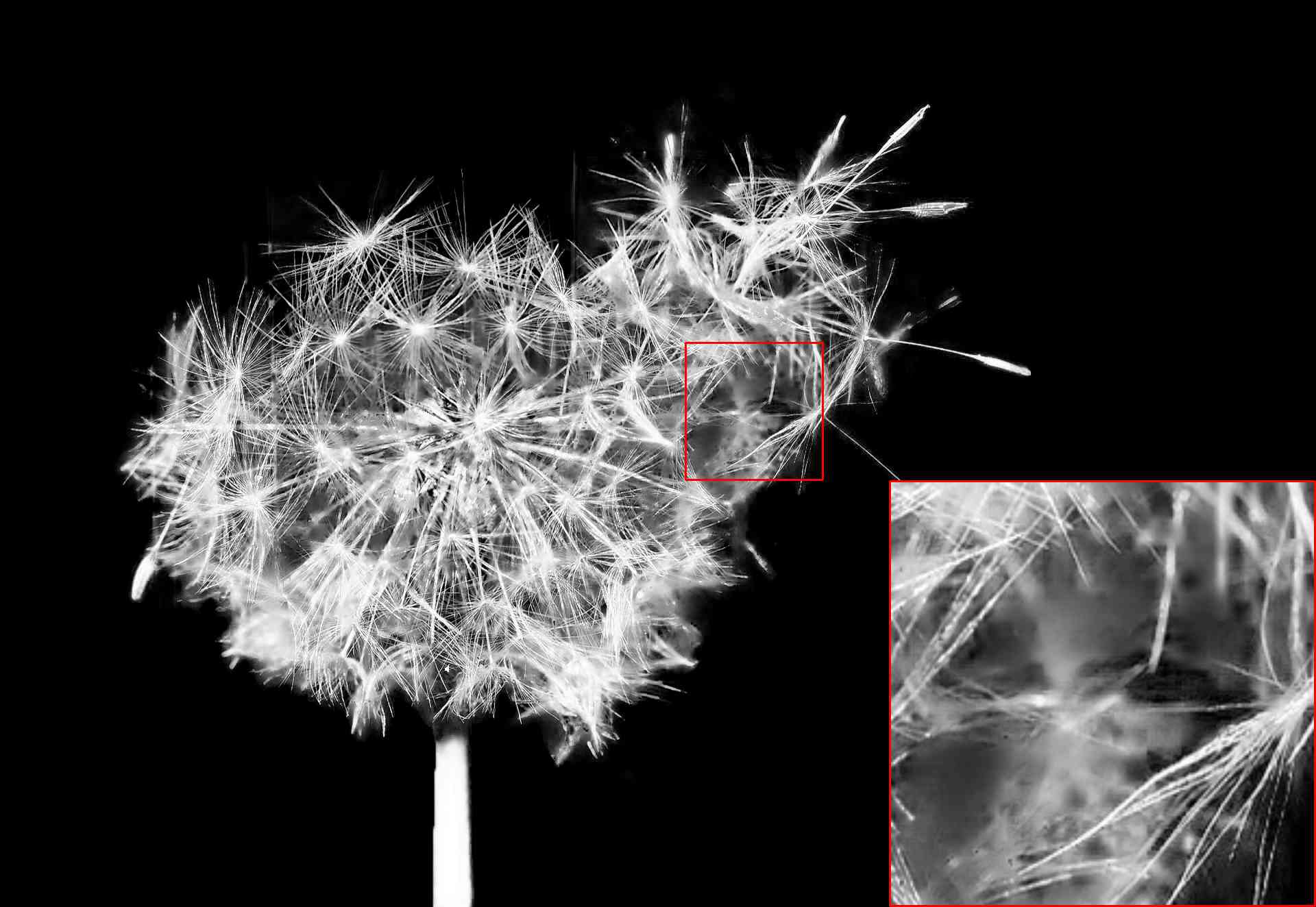}
\end{subfigure}
\begin{subfigure}{.16\linewidth}
  \centering
  \includegraphics[width=.99\linewidth]{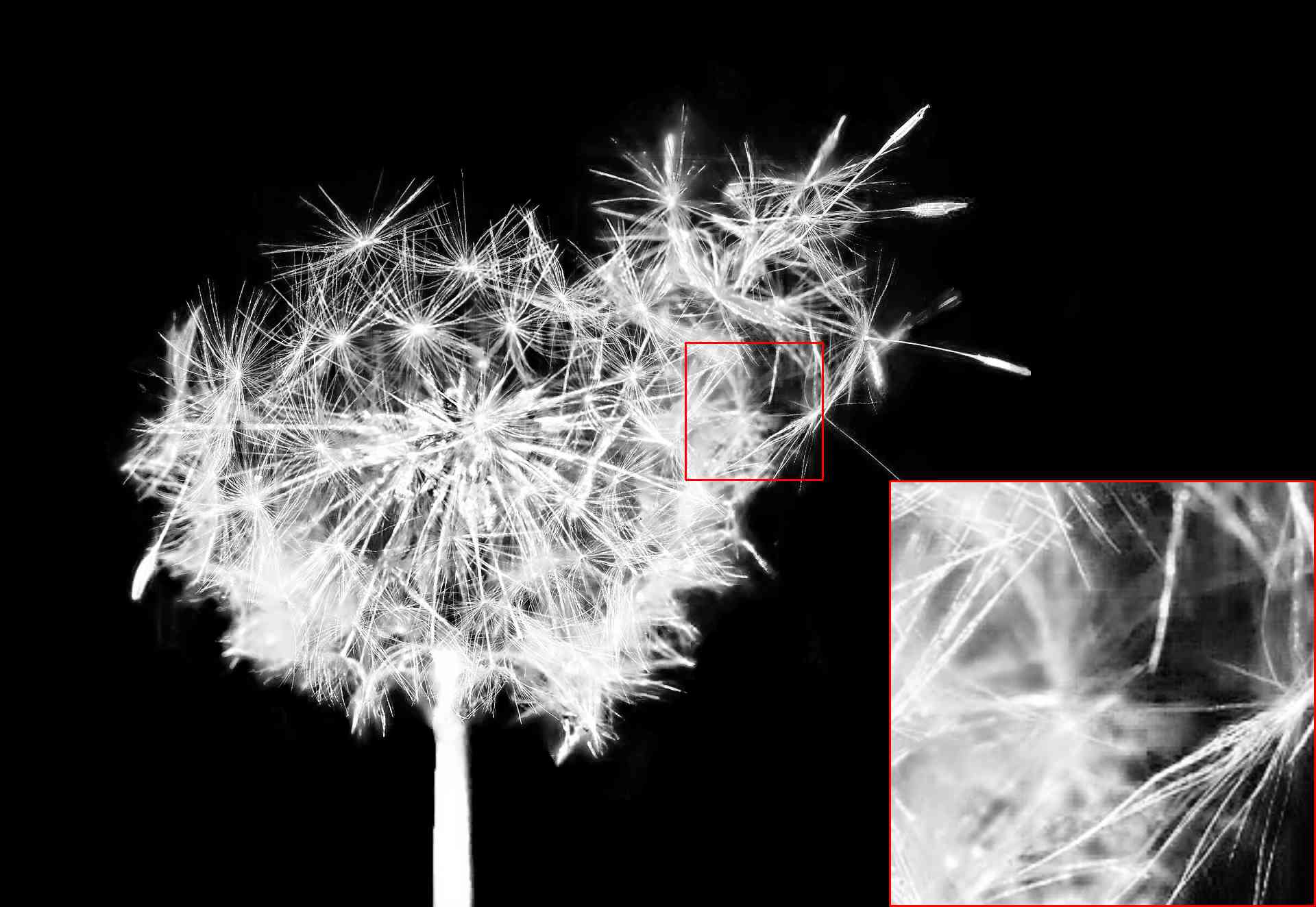}
\end{subfigure}

\begin{subfigure}{.16\linewidth}
  \centering
  \includegraphics[width=.99\linewidth]{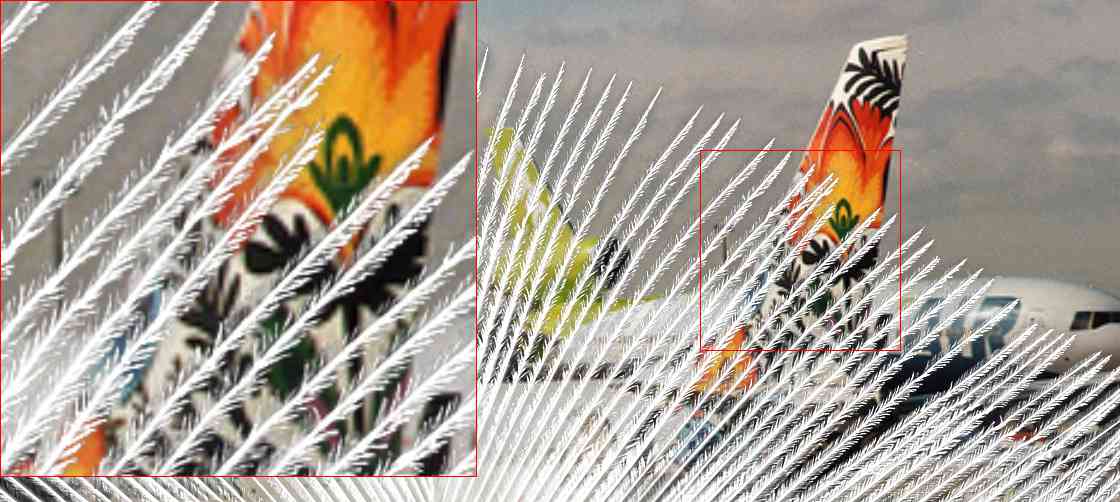}
  \caption{Image}
\end{subfigure}
\begin{subfigure}{.16\linewidth}
  \centering
  \includegraphics[width=.99\linewidth]{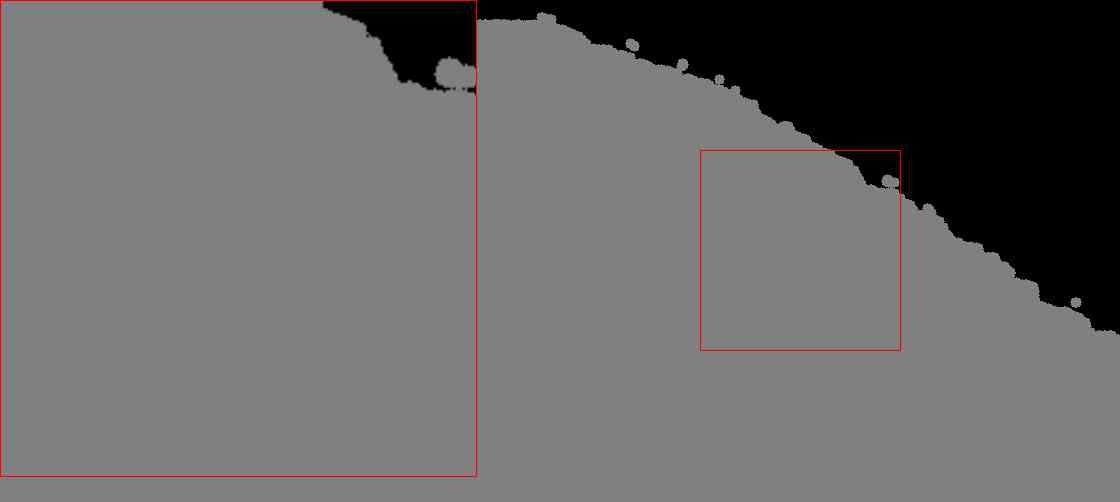}
  \caption{Trimap}
\end{subfigure}
\begin{subfigure}{.16\linewidth}
  \centering
  \includegraphics[width=.99\linewidth]{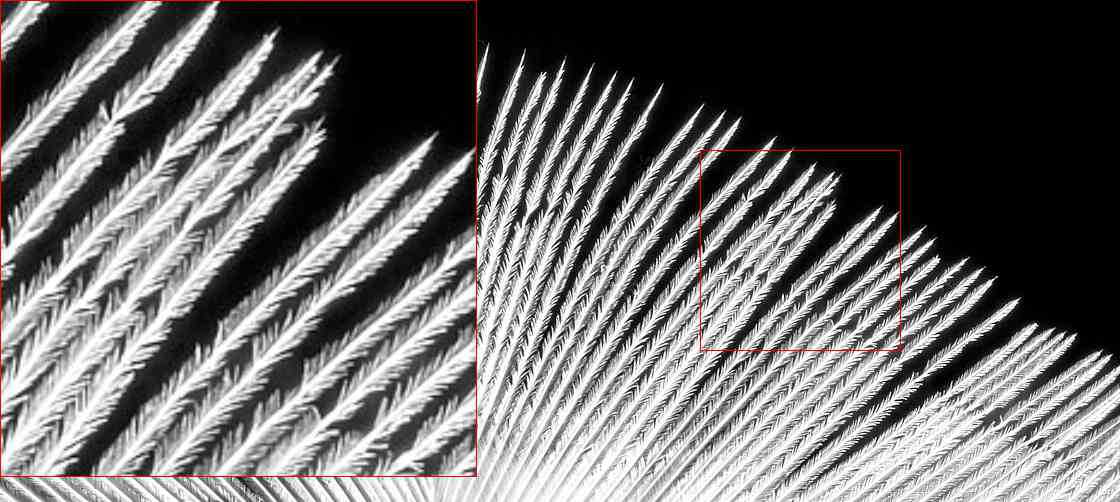}
  \caption{Ground Truth}
\end{subfigure}
\begin{subfigure}{.16\linewidth}
  \centering
  \includegraphics[width=.99\linewidth]{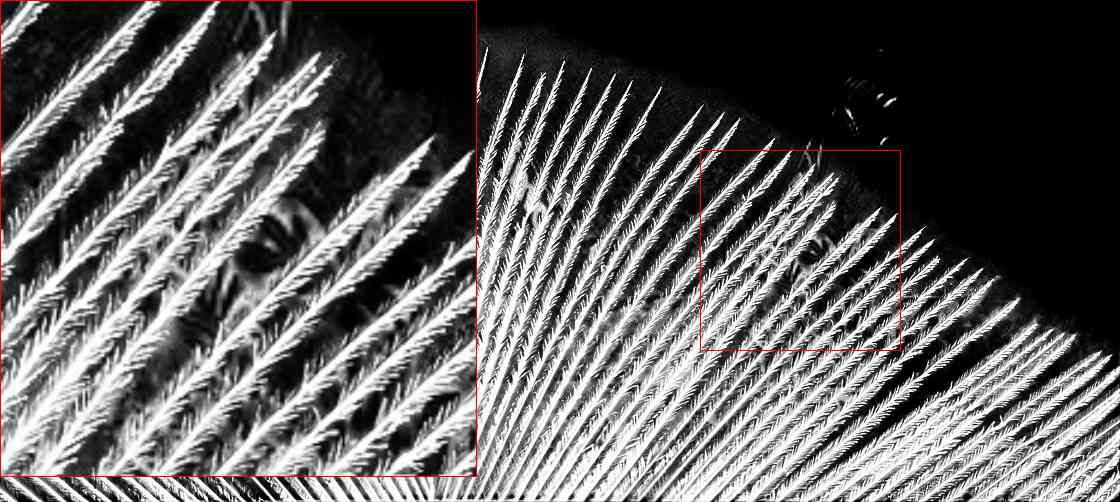}
  \caption{ContexNet}
\end{subfigure}
\begin{subfigure}{.16\linewidth}
  \centering
  \includegraphics[width=.99\linewidth]{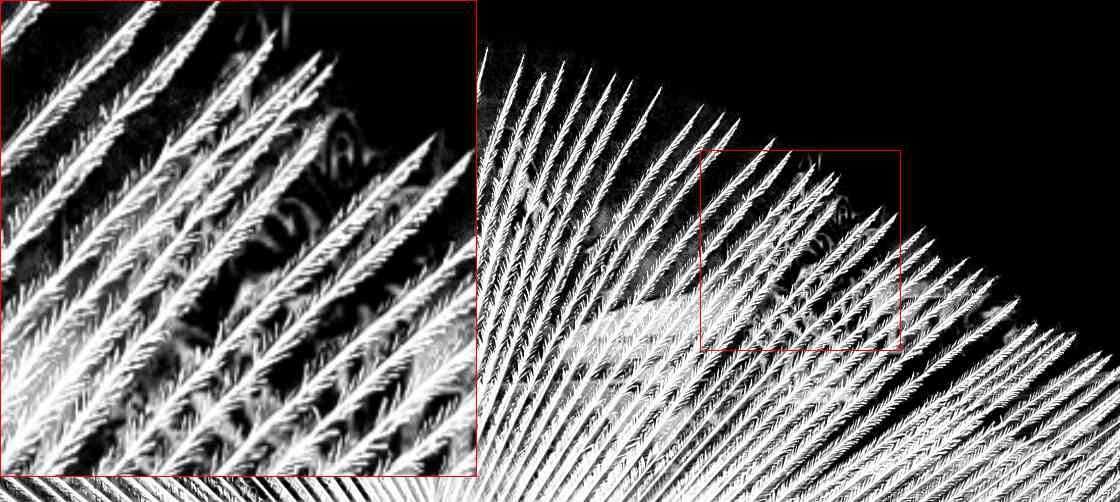}
  \caption{IndexNet}
\end{subfigure}
\begin{subfigure}{.16\linewidth}
  \centering
  \includegraphics[width=.99\linewidth]{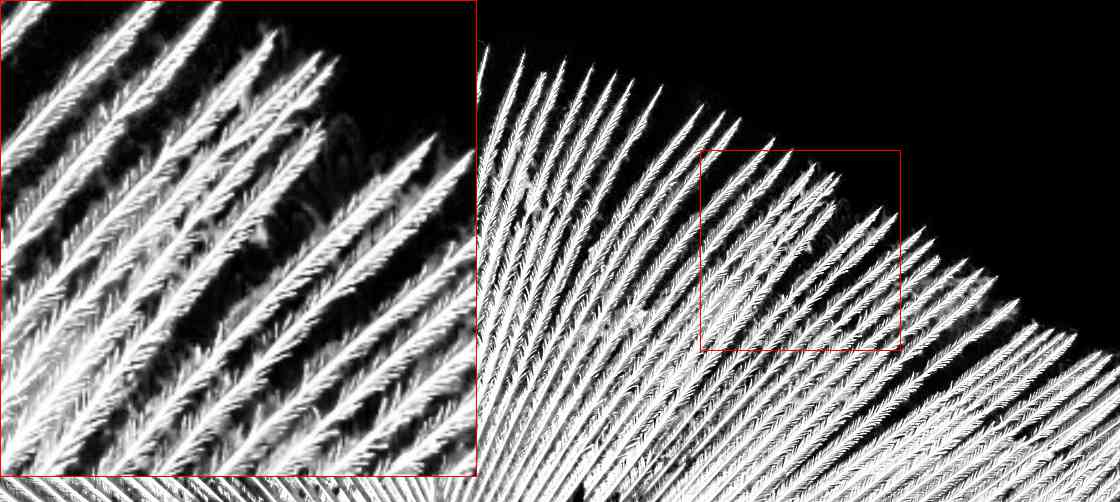}
  \caption{HDMatt (Ours)}
\end{subfigure}
\caption{Visual results on AIM testset. Our method has an obvious advantage in large unknown regions.}
\label{fig:visual_sota}
\vspace{-1.5em}
\end{figure*}

\subsection{Dataset}
We trained our models on Adobe Image Matting (AIM) dataset~\cite{xu2017deep}. AIM has 431 foreground images for training, each of which has a fine-annotated alpha matte. We augmented the data following Tang~\etal~\cite{tang2019learning}. Specifically, we first augmented the ground truth alpha matte by compositing two foreground images. To generate the associated trimaps, we randomly dilated ground truth alpha mattes. The synthetic training images will be the compositions of a foreground images in augmented AIM training set with a randomly sampled background image from COCO dataset~\cite{lin2014microsoft}. For each training image, we sampled a pixel from unknown area and crop a square patch centered at this pixel with side length in $\{640, 480, 320\}$. Then, these patches are resized to $320\times320$ and randomly flipped horizontally and rotated by an angle less than $15^\circ$. We tested our models on AIM testset, AlphaMatting~\cite{rhemann2009perceptually} and newly collected HR real-world images.

\subsection{Implementation}
To optimize the entire framework, we used a two-stage training strategy. On the first stage, we pre-trained a Resnet-34 classification model on ImageNet~\cite{imagenet_cvpr09}. We follow the same training configuration as the public PyTorch implementation~\footnote{https://github.com/pytorch/examples/blob/master/imagenet}. Then all layers from the model before the fully-connected layer were used as our matting encoder. We changed the last two convolution blocks with dilated convolution to keep HR feature maps as in \cite{chen2017rethinking}. On the second stage, the encoder-decoder model together with CPC module was jointly trained end-to-end for image matting. During both training and testing, we set context patch number $K=3$ (\ie~top-3 patches) by default in all the following experiments unless otherwise stated. We also make an ablation study on different choices of $K$. When input image is of ultra-high resolution, we set the candidate context patches number $N=30$ to save computational resources. Adam~\cite{kingma2014adam} optimizer was used with initial learning rate $0.5\times10^{-3}$ and decayed by cosine scheduler. The model is trained for $200k$ steps with batch size 32 and weight decay $10^{-4}$. Our experiments are implemented in PyTorch~\cite{paszke2017automatic}.

\begin{figure*}[th]
\centering
\begin{subfigure}{.193\linewidth}
  \centering
  \includegraphics[width=.99\linewidth]{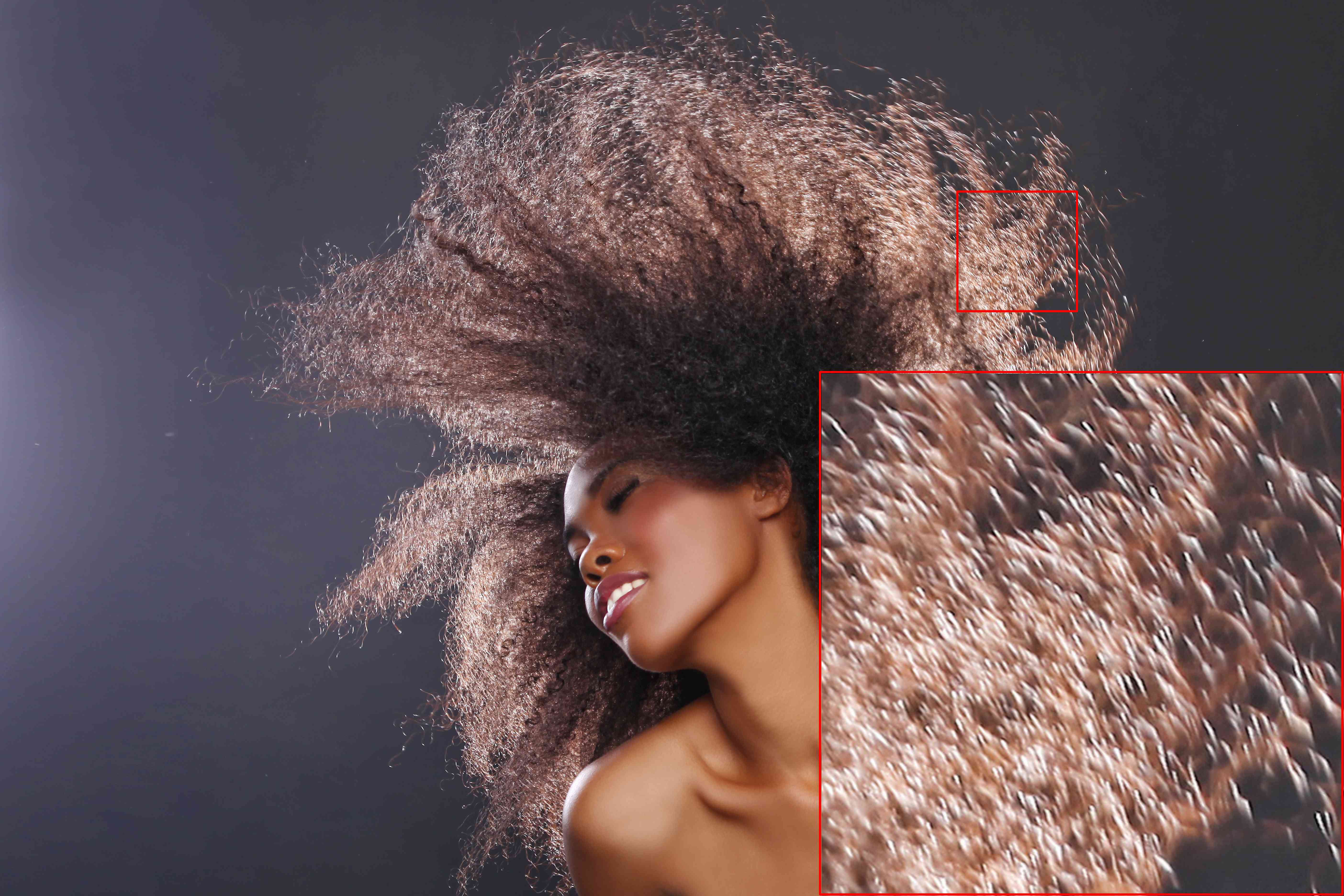} 
\end{subfigure}
\begin{subfigure}{.193\linewidth}
  \centering
  \includegraphics[width=.99\linewidth]{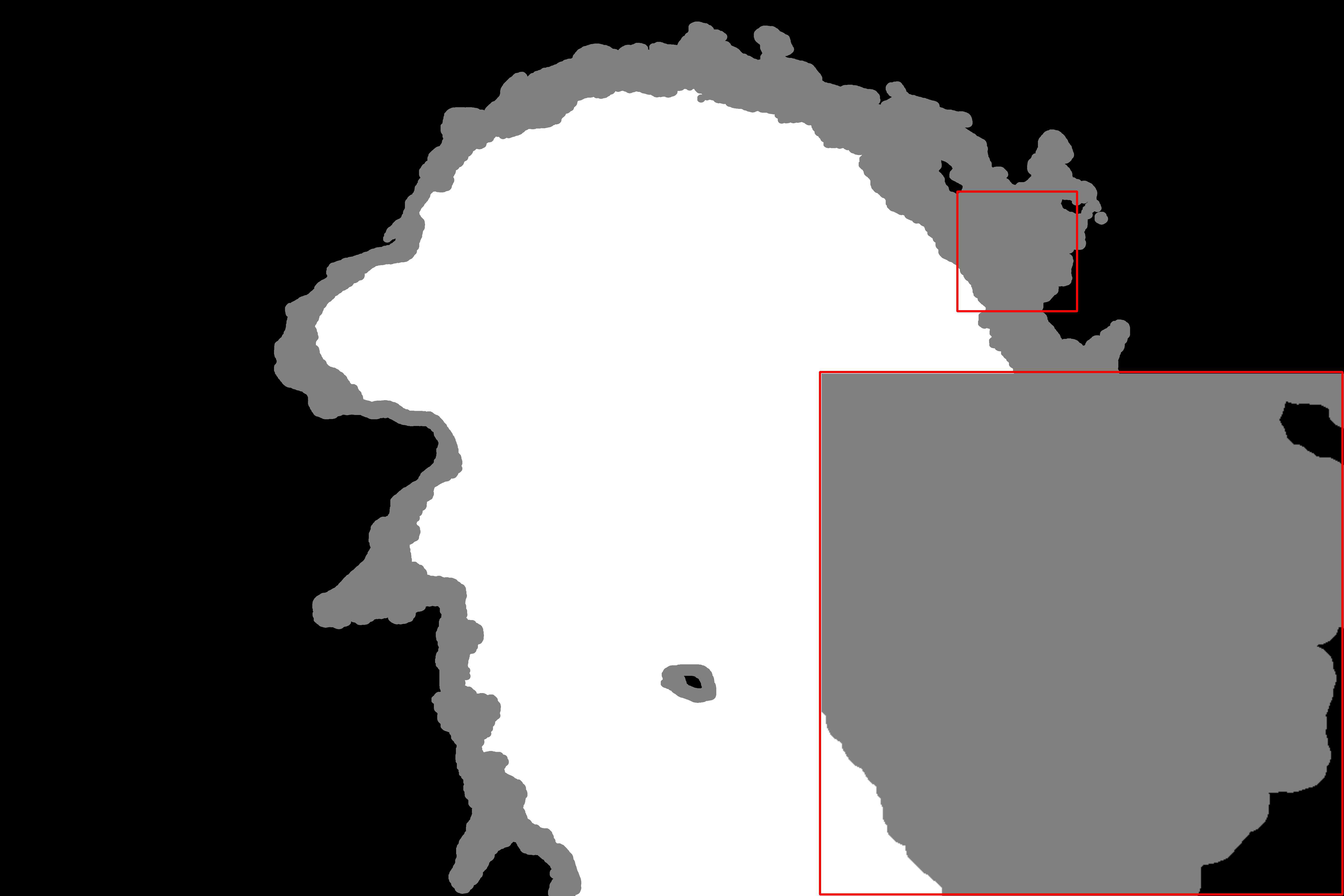}
\end{subfigure}
\begin{subfigure}{.193\linewidth}
  \centering
  \includegraphics[width=.99\linewidth]{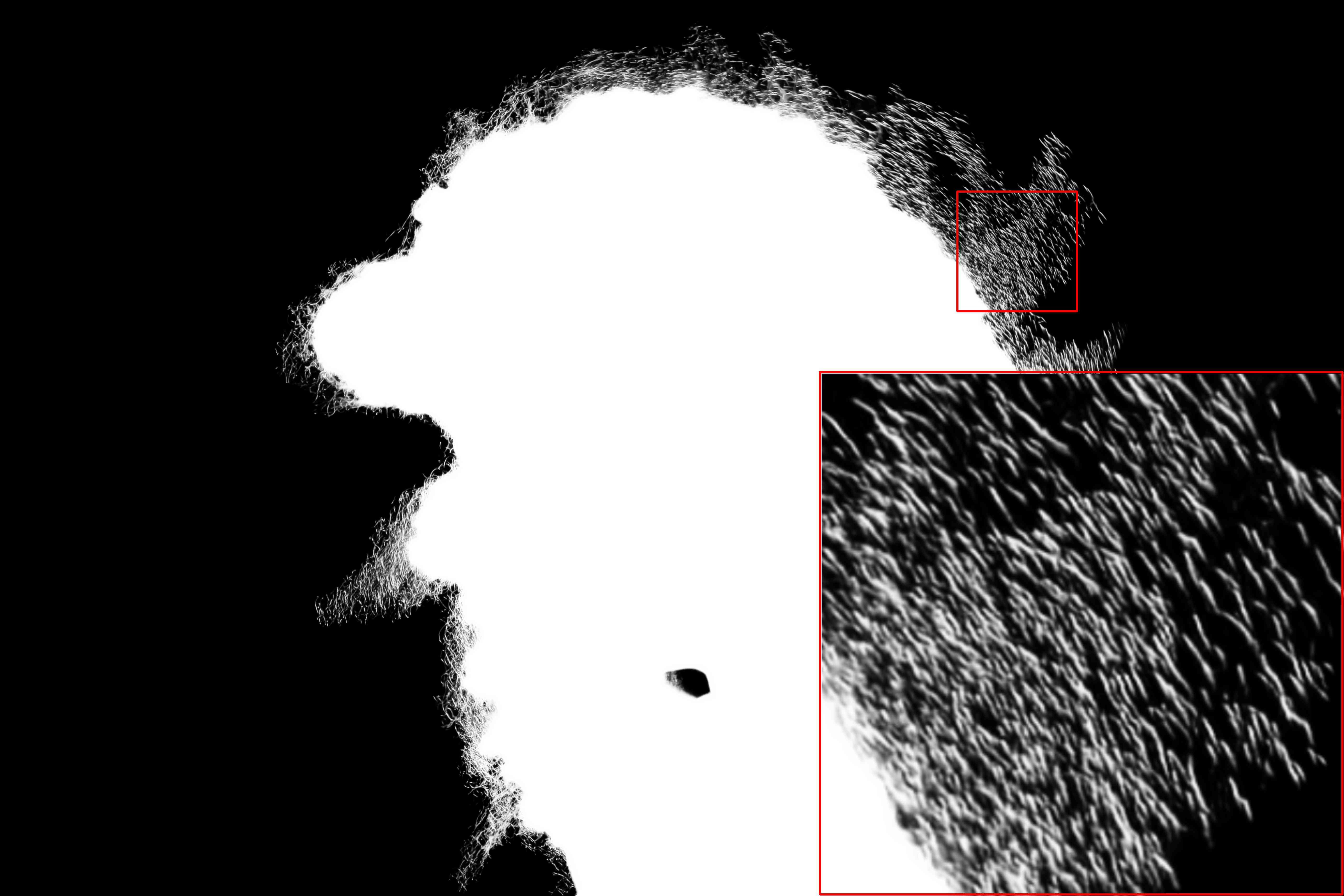}
\end{subfigure}
\begin{subfigure}{.193\linewidth}
  \centering
  \includegraphics[width=.99\linewidth]{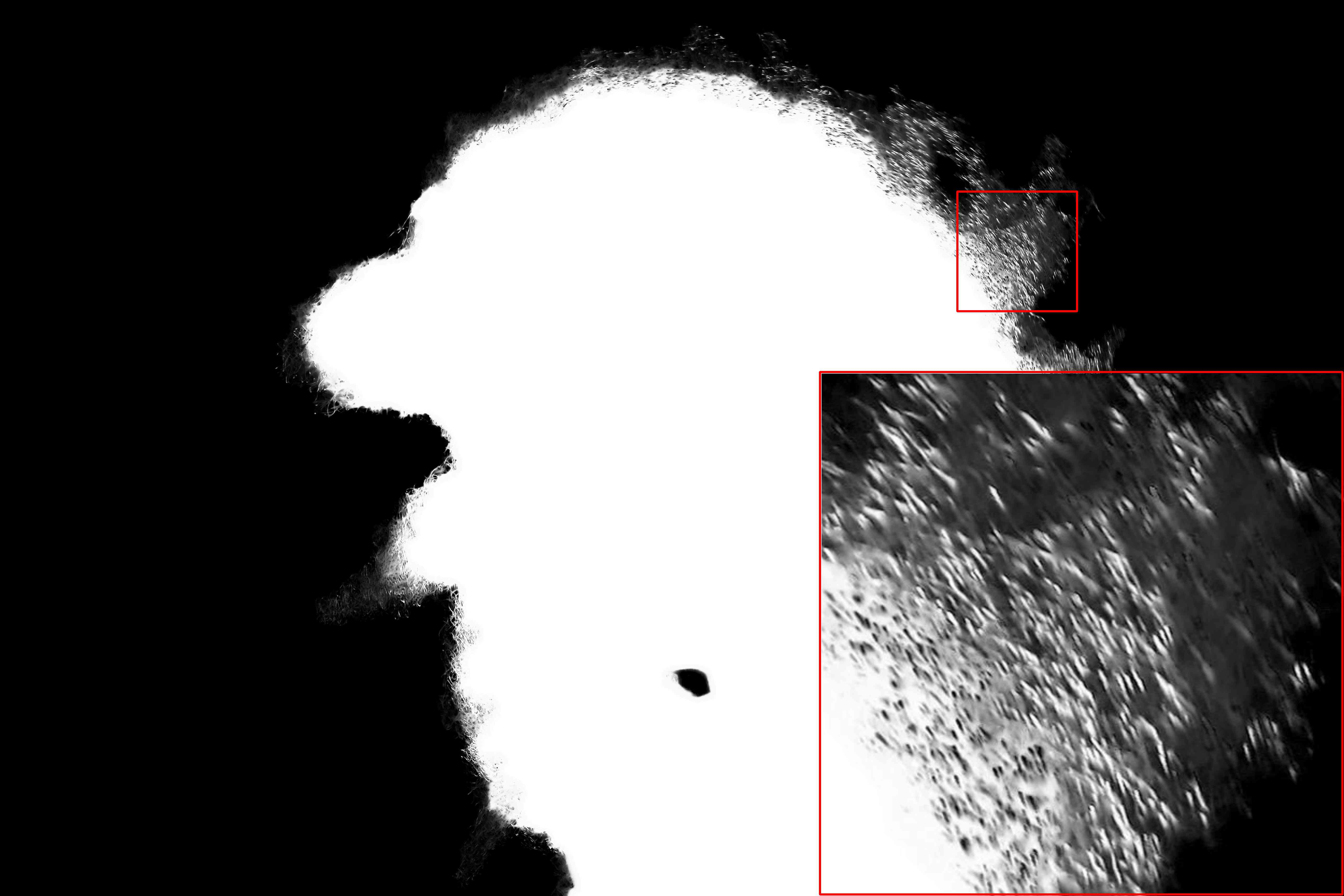}
\end{subfigure}
\begin{subfigure}{.193\linewidth}
  \centering
  \includegraphics[width=.99\linewidth]{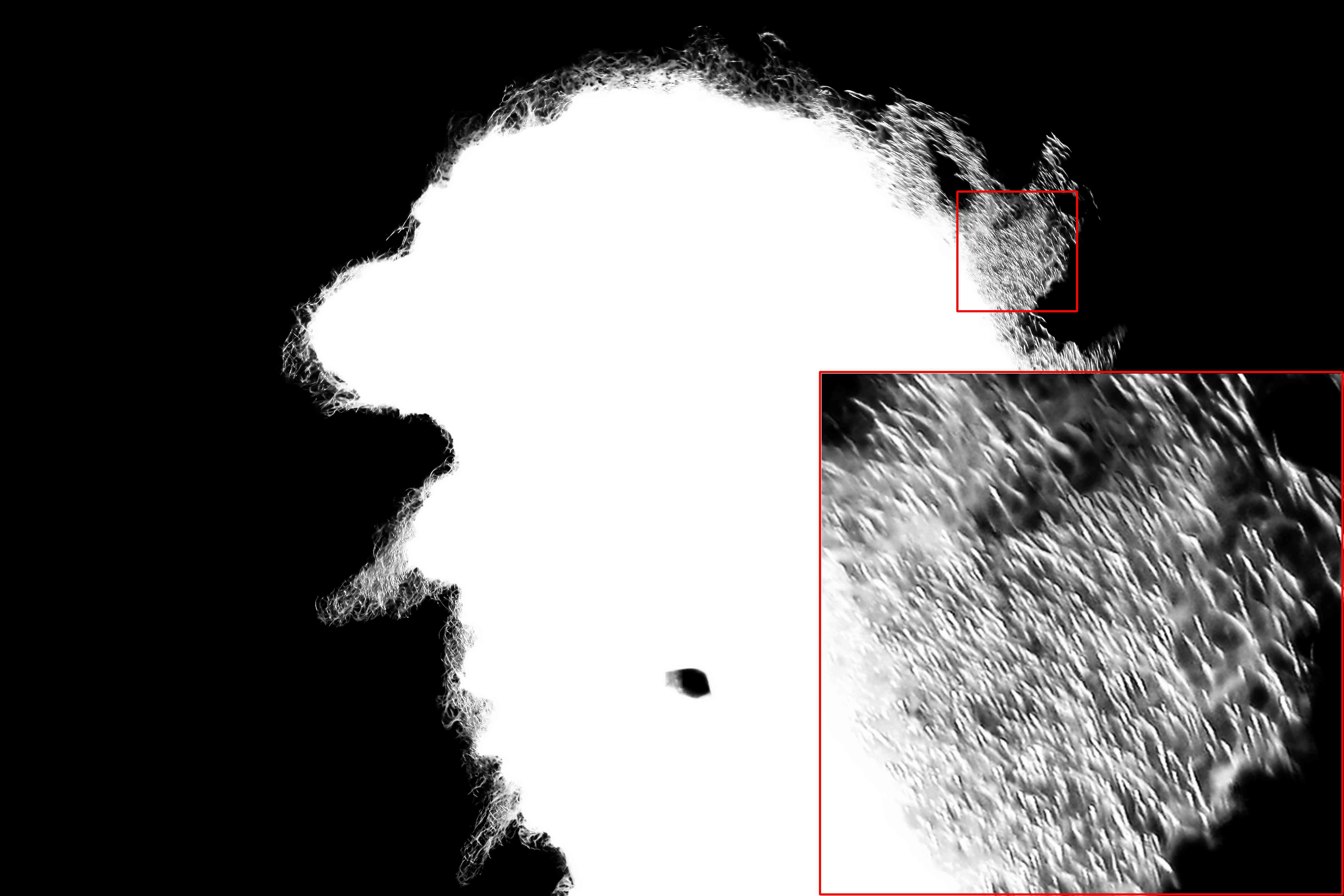}
\end{subfigure}
\begin{subfigure}{.193\linewidth}
  \centering
  \includegraphics[width=.99\linewidth]{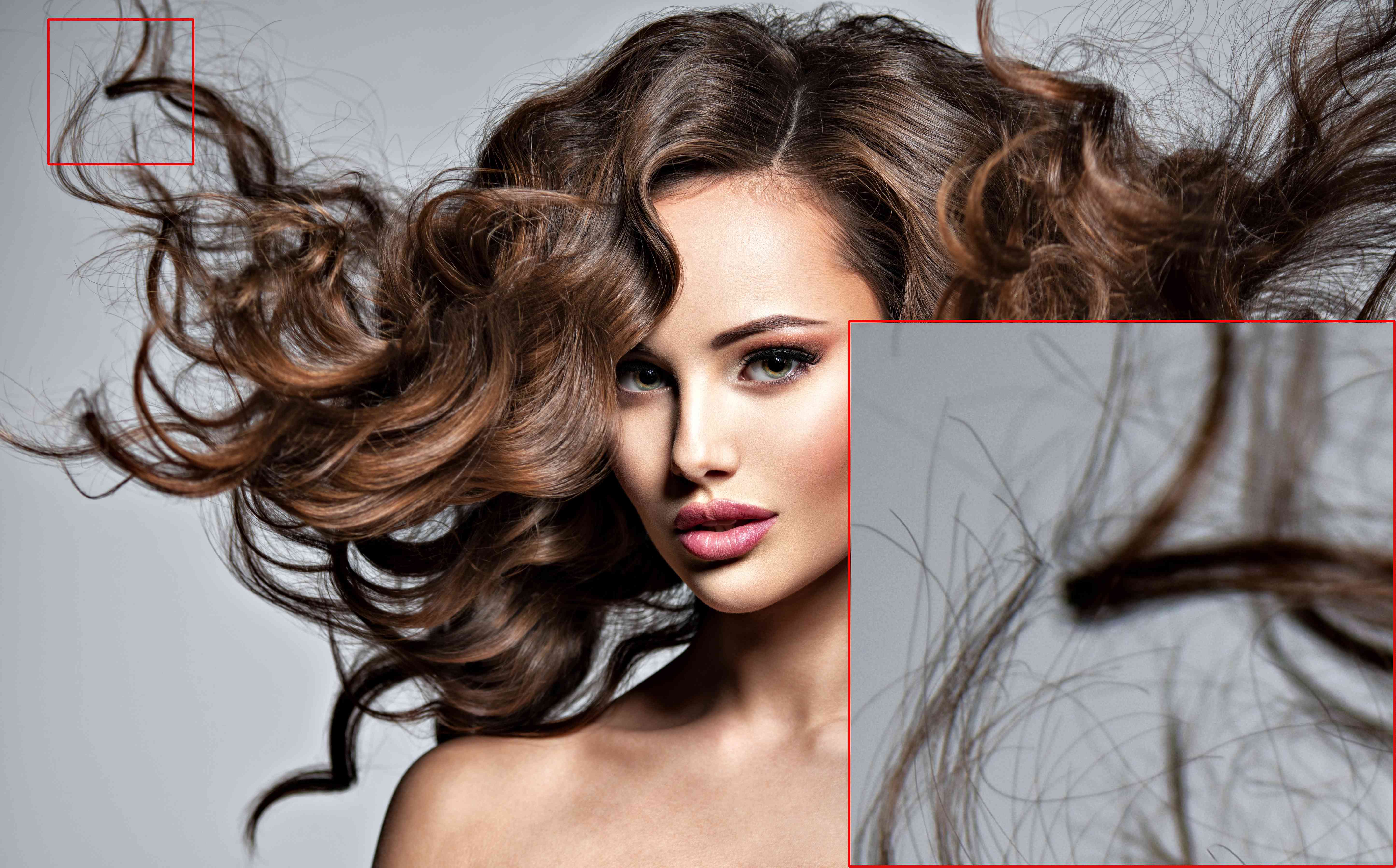} 
\end{subfigure}
\begin{subfigure}{.193\linewidth}
  \centering
  \includegraphics[width=.99\linewidth]{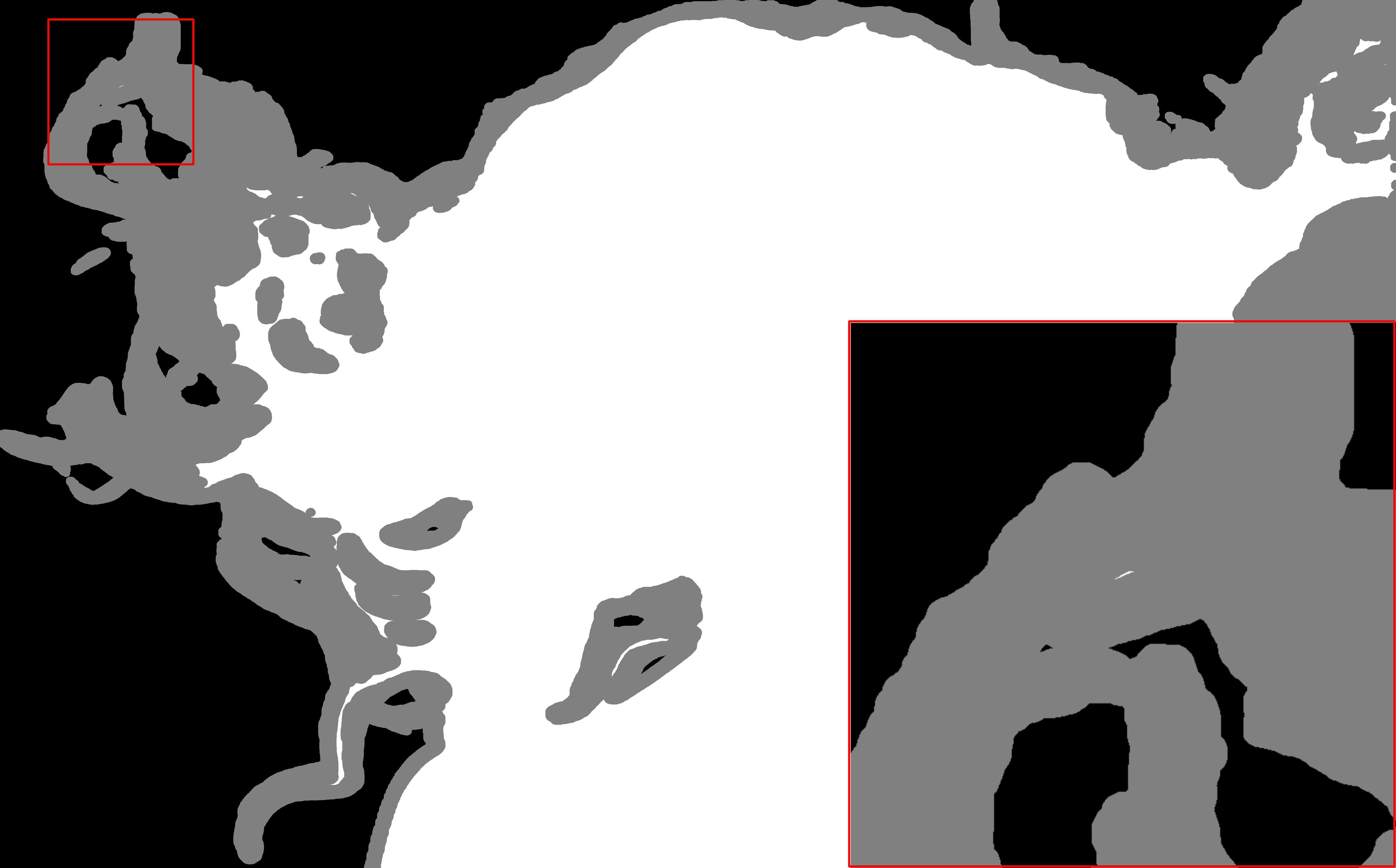}
\end{subfigure}
\begin{subfigure}{.193\linewidth}
  \centering
  \includegraphics[width=.99\linewidth]{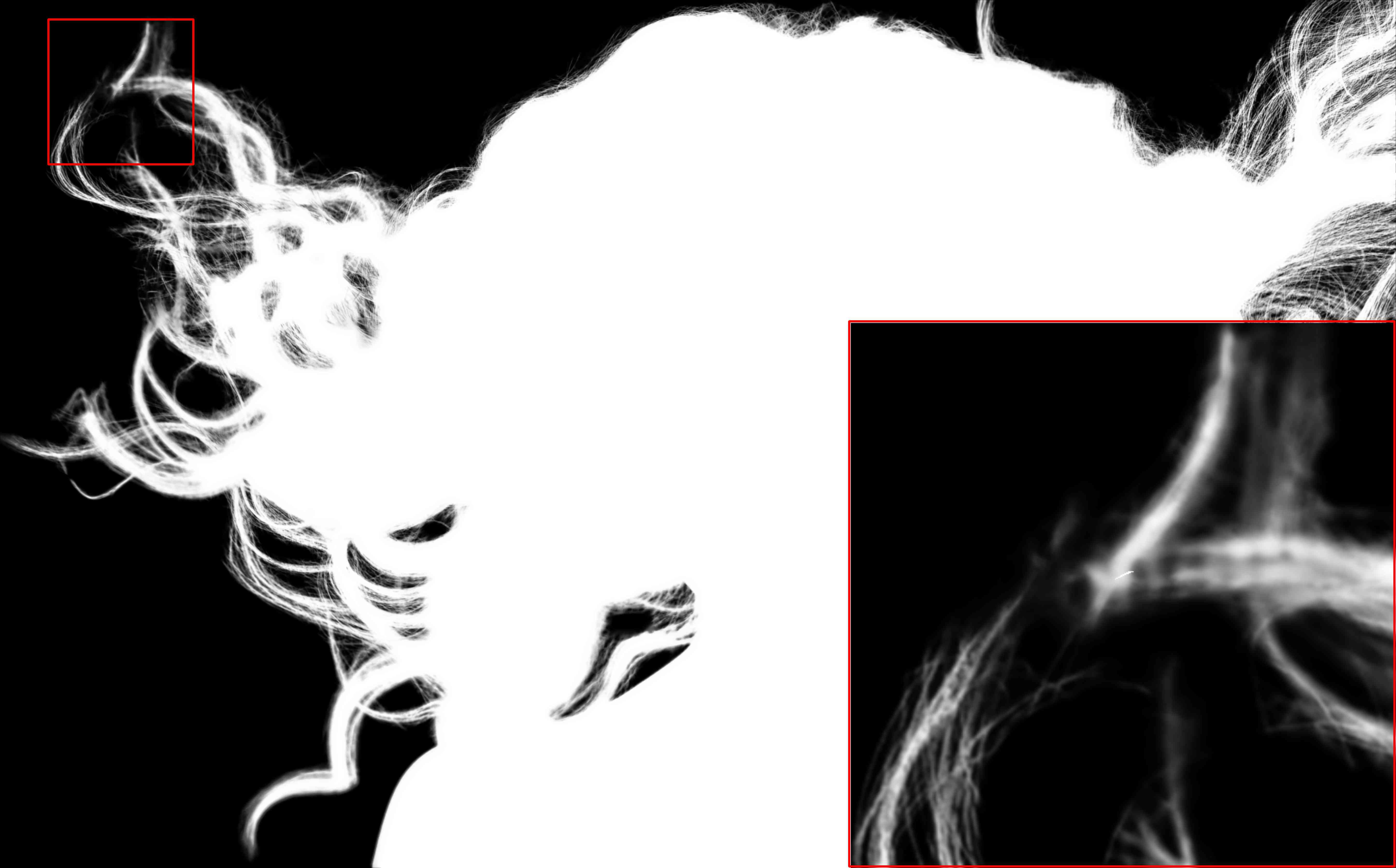}
\end{subfigure}
\begin{subfigure}{.193\linewidth}
  \centering
  \includegraphics[width=.99\linewidth]{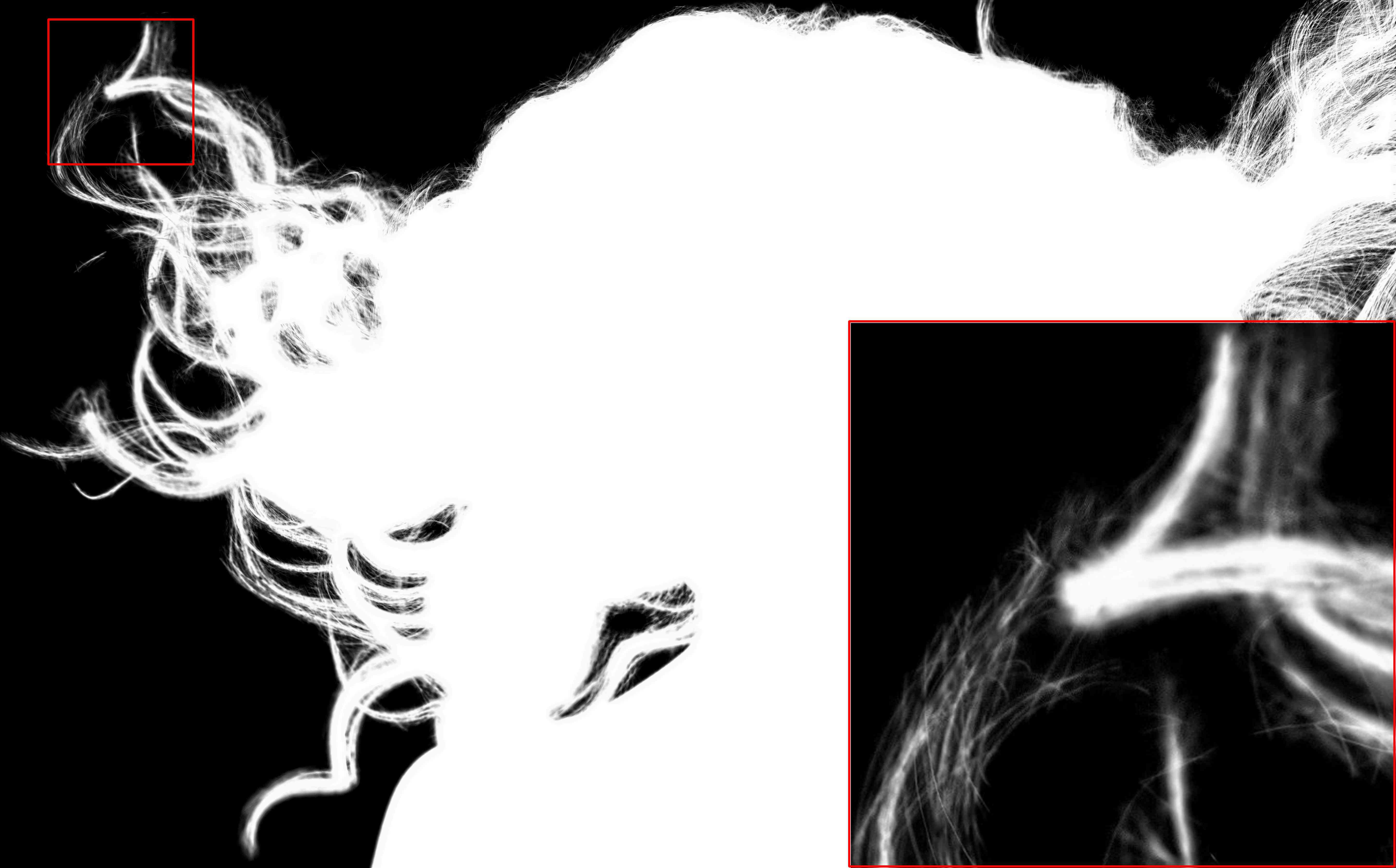}
\end{subfigure}
\begin{subfigure}{.193\linewidth}
  \centering
  \includegraphics[width=.99\linewidth]{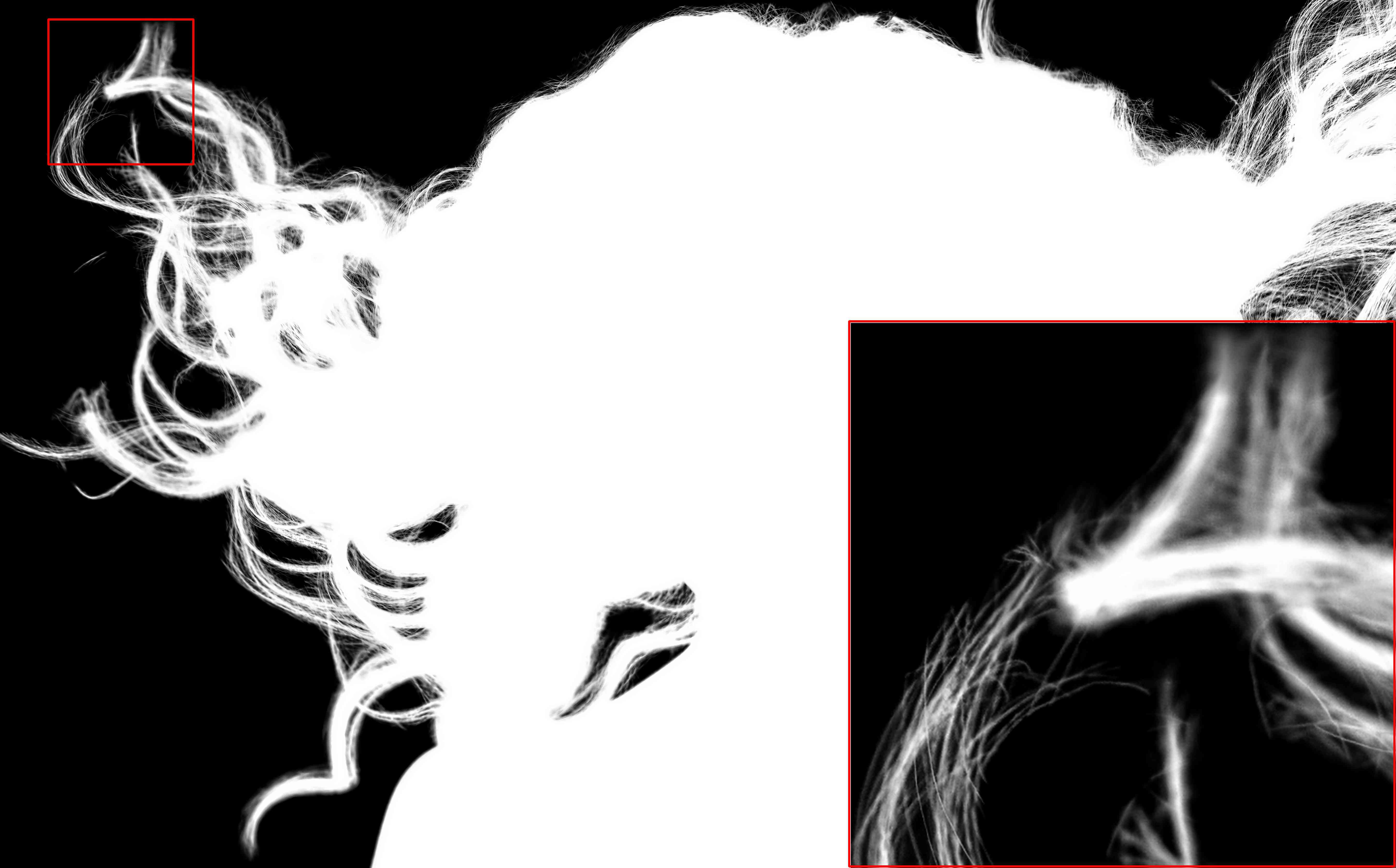}
\end{subfigure}
\begin{subfigure}{.193\linewidth}
  \centering
  \includegraphics[width=.99\linewidth]{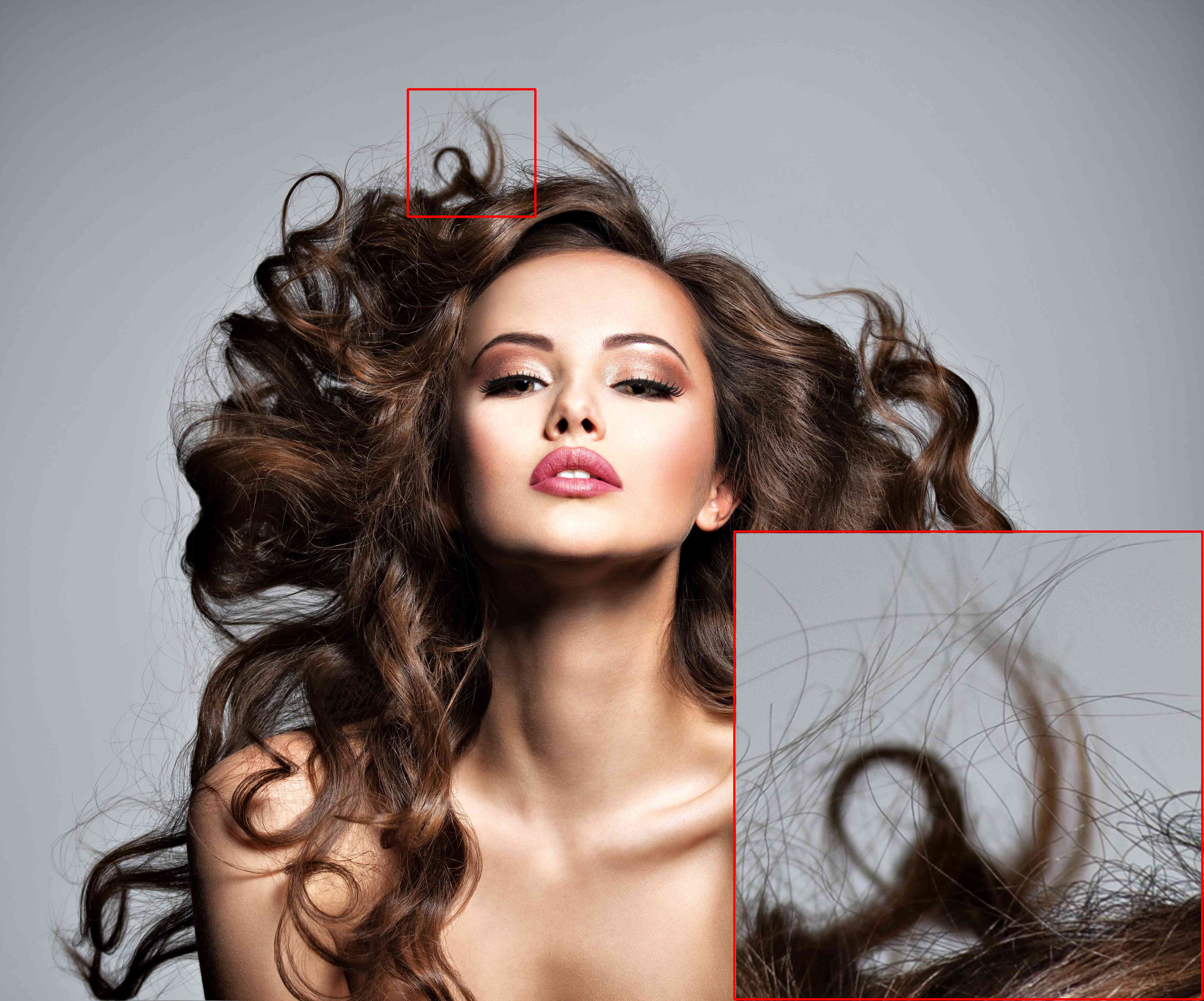} 
  \caption{HR Image}
\end{subfigure}
\begin{subfigure}{.193\linewidth}
  \centering
  \includegraphics[width=.99\linewidth]{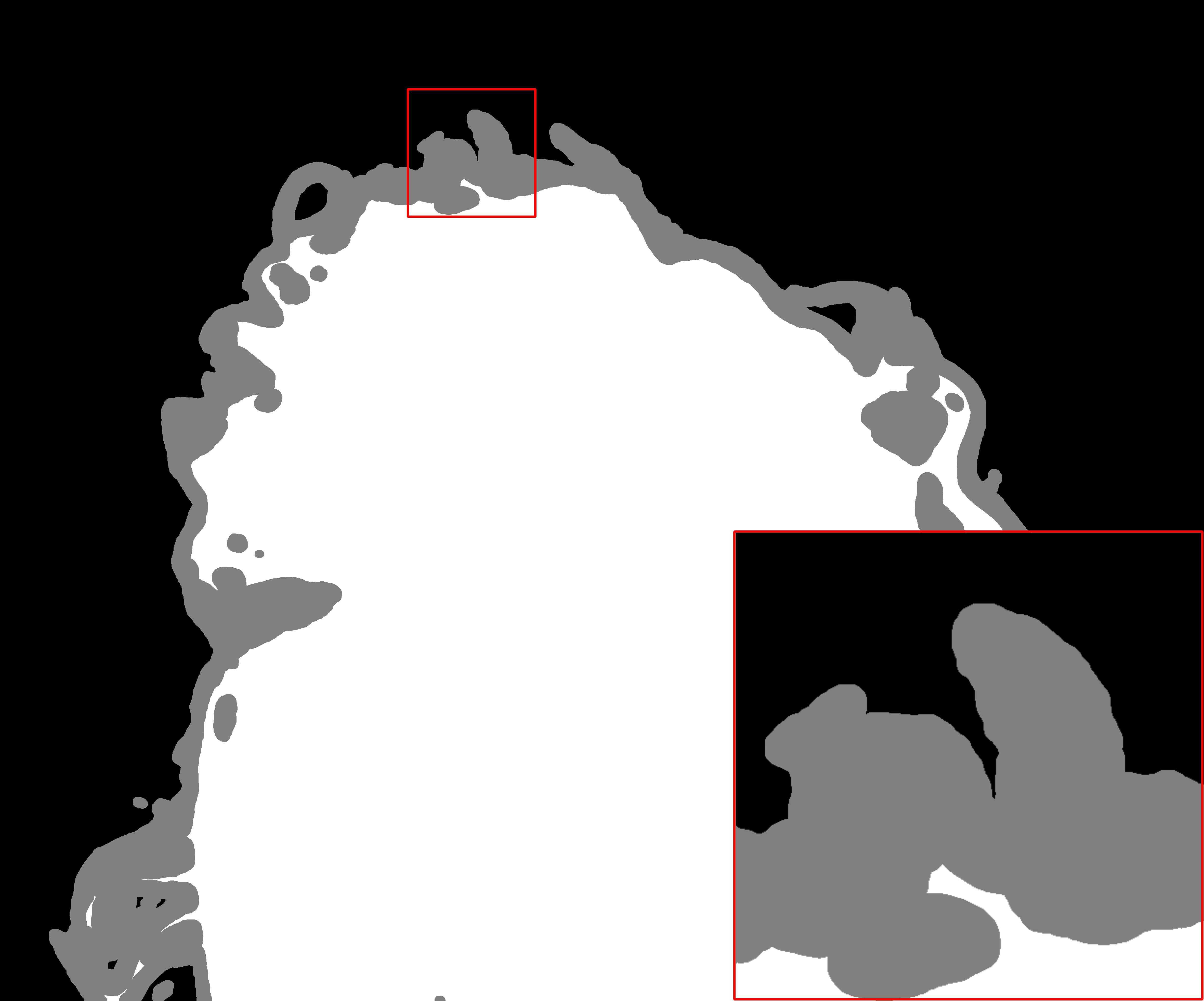}
  \caption{Trimap}
\end{subfigure}
\begin{subfigure}{.193\linewidth}
  \centering
  \includegraphics[width=.99\linewidth]{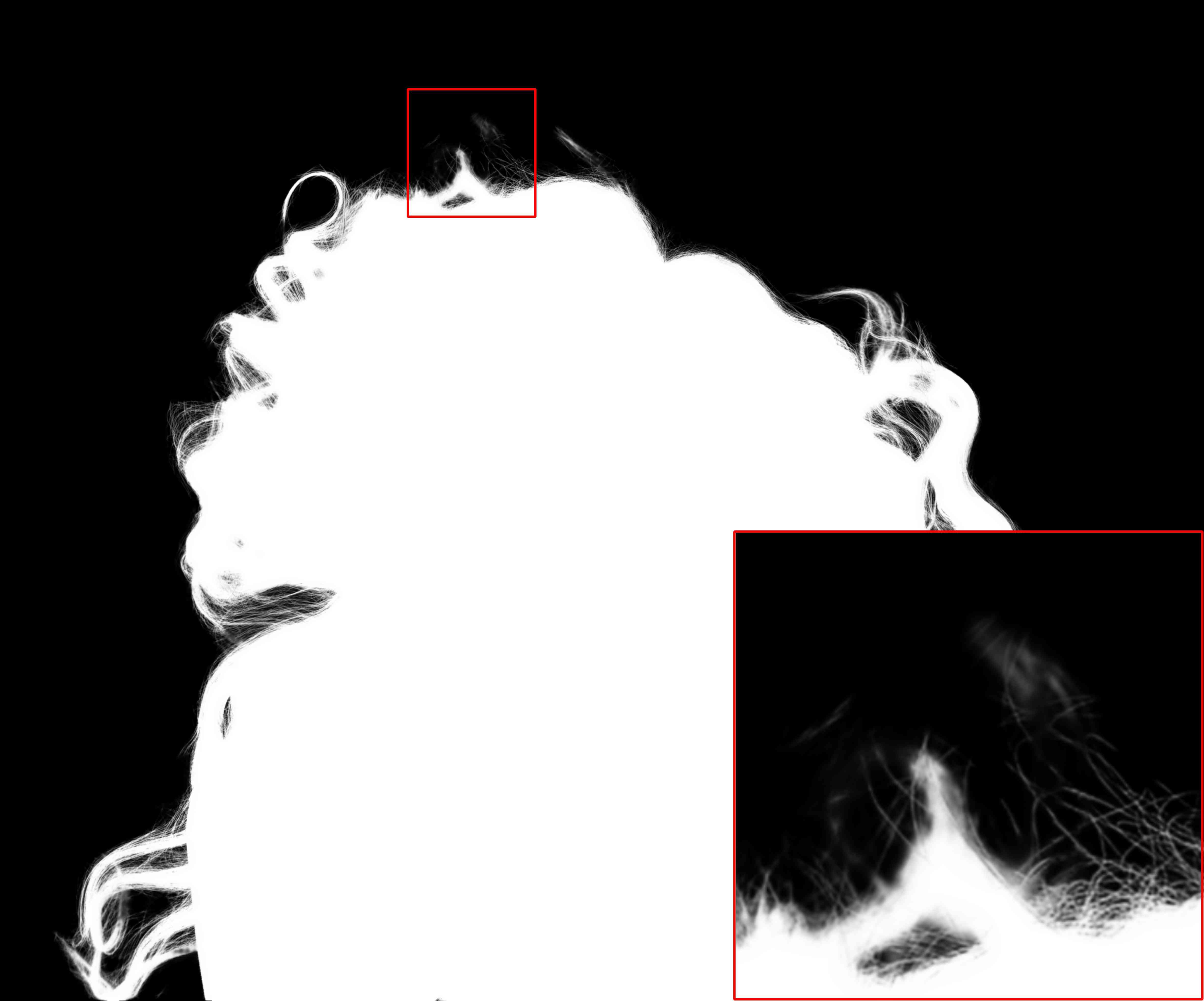}
  \caption{ContexNet}
\end{subfigure}
\begin{subfigure}{.193\linewidth}
  \centering
  \includegraphics[width=.99\linewidth]{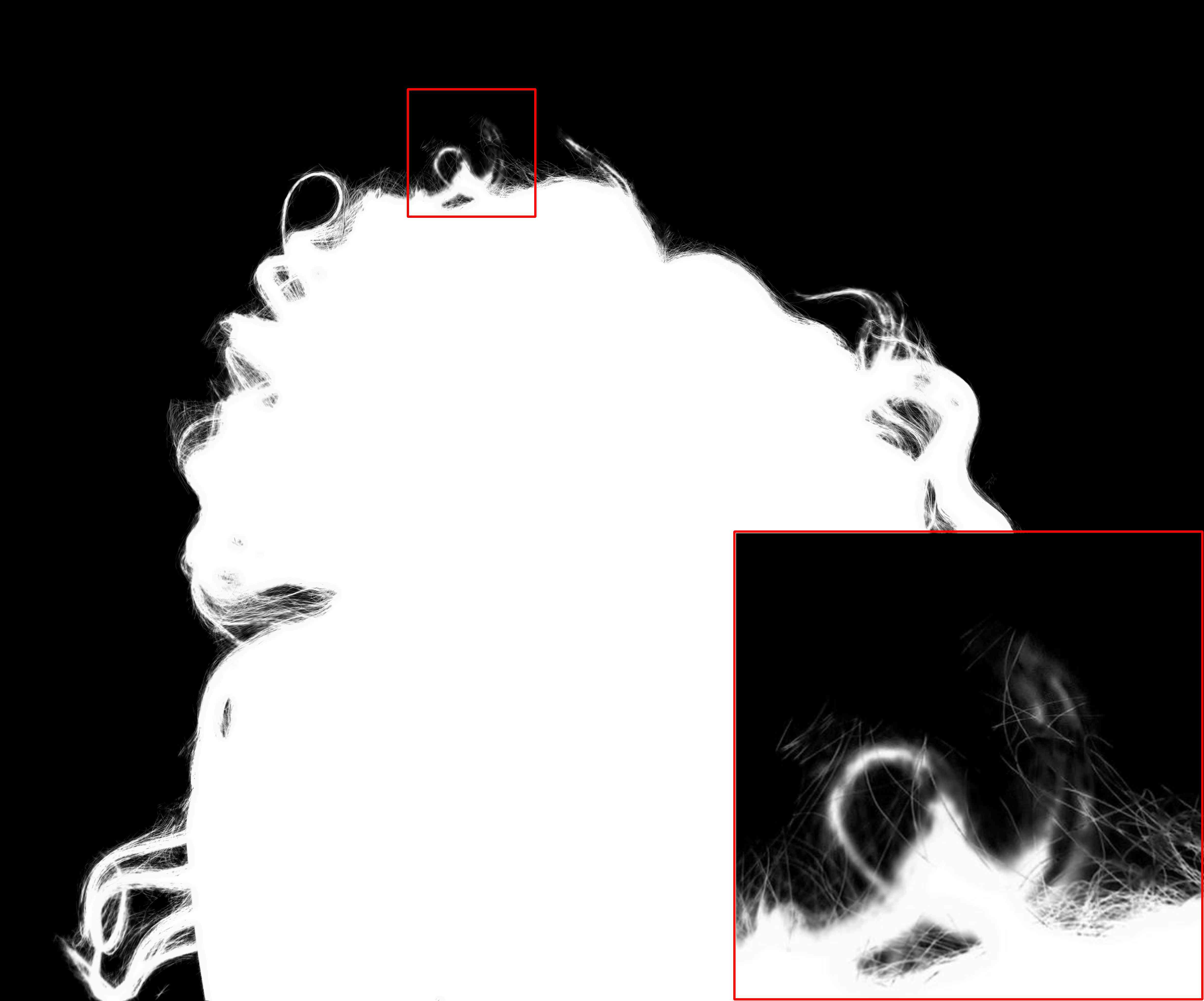}
  \caption{IndexNet}
\end{subfigure}
\begin{subfigure}{.193\linewidth}
  \centering
  \includegraphics[width=.99\linewidth]{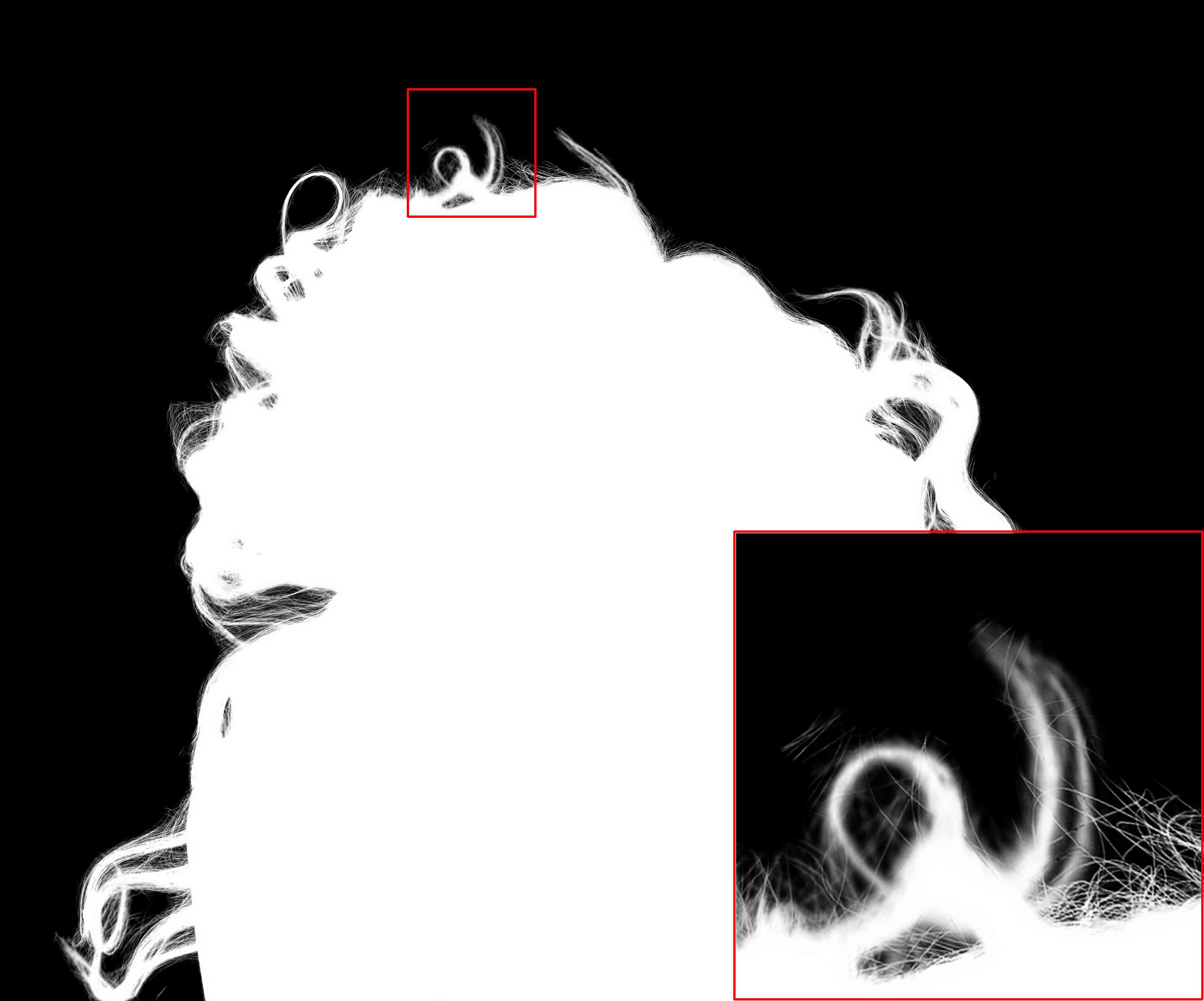}
  \caption{HDMatt (Ours)}
\end{subfigure}
\caption{Visual comparison on real-world HR images. We test ContextNet and IndexNet on CPUs on the full images. Zoom in for details. Image sizes from top to bottom: $5616\times3744$, $5779\times3594$, $4724\times3929$.  }
\label{fig:real1}
\vspace{-0.5em}
\end{figure*}

\begin{figure*}[th]
\centering
\begin{subfigure}{.137\linewidth}
  \centering
  \includegraphics[width=.99\linewidth]{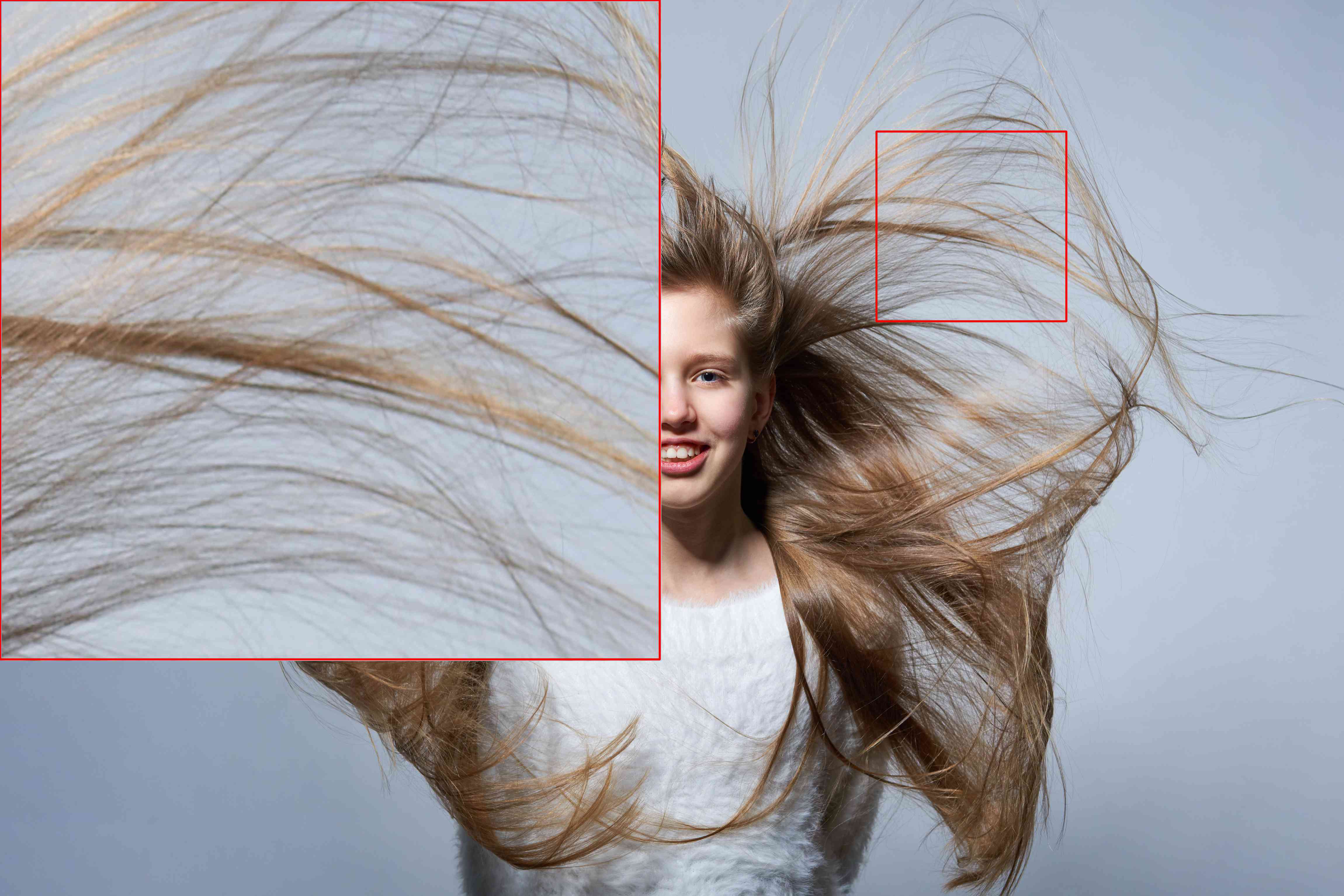} 
\end{subfigure}
\begin{subfigure}{.137\linewidth}
  \centering
  \includegraphics[width=.99\linewidth]{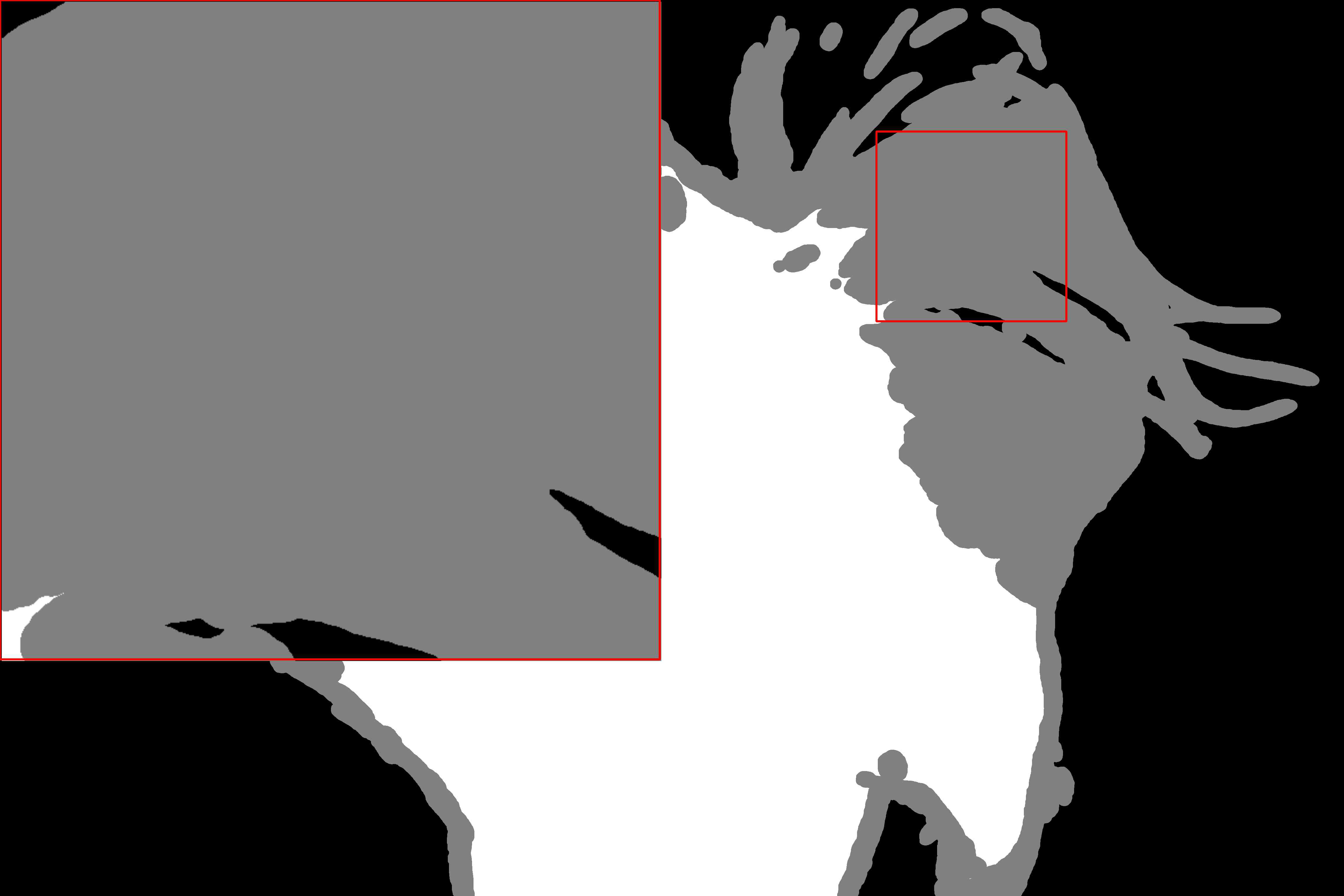}
\end{subfigure}
\begin{subfigure}{.137\linewidth}
  \centering
  \includegraphics[width=.99\linewidth]{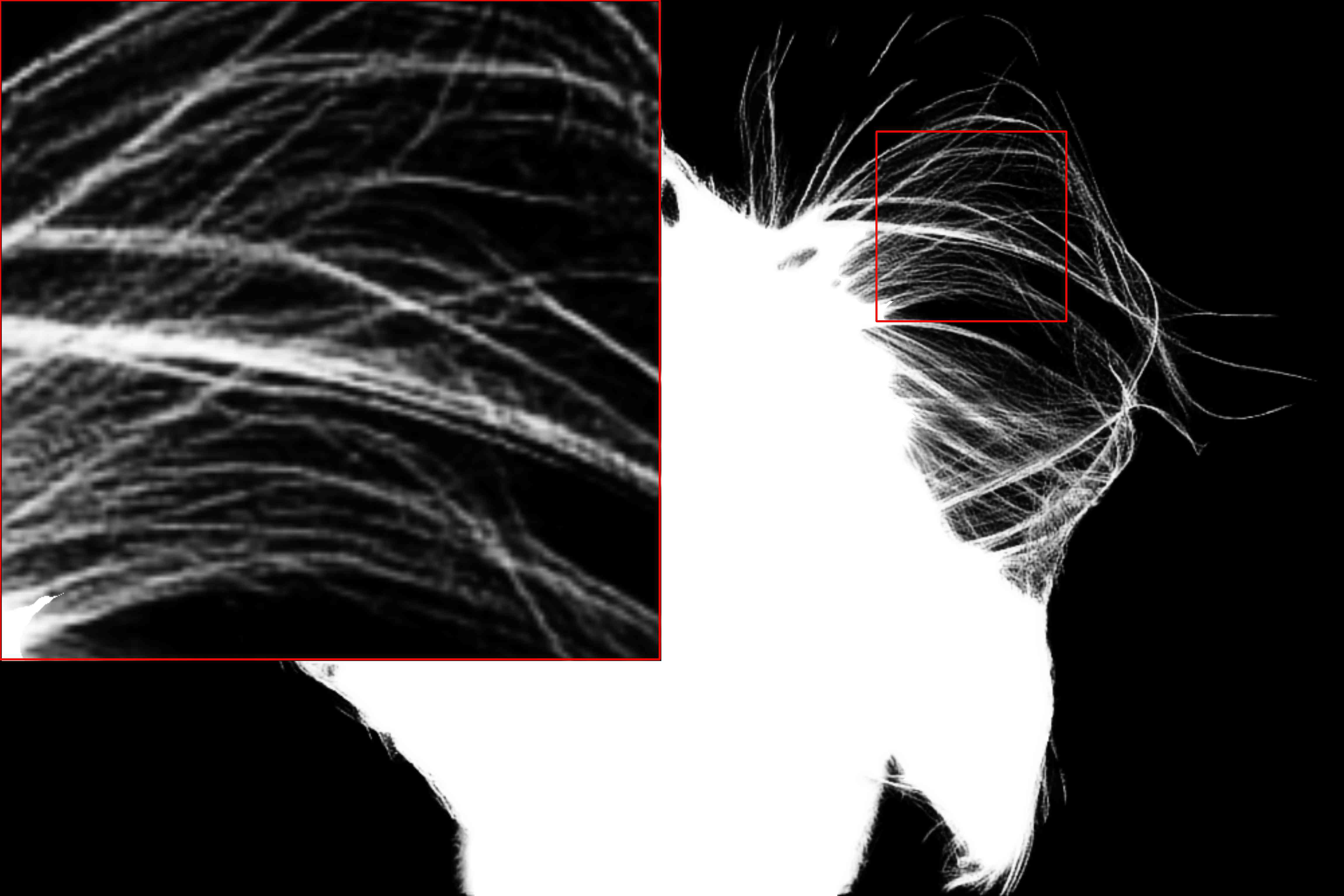}
\end{subfigure}
\begin{subfigure}{.137\linewidth}
  \centering
  \includegraphics[width=.99\linewidth]{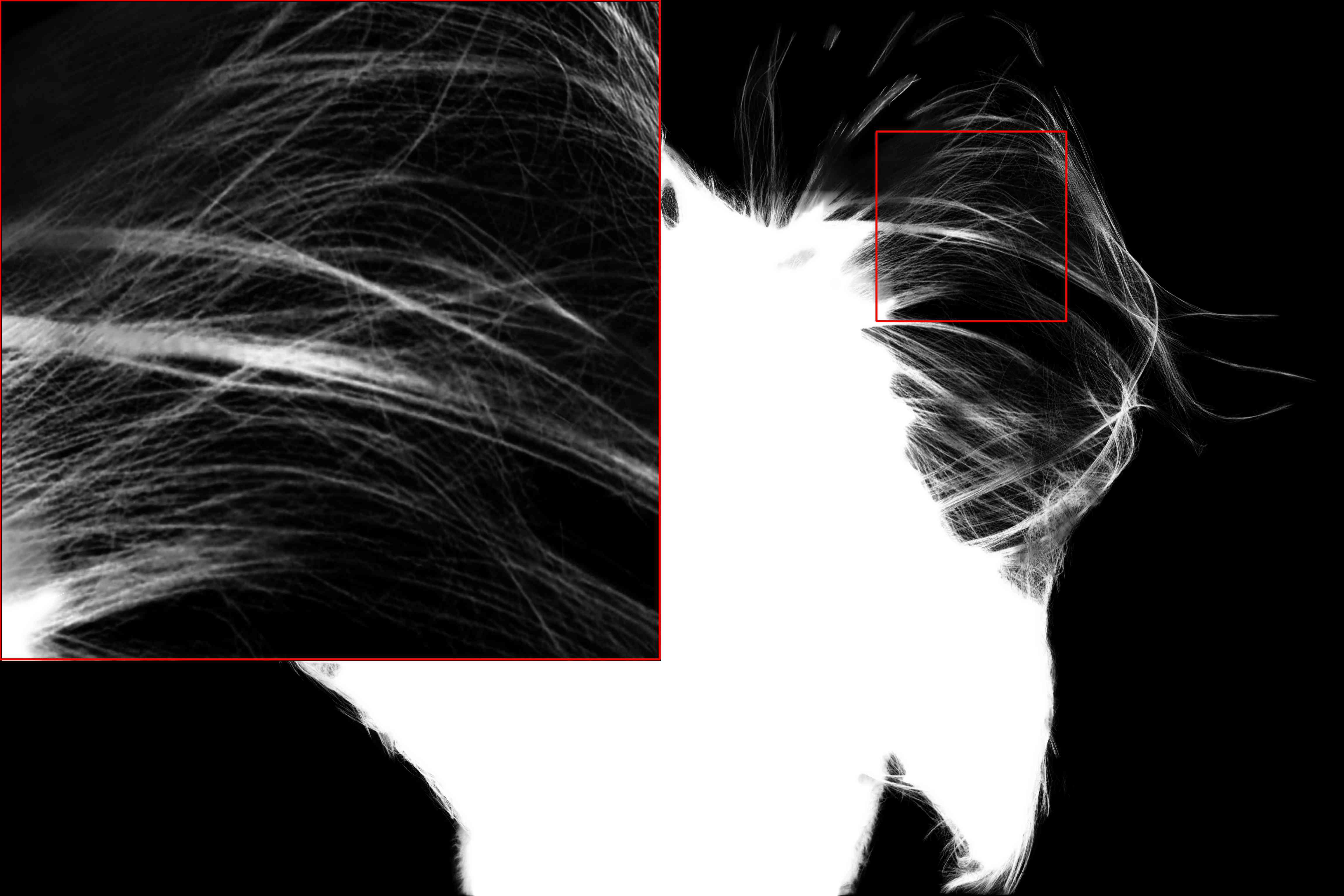}
\end{subfigure}
\begin{subfigure}{.137\linewidth}
  \centering
  \includegraphics[width=.99\linewidth]{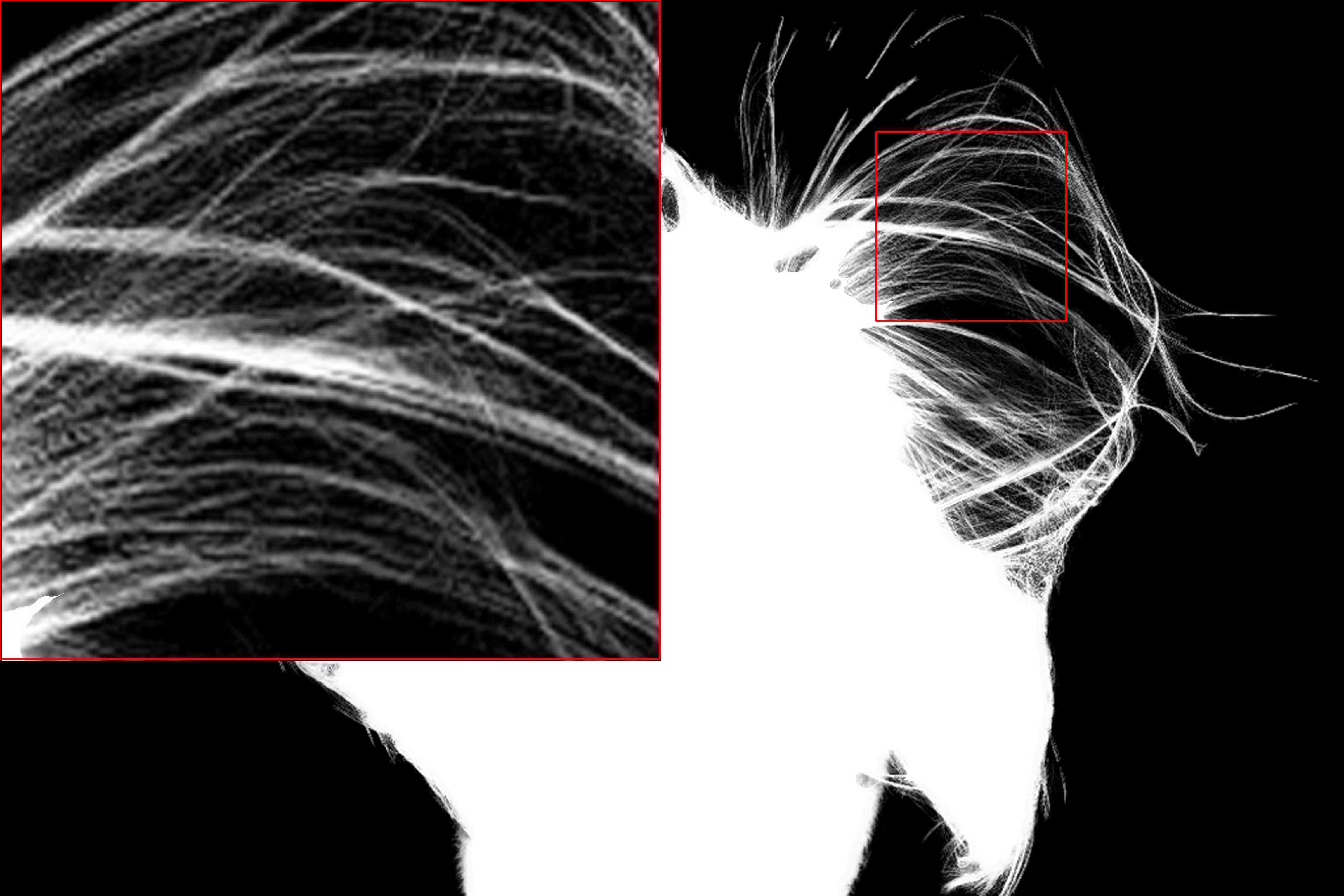}
\end{subfigure}
\begin{subfigure}{.137\linewidth}
  \centering
  \includegraphics[width=.99\linewidth]{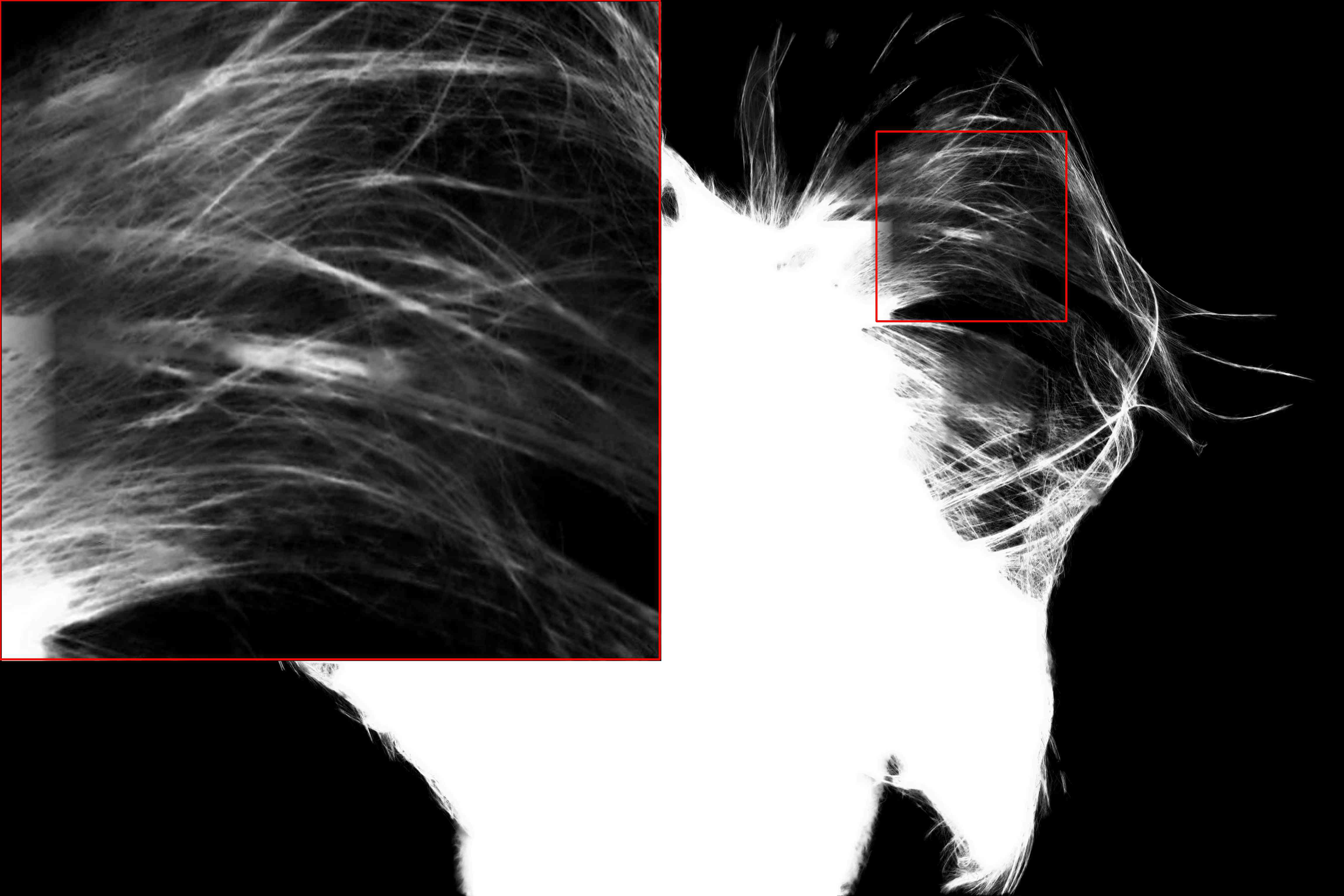}
\end{subfigure}
\begin{subfigure}{.137\linewidth}
  \centering
  \includegraphics[width=.99\linewidth]{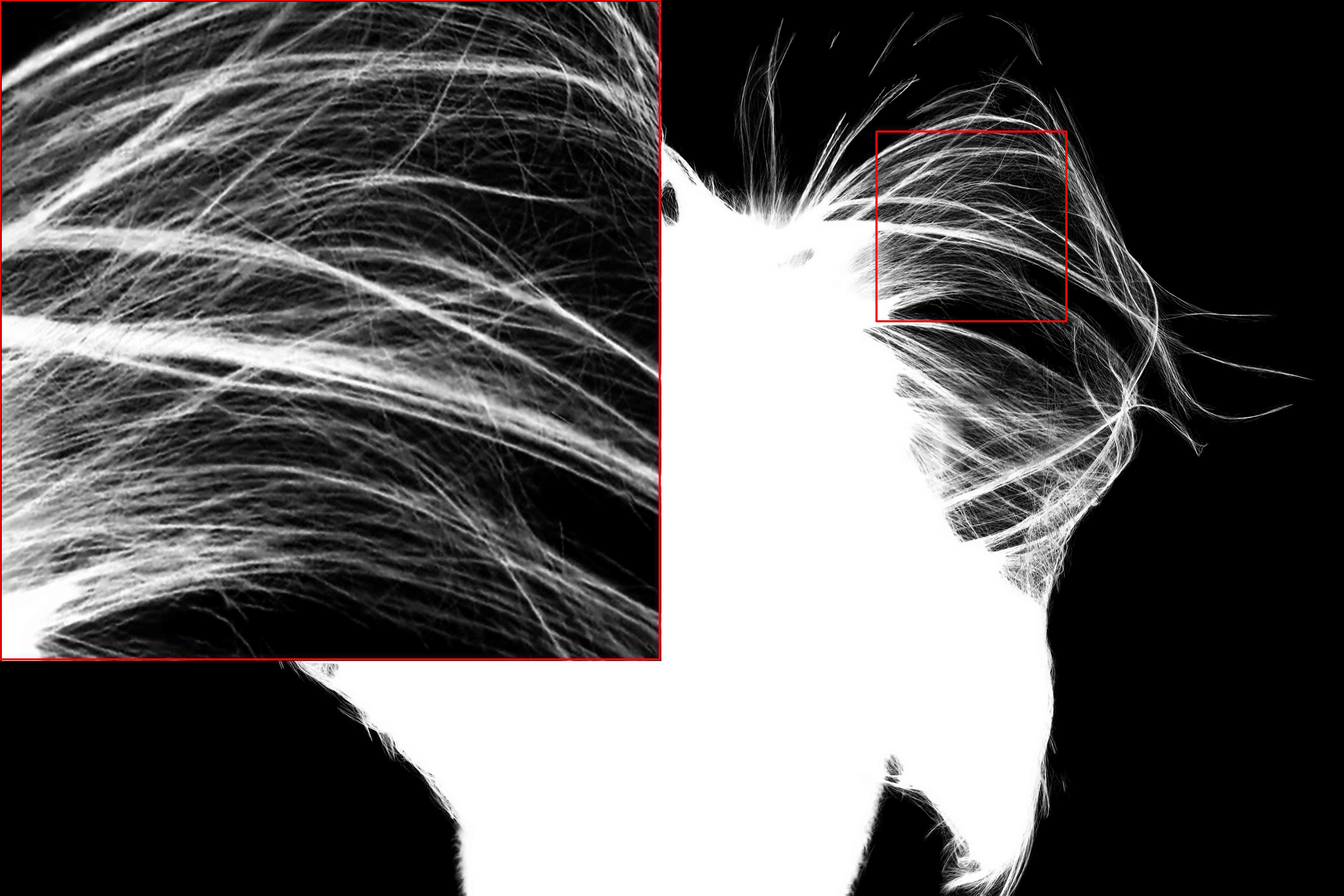}
\end{subfigure}
\begin{subfigure}{.137\linewidth}
  \centering
  \includegraphics[width=.99\linewidth]{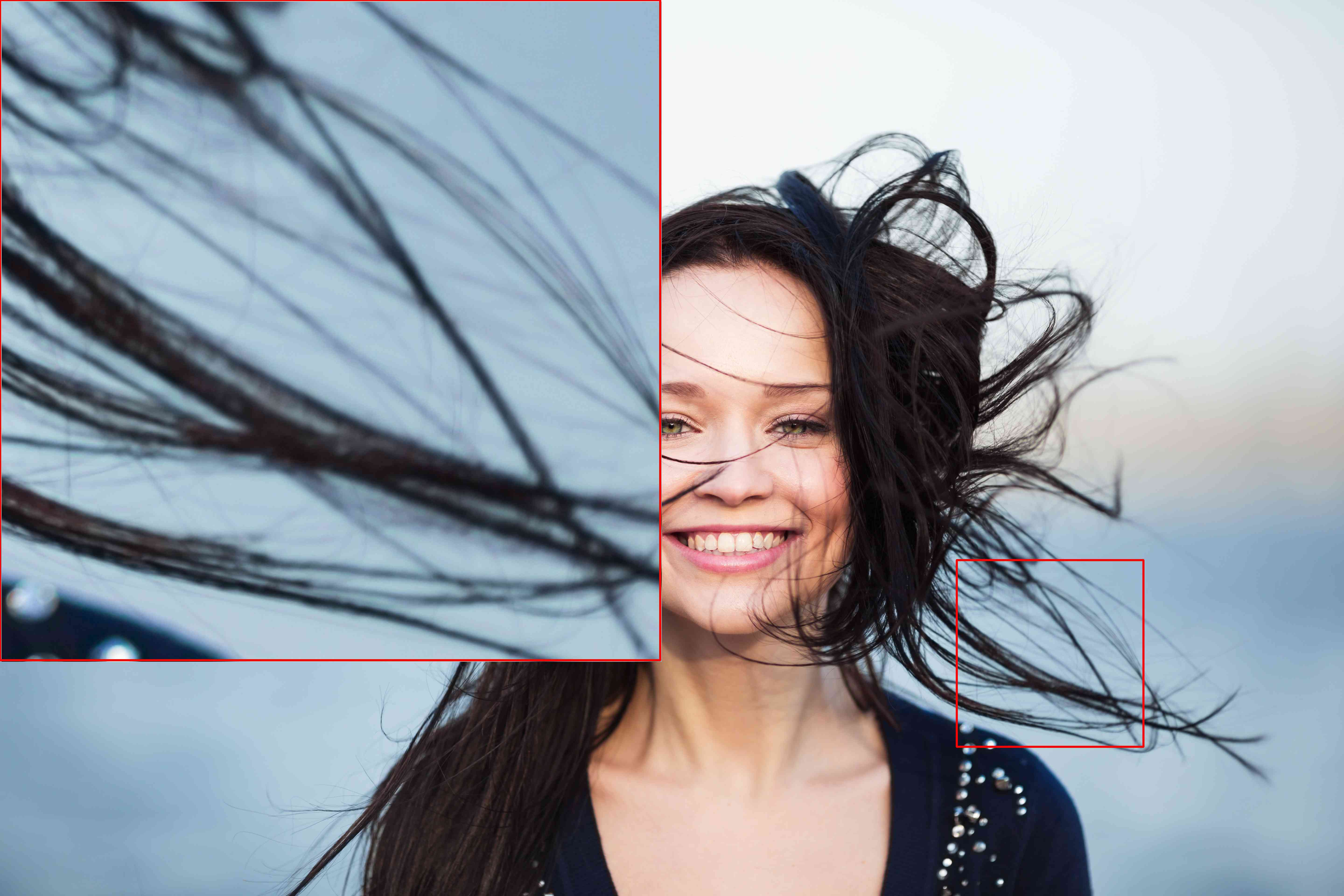} 
\end{subfigure}
\begin{subfigure}{.137\linewidth}
  \centering
  \includegraphics[width=.99\linewidth]{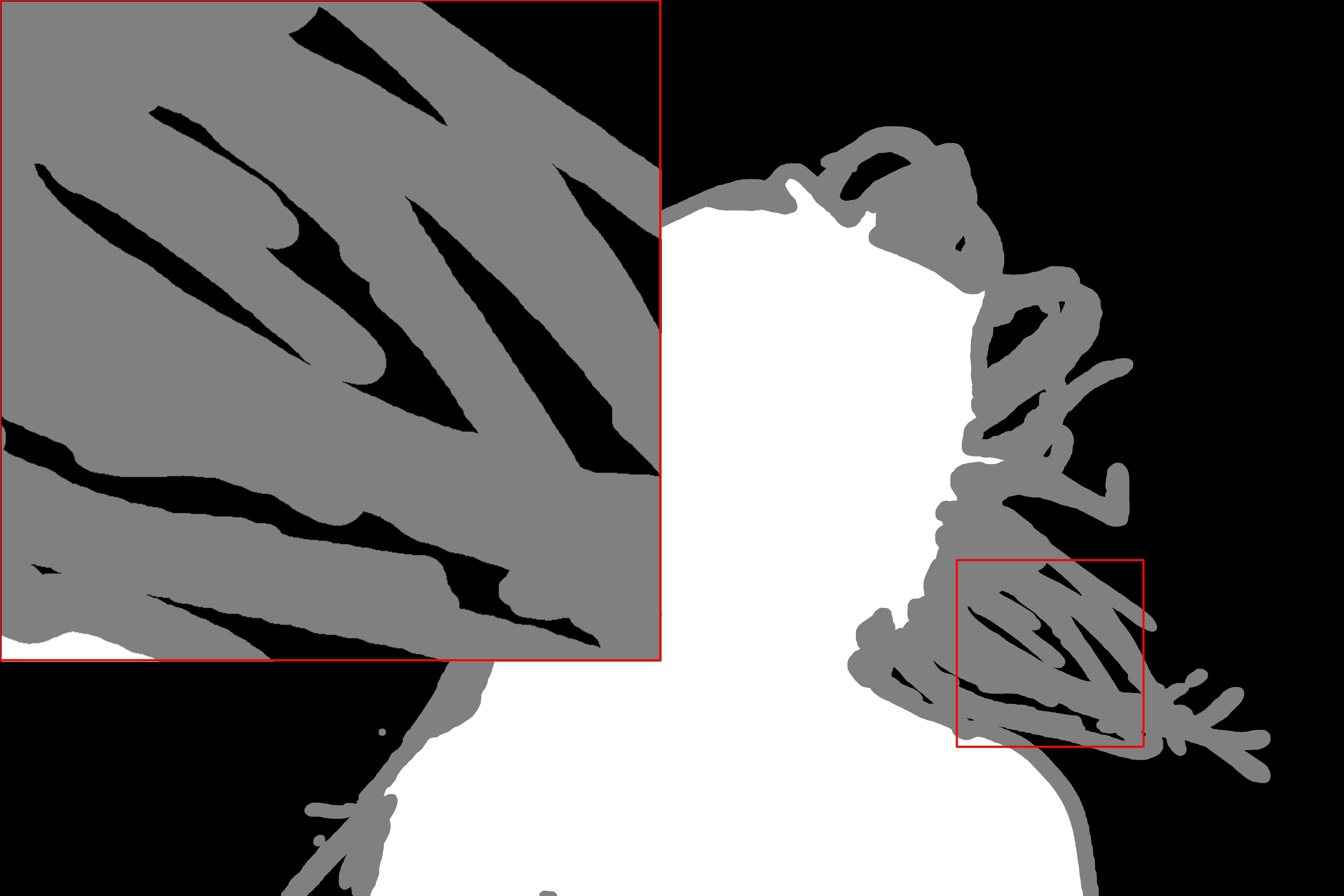}
\end{subfigure}
\begin{subfigure}{.137\linewidth}
  \centering
  \includegraphics[width=.99\linewidth]{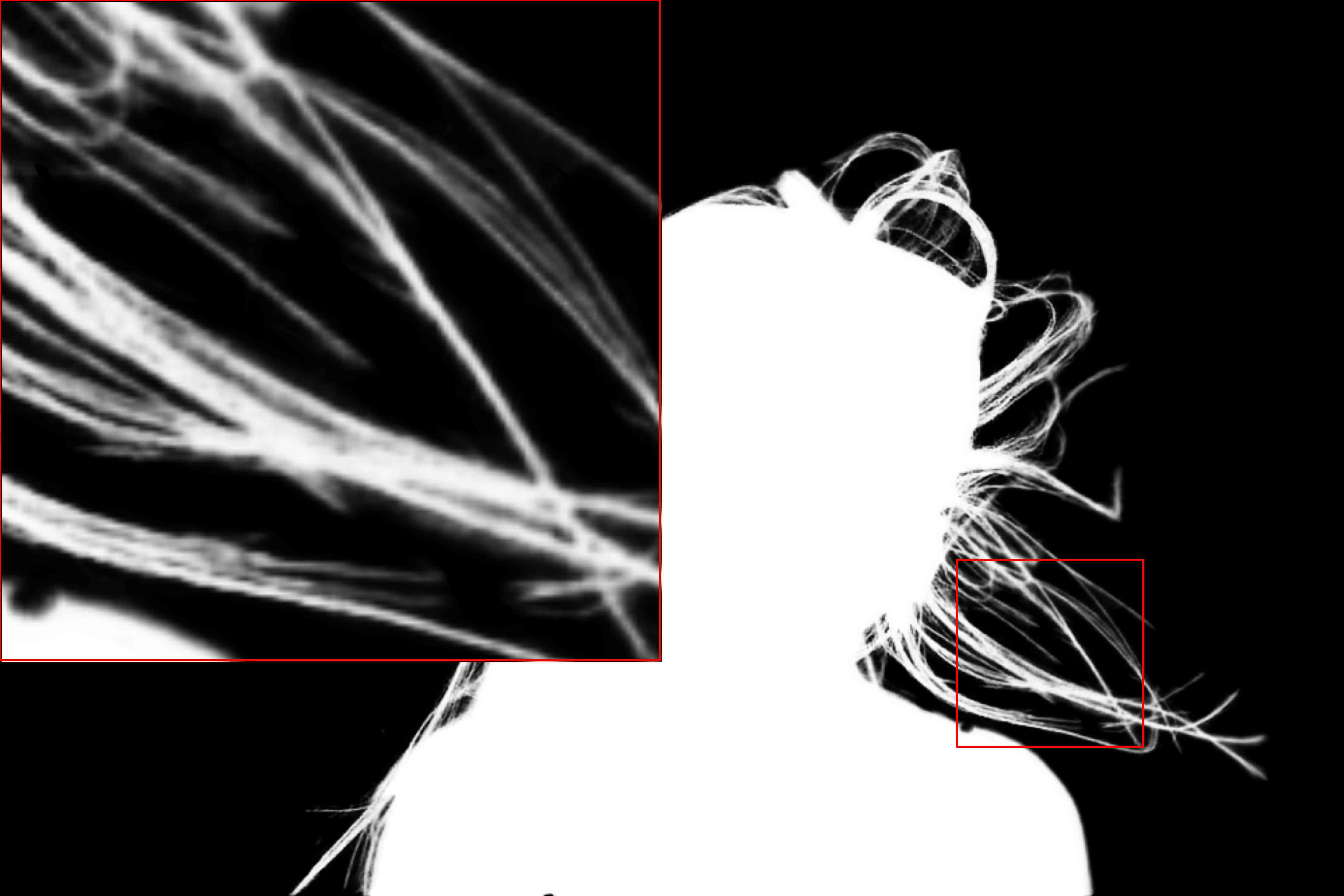}
\end{subfigure}
\begin{subfigure}{.137\linewidth}
  \centering
  \includegraphics[width=.99\linewidth]{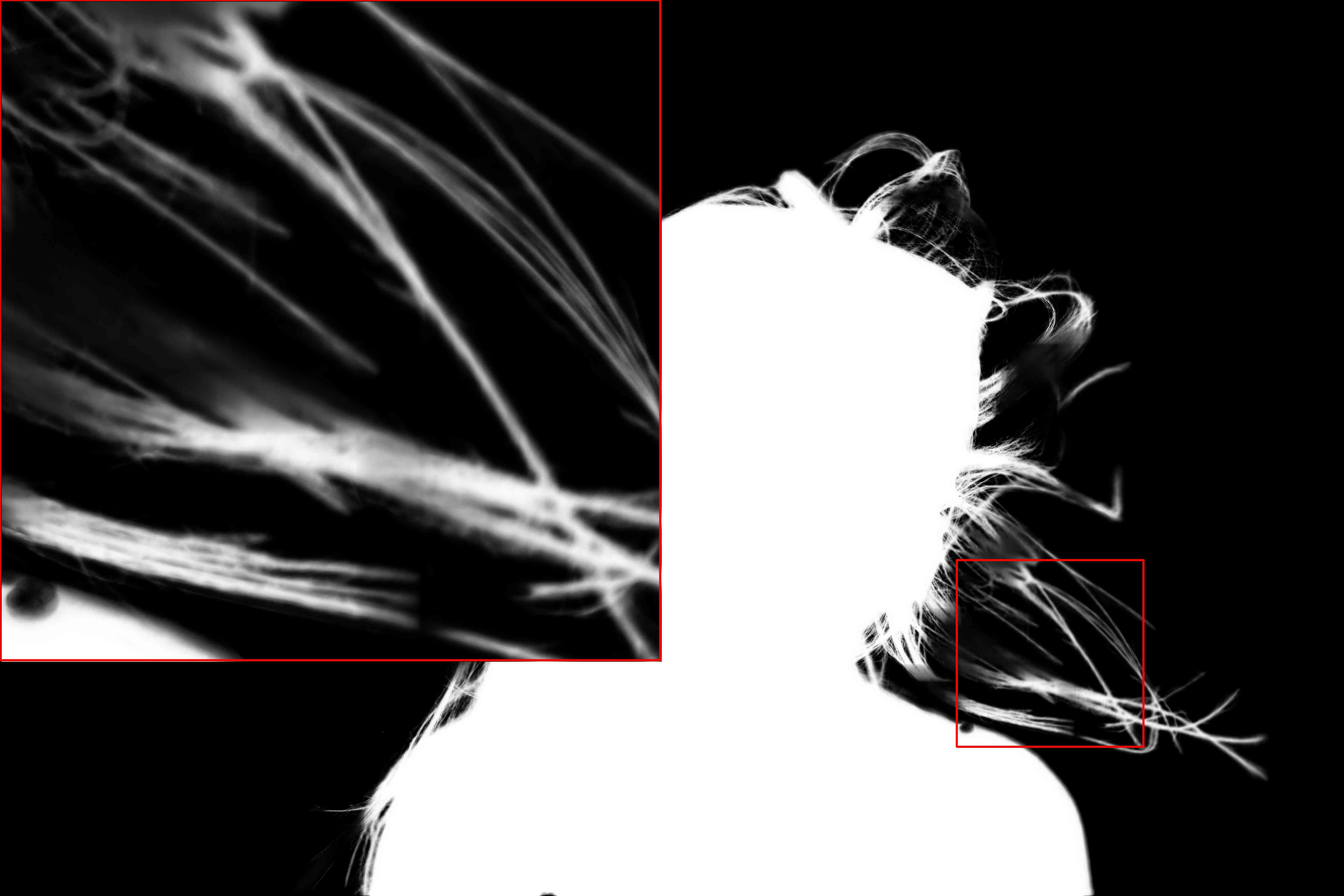}
\end{subfigure}
\begin{subfigure}{.137\linewidth}
  \centering
  \includegraphics[width=.99\linewidth]{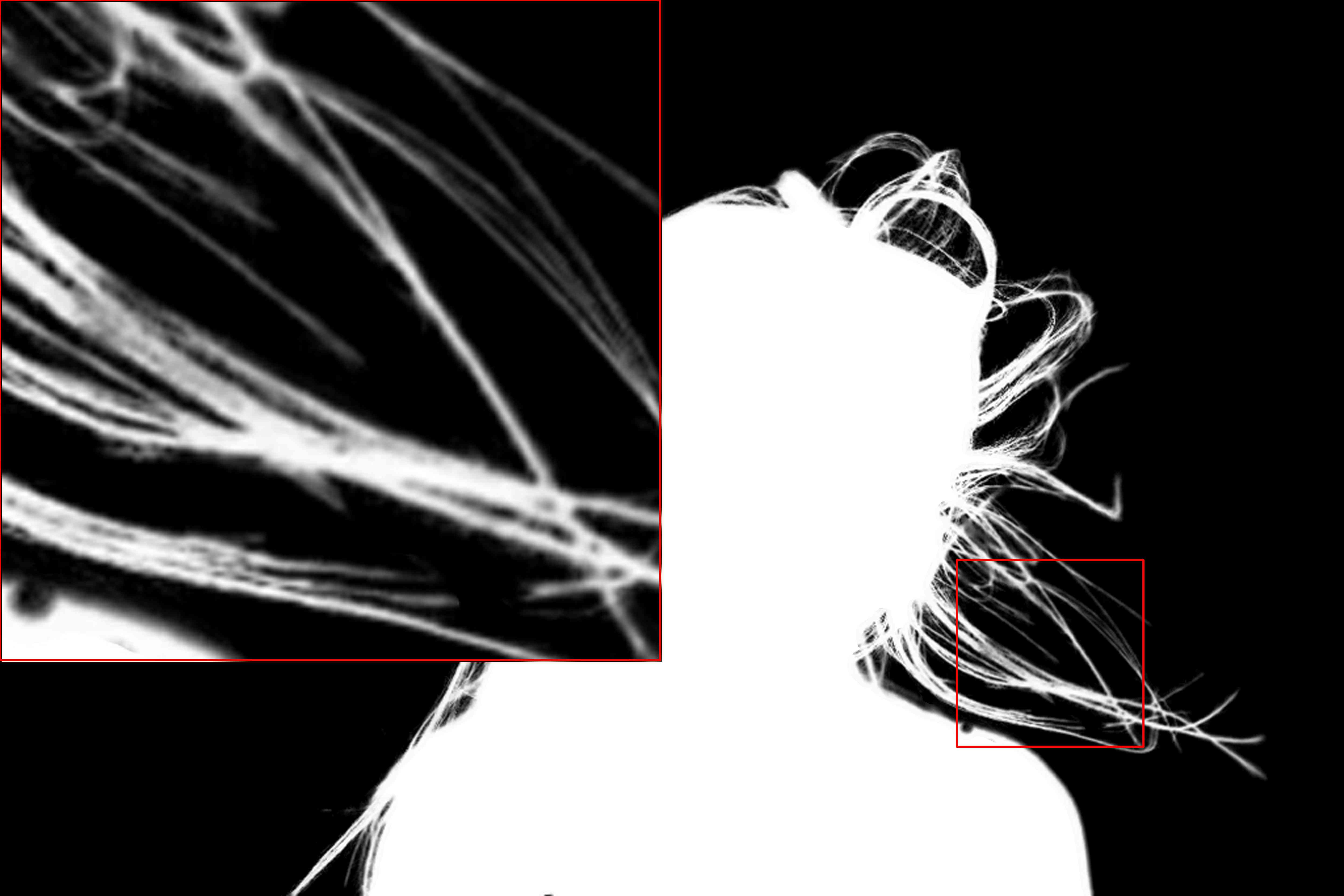}
\end{subfigure}
\begin{subfigure}{.137\linewidth}
  \centering
  \includegraphics[width=.99\linewidth]{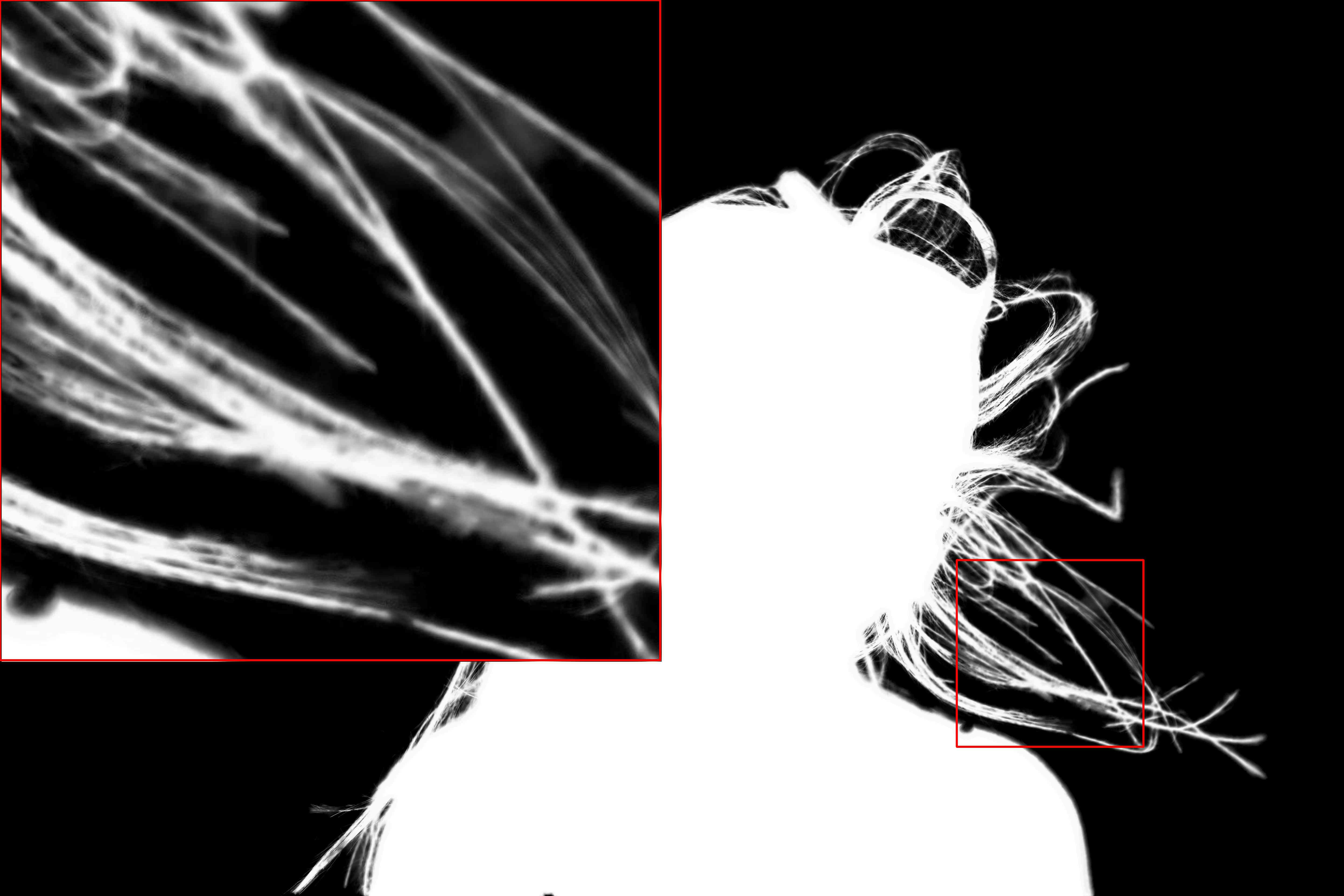}
\end{subfigure}
\begin{subfigure}{.137\linewidth}
  \centering
  \includegraphics[width=.99\linewidth]{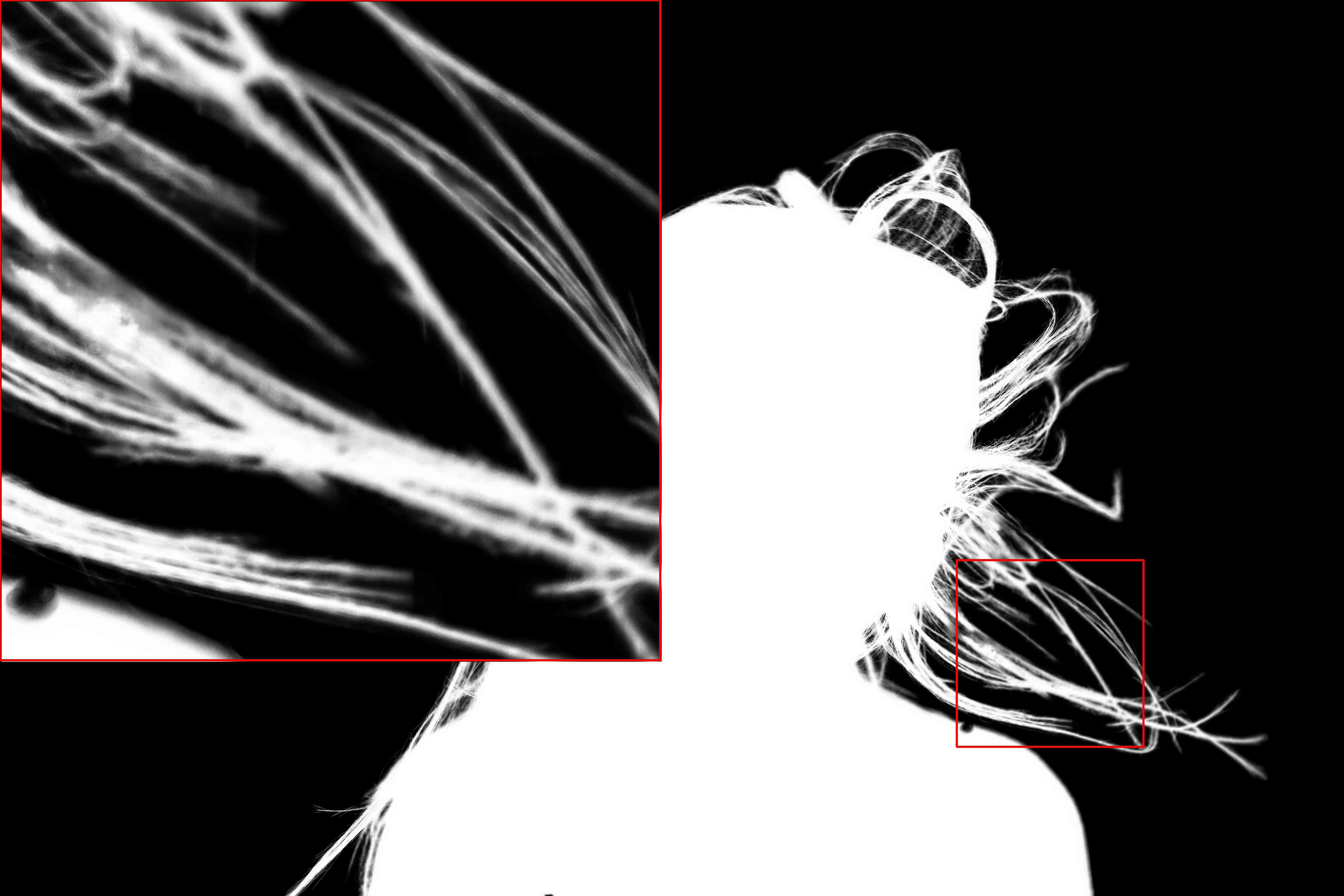}
\end{subfigure}
\begin{subfigure}{.137\linewidth}
  \centering
  \includegraphics[width=.99\linewidth]{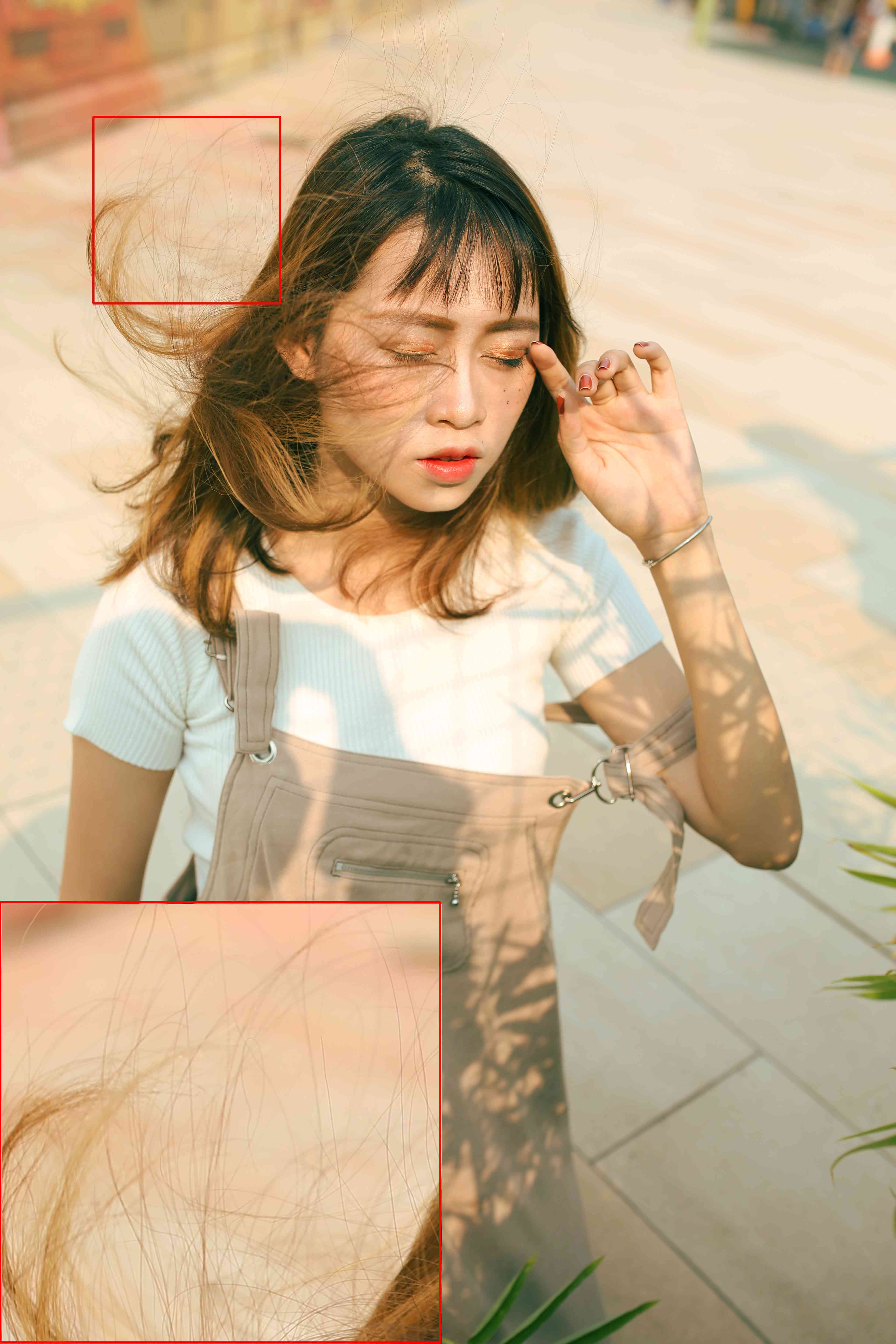} 
  \caption{HR Image}
\end{subfigure}
\begin{subfigure}{.137\linewidth}
  \centering
  \includegraphics[width=.99\linewidth]{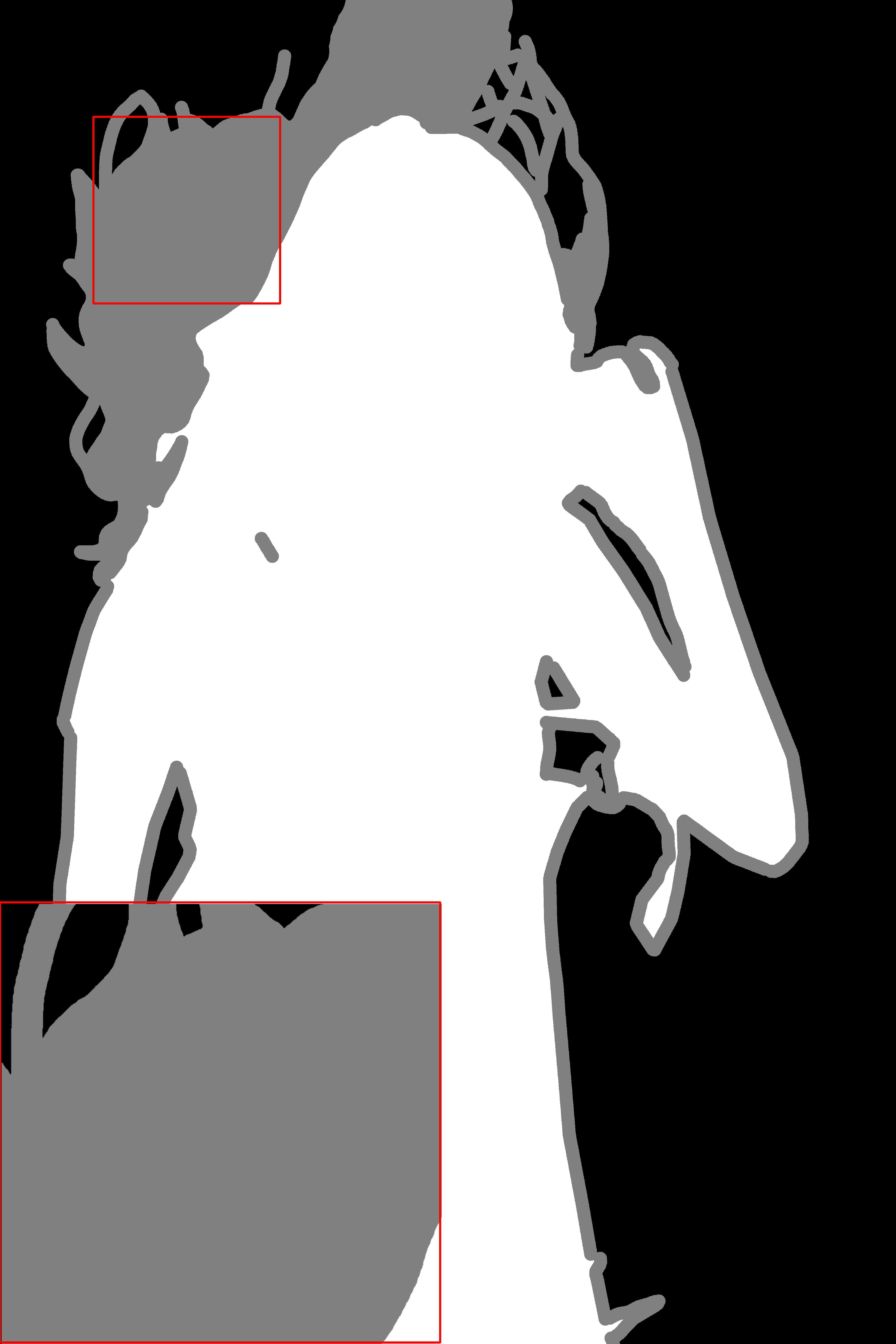}
  \caption{Trimap}
\end{subfigure}
\begin{subfigure}{.137\linewidth}
  \centering
  \includegraphics[width=.99\linewidth]{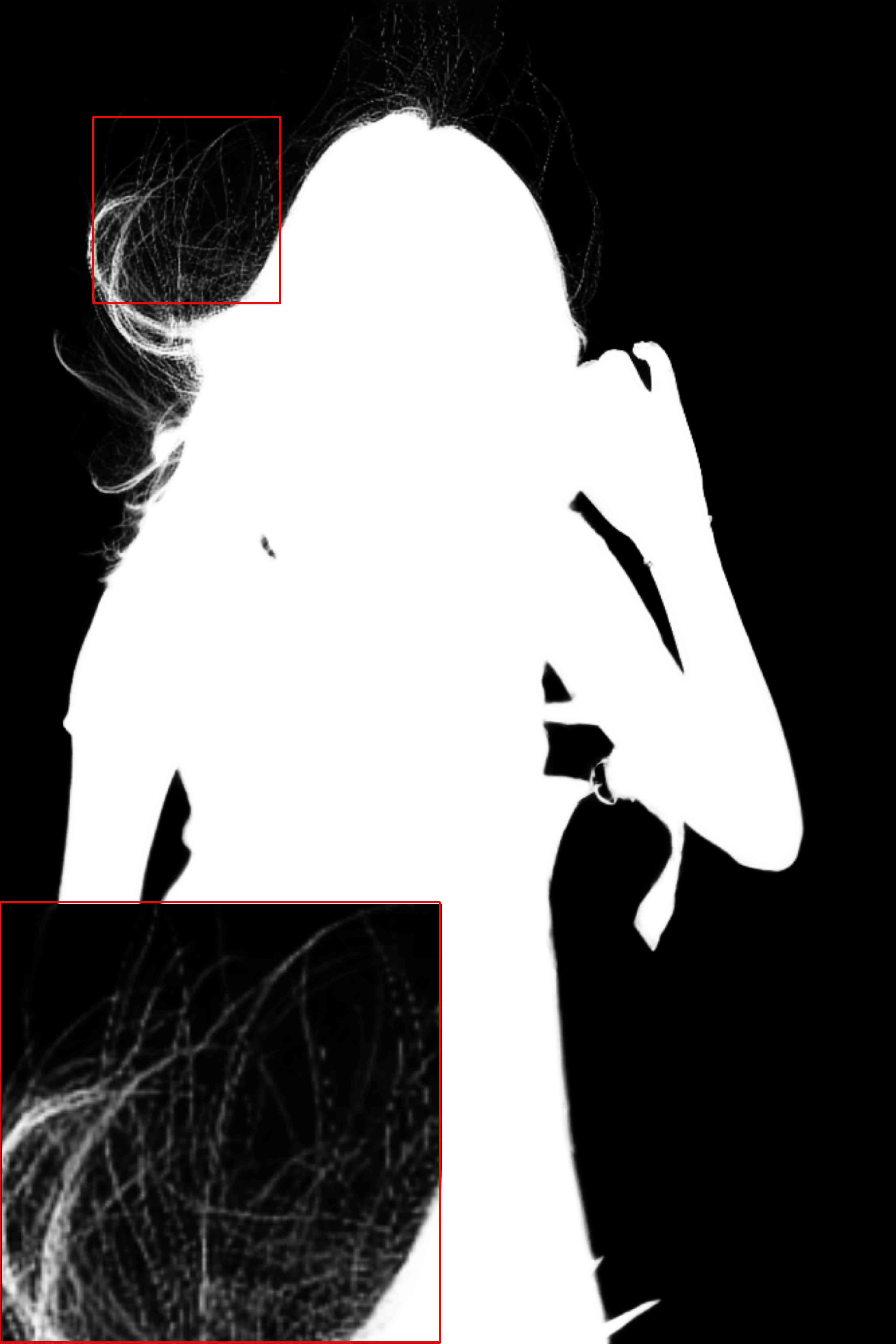}
  \caption{ContexNet-DS}
\end{subfigure}
\begin{subfigure}{.137\linewidth}
  \centering
  \includegraphics[width=.99\linewidth]{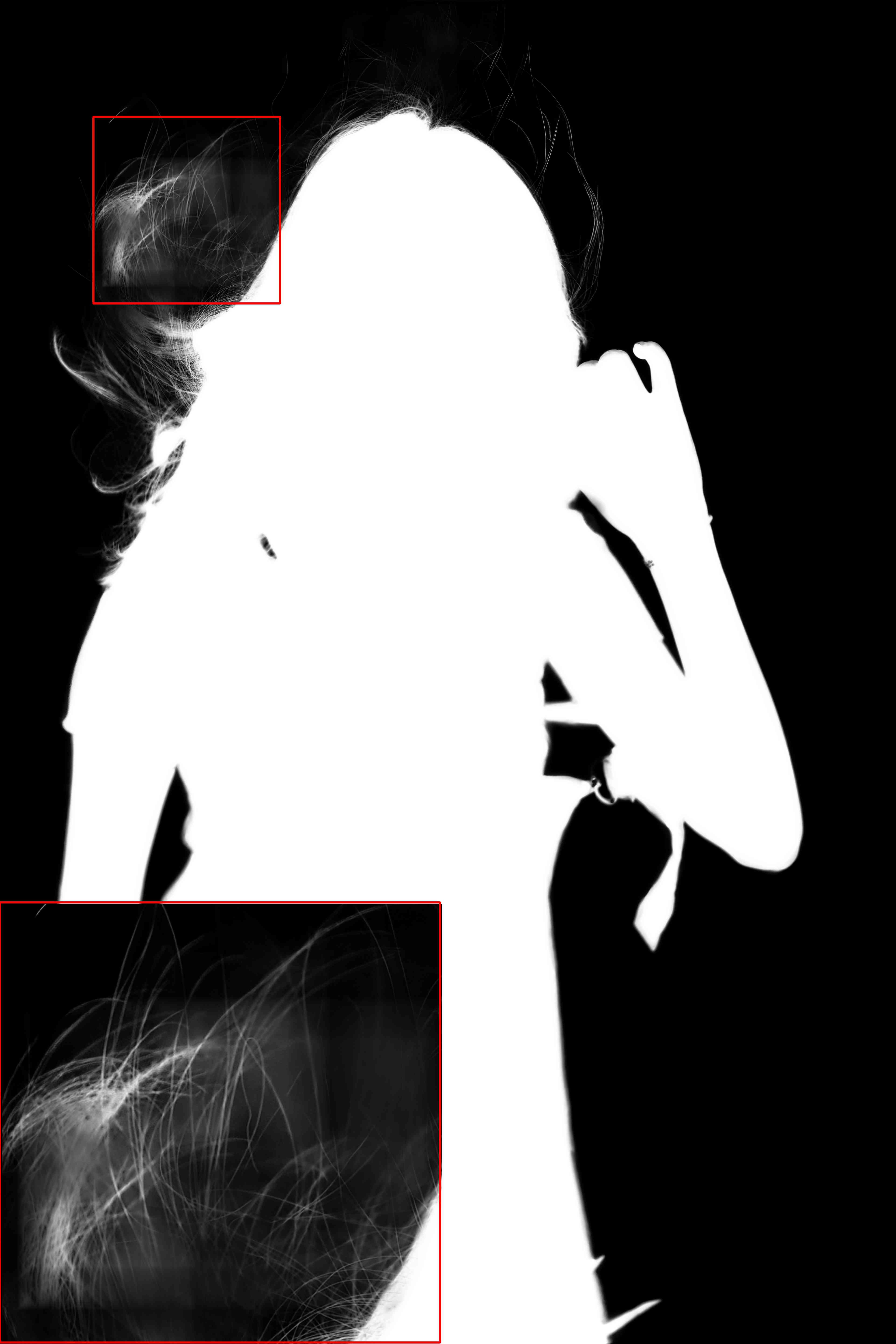}
  \caption{ContexNet-C}
\end{subfigure}
\begin{subfigure}{.137\linewidth}
  \centering
  \includegraphics[width=.99\linewidth]{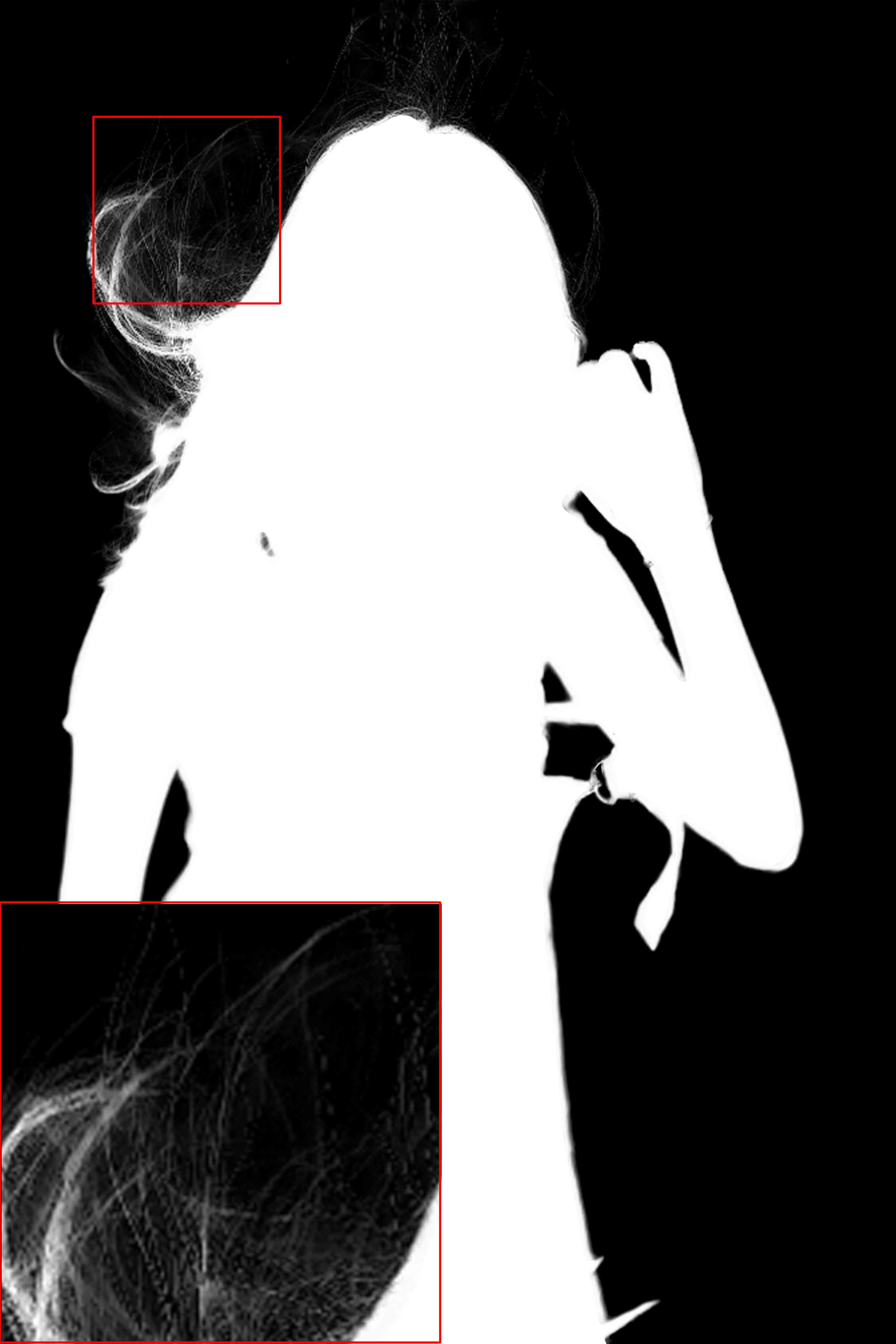}
  \caption{IndexNet-DS}
\end{subfigure}
\begin{subfigure}{.137\linewidth}
  \centering
  \includegraphics[width=.99\linewidth]{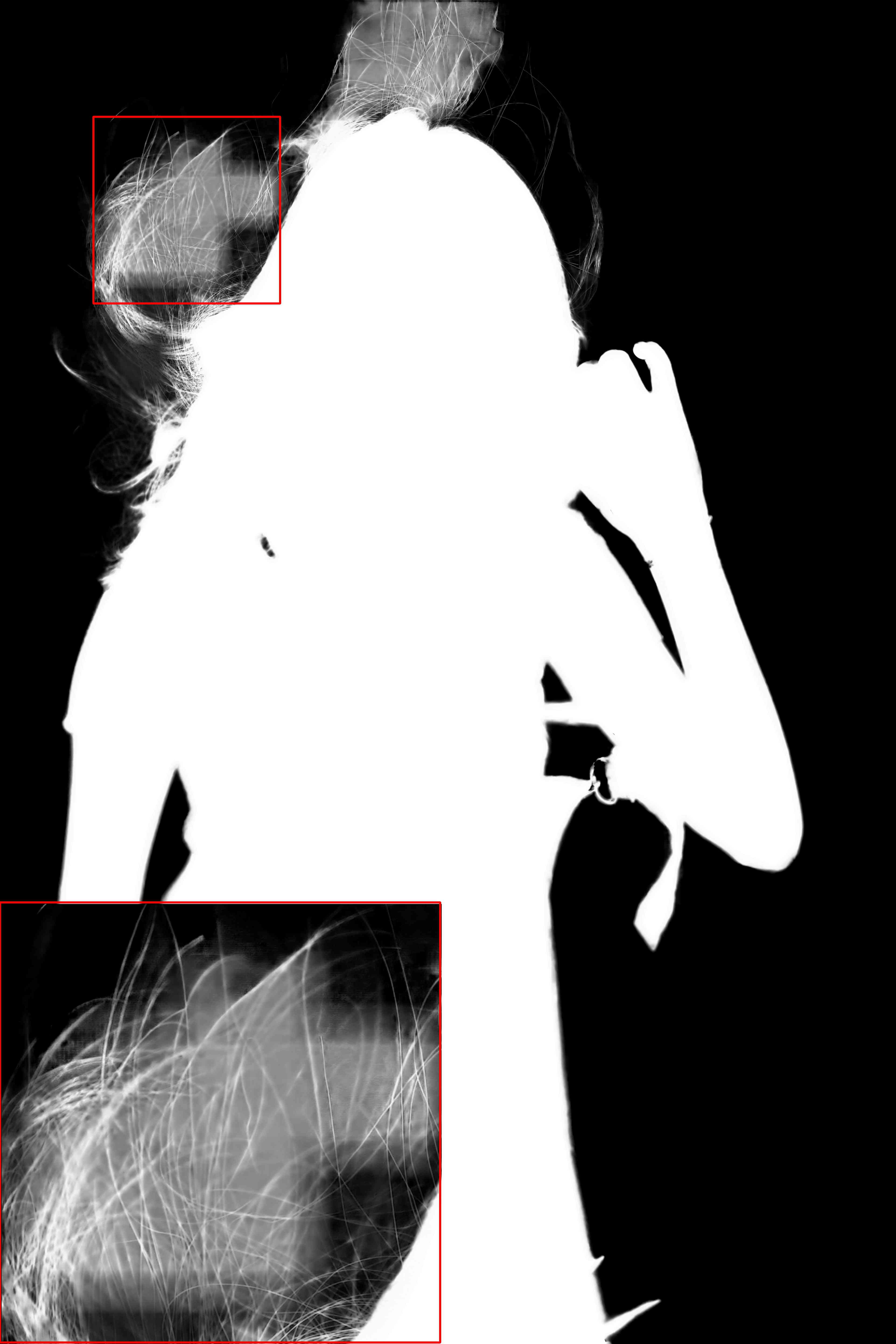}
  \caption{IndexNet-C}
\end{subfigure}
\begin{subfigure}{.137\linewidth}
  \centering
  \includegraphics[width=.99\linewidth]{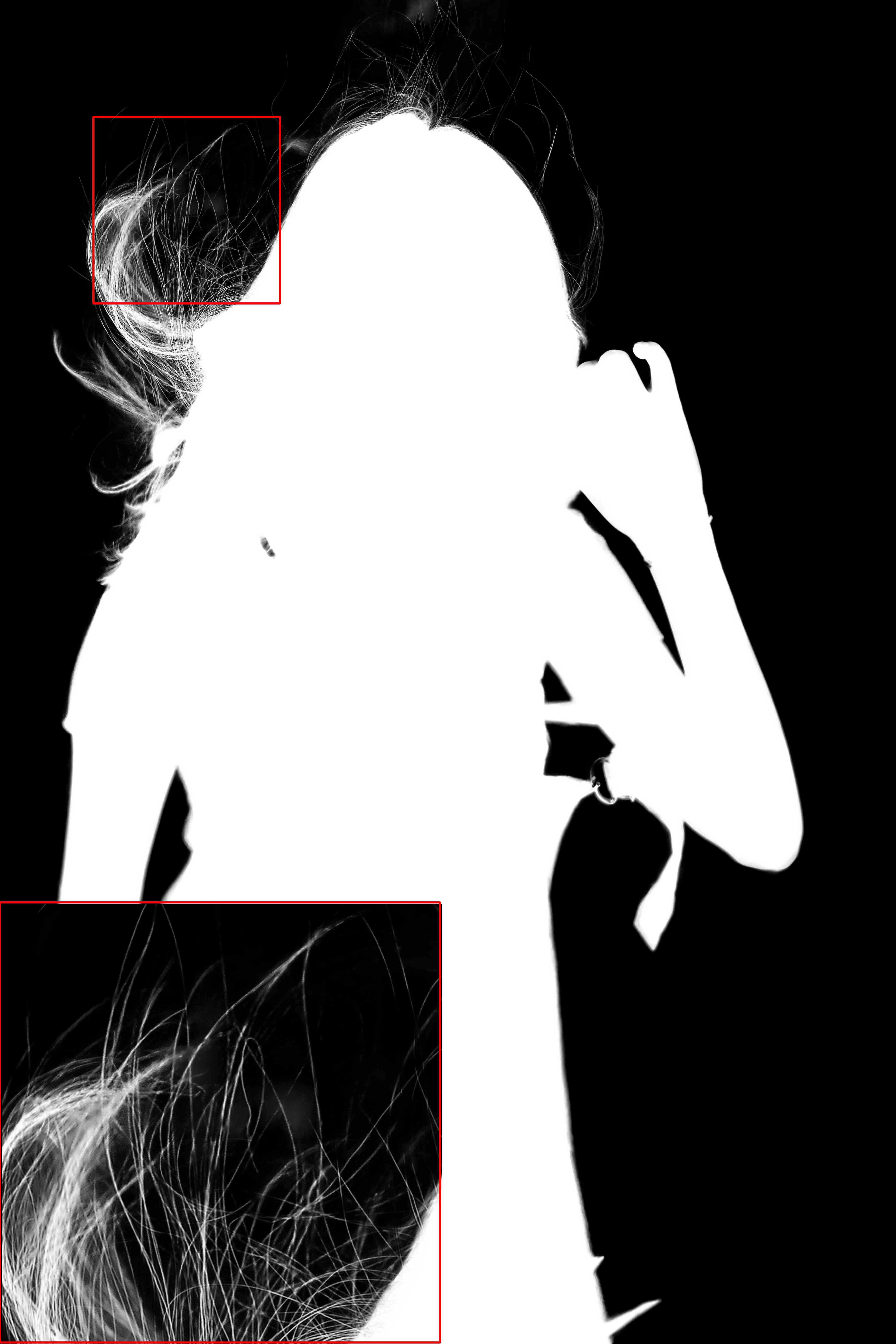}
  \caption{HDMatt (Ours)}
\end{subfigure}
\caption{Visual comparison on real-world HR images. Zoom in for details. Image sizes from top to bottom: $4601\times3069$, $5760\times3840$, $3840\times5760$. DS: Down-sampling. C: Patch-based cropping.}
\label{fig:real}
\vspace{-1em}
\end{figure*}

\subsection{Adobe Image Matting Benchmark}
We tested our methods on Adobe Image Matting testset. This dataset has $1000$ test images with alpha matte, which are synthesized from $50$ foreground images and $1000$ background images from Pascal~VOC~\cite{everingham2015pascal}. We use four evaluation metrics, SAD, MSE, Gradient and Connectivity as in \cite{rhemann2009perceptually}. Tab.~\ref{tab:aim} shows our results together with other state-of-the-arts. We achieve the best performance with other top-ranked methods in all the evaluation metrics. In addition to the results on whole images, we also test several methods in a naive patch-wise manner (Note that they use the same patch size and blending function as our method). It is clear that their results on patches are worse than those on whole images, indicating that a naive adaptation of previous methods under limited computation resources will have degraded performance.

In Fig.~\ref{fig:visual_sota}, we qualitatively compare with recent state-of-the-art matting methods including IndexNet~\cite{lu2019indices} and ContextNet~\cite{hou2019context}. For IndexNet, we used the official released code.
For ContextNet, we used the publicly available test results provided by the authors.
It demonstrates that our method works better especially in large unknown regions where little foreground or background information is available.
Both IndexNet and ContextNet take whole image as input. For each pixel,  they capture contextual dependency within a fixed receptive field formed by a stack of convolutional and pooling layers. Although this may capture local context, but it is not very effective to build a strong long-range contextual dependency.  In contrast, our method globally samples context patches that contain useful background and foreground information and explicitly correlates them with the given patch.

\subsection{AlphaMatting Benchmark}
AlphaMatting~\cite{rhemann2009perceptually} is a popular image matting benchmark while all of its test images are around 800$\times$600. Although our method is particularly effective for high-resolution images, our method still achieves the top-1 performance under SAD, MSE and Gradient metrics among all the published methods at the time of submission, which demonstrates that our method is general and effective for images of various resolutions. See Tab.~\ref{tab:alphamat} for details. See Appendix~\ref{app:alpha} for visual results.

\begin{table*}[th]
    \begin{center}
    \caption{
    Top-4 methods on AlphaMatting benchmark~\cite{rhemann2009perceptually}. Our method achieves best overall performance in SAD. Cases we get best results are in bold.}
    \scalebox{.61}{
    \begin{tabular}{l|C{0.4cm}|C{0.4cm}|C{0.4cm}|C{0.45cm}|C{0.45cm}C{0.45cm}C{0.45cm}|C{0.4cm}C{0.4cm}C{0.4cm}|C{0.4cm}C{0.4cm}C{0.4cm}|C{0.4cm}C{0.4cm}C{0.4cm}|C{0.4cm}C{0.4cm}C{0.4cm}|C{0.4cm}C{0.4cm}C{0.4cm}|C{0.45cm}C{0.45cm}C{0.45cm}|C{0.45cm}C{0.45cm}C{0.45cm}}
        \hline
        \multirow{2}{*}{} & \multicolumn{4}{c|}{Average} & \multicolumn{3}{c|}{Troll} & \multicolumn{3}{c|}{Doll} & \multicolumn{3}{c|}{Donkey} & \multicolumn{3}{c|}{Elephant} & \multicolumn{3}{c|}{Plant} & \multicolumn{3}{c|}{Pineapple} & \multicolumn{3}{c|}{Plastic bag} & \multicolumn{3}{c}{Net} \\
        \cline{2-29}
        & All & S & L & U &     S & L & U &    S & L & U &    S & L & U &    S & L & U &    S & L & U &    S & L & U &    S & L & U &    S & L & U \\
        \hline
        \textbf{HDMatt (Ours)} & \textbf{5} & 6.3 & \textbf{3.9}&	\textbf{5}	&9.5 &	10 &	10.7 &	4.7 &	\textbf{4.8} &	5.8 &	2.9 &	3 &	\textbf{2.6} &	1.1 &	1.2 &	\textbf{1.3} &	5.2 &	\textbf{5.9} &	\textbf{6.7} &	2.4 &	\textbf{2.6} &	\textbf{3.1} &	\textbf{17.3} &	\textbf{17.3} &	17 &	21.5 &	22.4 &	23.2  \\
        AdaMatting~\cite{cai2019disentangled} &	6.9	&5.9	&6	&8.9	&10.2 &	11.1 &	10.8 &	4.9 &	5.4 &	6.6 &	3.6 &	3.4 &	3.4 &	0.9 &	0.9 &	1.8 &	4.7 &	6.8 &	9.3 &	2.2 &	2.6 &	3.3 &	19.2 &	19.8 &	18.7 &	17.8 &	19.1 &	18.6 \\
        SampleNet~\cite{tang2019learning} &7.3&	5.4&	6.9&	9.8	&9.1&	9.7 &	9.8 &	4.3 &	4.8 &	5.1 &	3.4 &	3.7 &	3.2 &	0.9 &	1.1 &	2 &	5.1 &	6.8 &	9.7 &	2.5 &	4 &	3.7 &	18.6 &	19.3 &	19.1 &	20 &	21.6 &	23.2 \\
        GCAMatting~\cite{li2020natural}	&8.4&	9	&5.8	&10.4	&8.8 &	9.5 &	11.1 &	4.9 &	4.8 &	5.8 &	3.4 &	3.7 &	3.2 &	1.1 &	1.2 &	1.3 &	5.7 &	6.9 &	7.6 &	2.8 &	3.1 &	4.5 &	18.3 &	19.2 &	18.5 &	20.8 &	21.7 &	24.7 \\
        \hline
    \end{tabular}\label{tab:alphamat}
    }
    \end{center}
    \vspace{-1.5em}
\end{table*}

\subsection{Real-world Images}
Although our method achieves the state-of-the-art results on existing benchmarks, the advantages of our method are not fully reflected given that the test images of existing benchmarks are not very high resolution. Therefore we collect dozens of online HR images with resolution up to $6000\times6000$. 
In Fig.~\ref{fig:real1}, we test IndexNet~\cite{lu2019indices} and ContextNet~\cite{hou2019context} with the whole image as input. Since these images are too large to be fed into a single GPU, we use CPU instead, which has a prohibited long inference time for each test image. From the results we can see that our method extracts finer and more accurate details than the other two state-of-the-art matting methods, while having a much faster inference speed. We also notice that our method also misses some finest details. A possible explanation for this issue is that the AIM training set lacks similar training examples with such small details.

In Fig.~\ref{fig:real}, we test the previous matting methods with more realistic settings. The first setting is to run inference on the downsampled images (\ie~1024$\times$1024) and then the predicted results are upsampled to the original resolution. The second setting is to run both methods in a crop-and-stitch manner with the same patch size (\ie~320$\times$320) and smooth blending function as our method. From the results of the prior works, we can clearly see that the downsampling strategy will lose a lot of details and produce blurry results while the naive-patch strategy will cause inconsistent results across patches due to the lack of cross-patch and long-range information. In contrast, our method is able to produce high-quality alpha mattes on the high-resolution images.

\begin{figure}[t]
    \centering
    \includegraphics[width=0.32\linewidth]{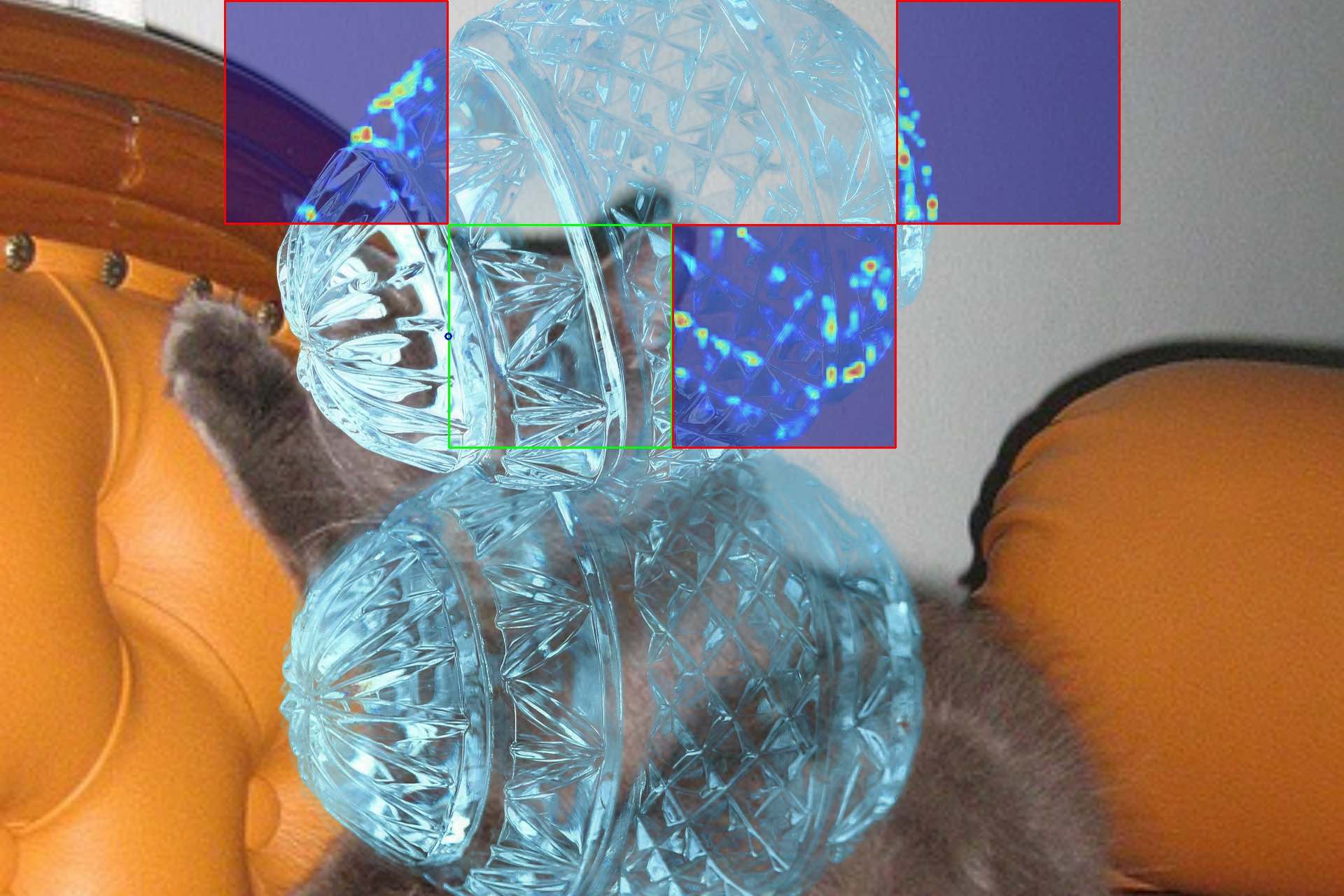}
    \includegraphics[width=0.32\linewidth]{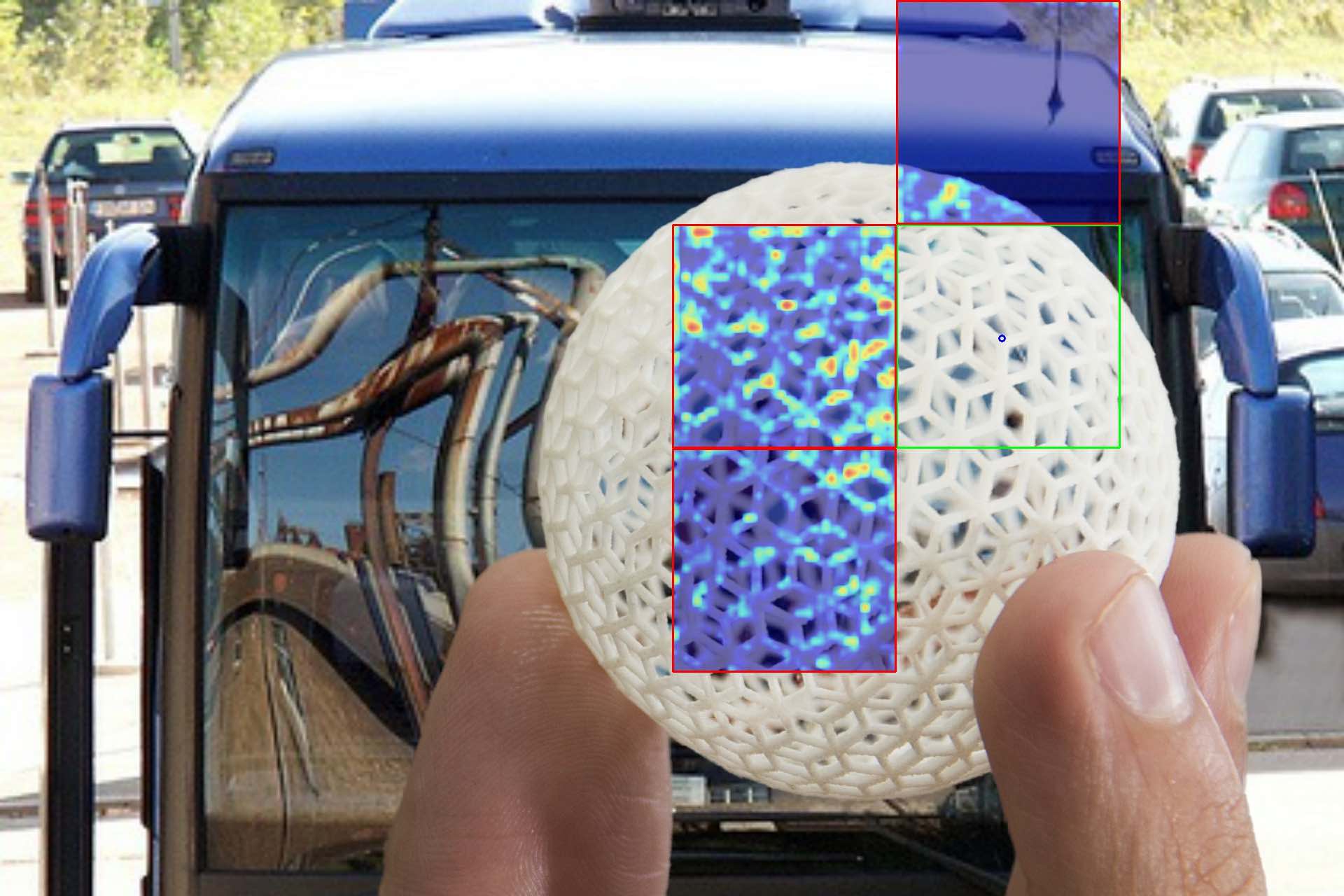}
    \includegraphics[width=0.32\linewidth]{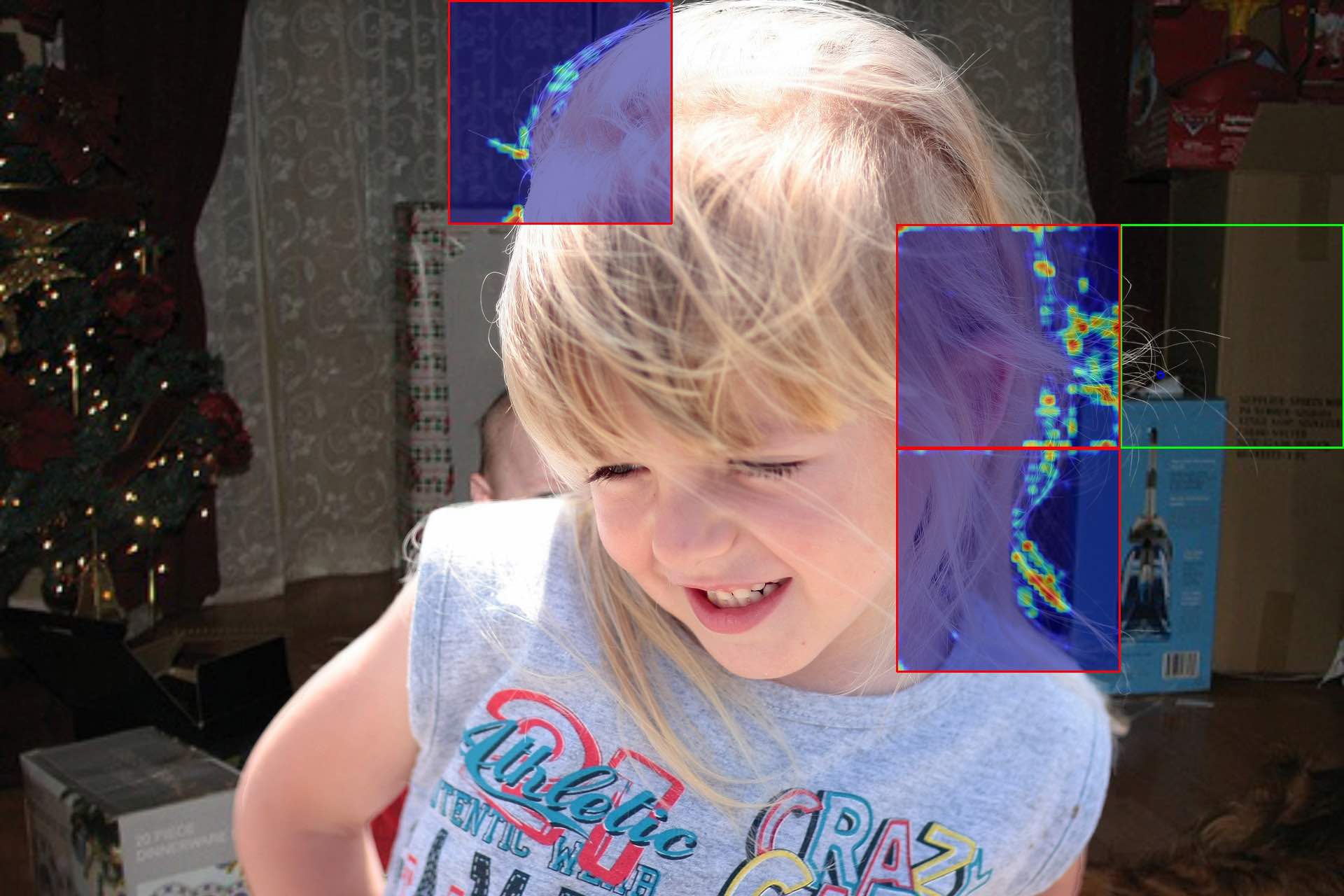}
    \includegraphics[width=0.32\linewidth]{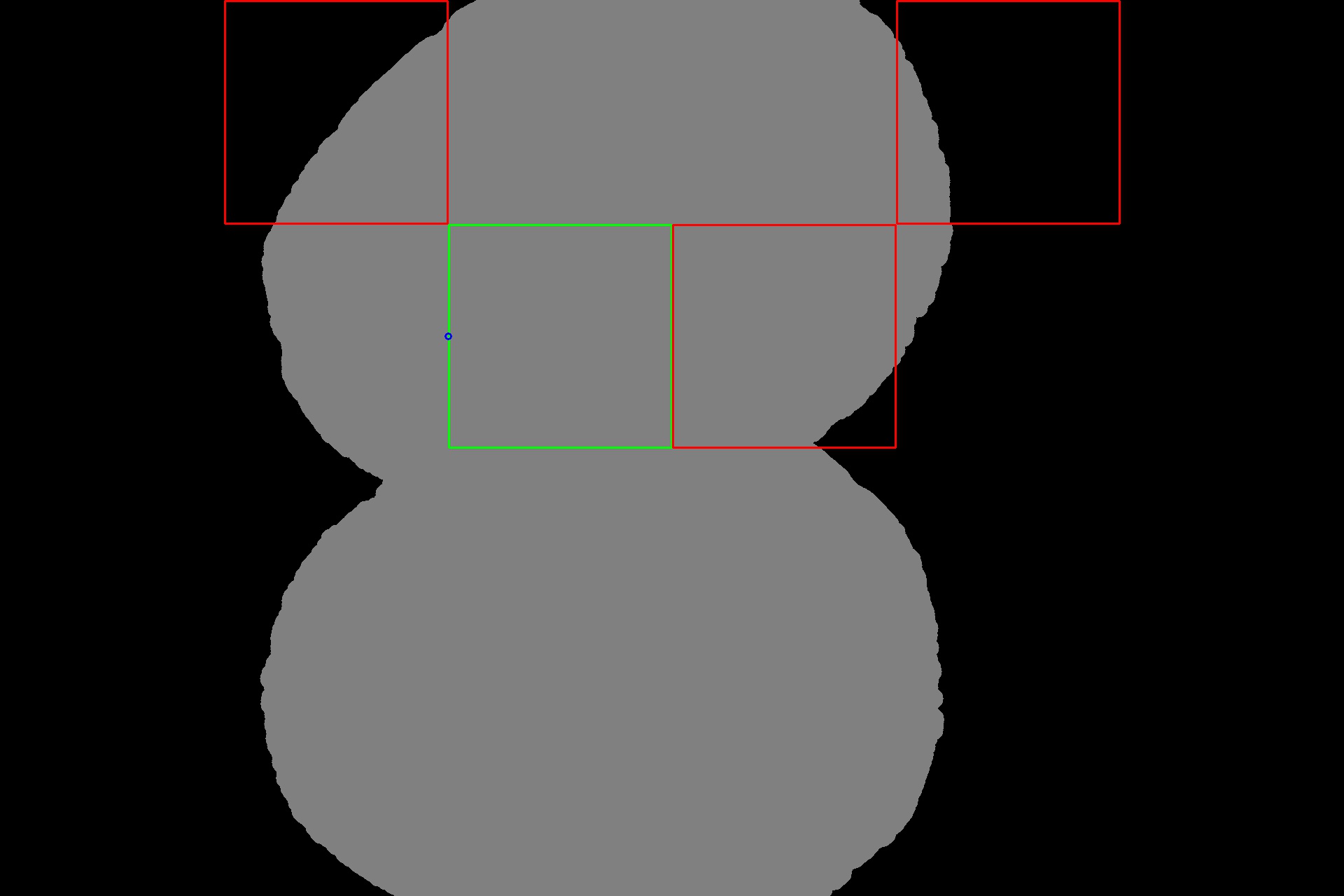}
    \includegraphics[width=0.32\linewidth]{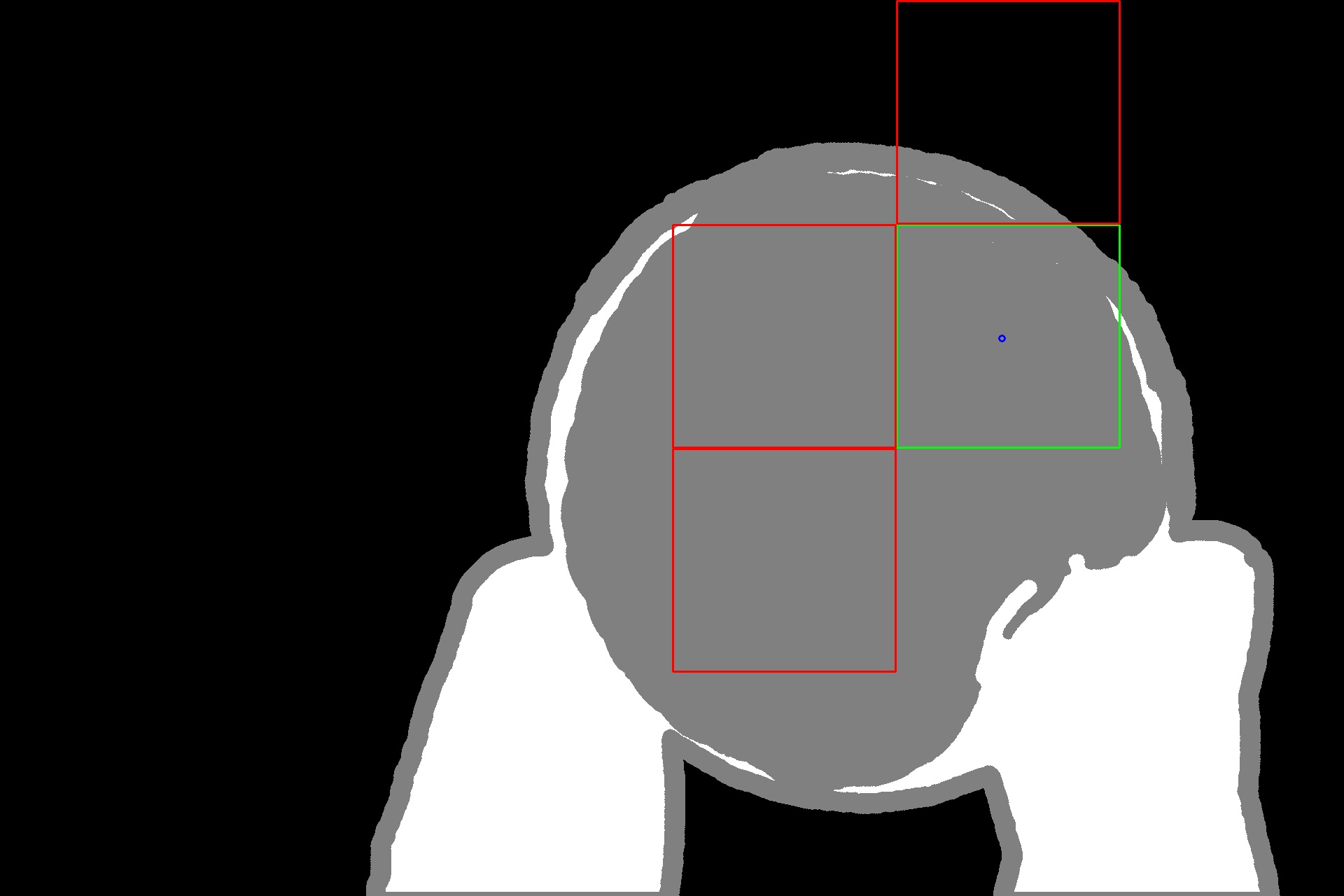}
    \includegraphics[width=0.32\linewidth]{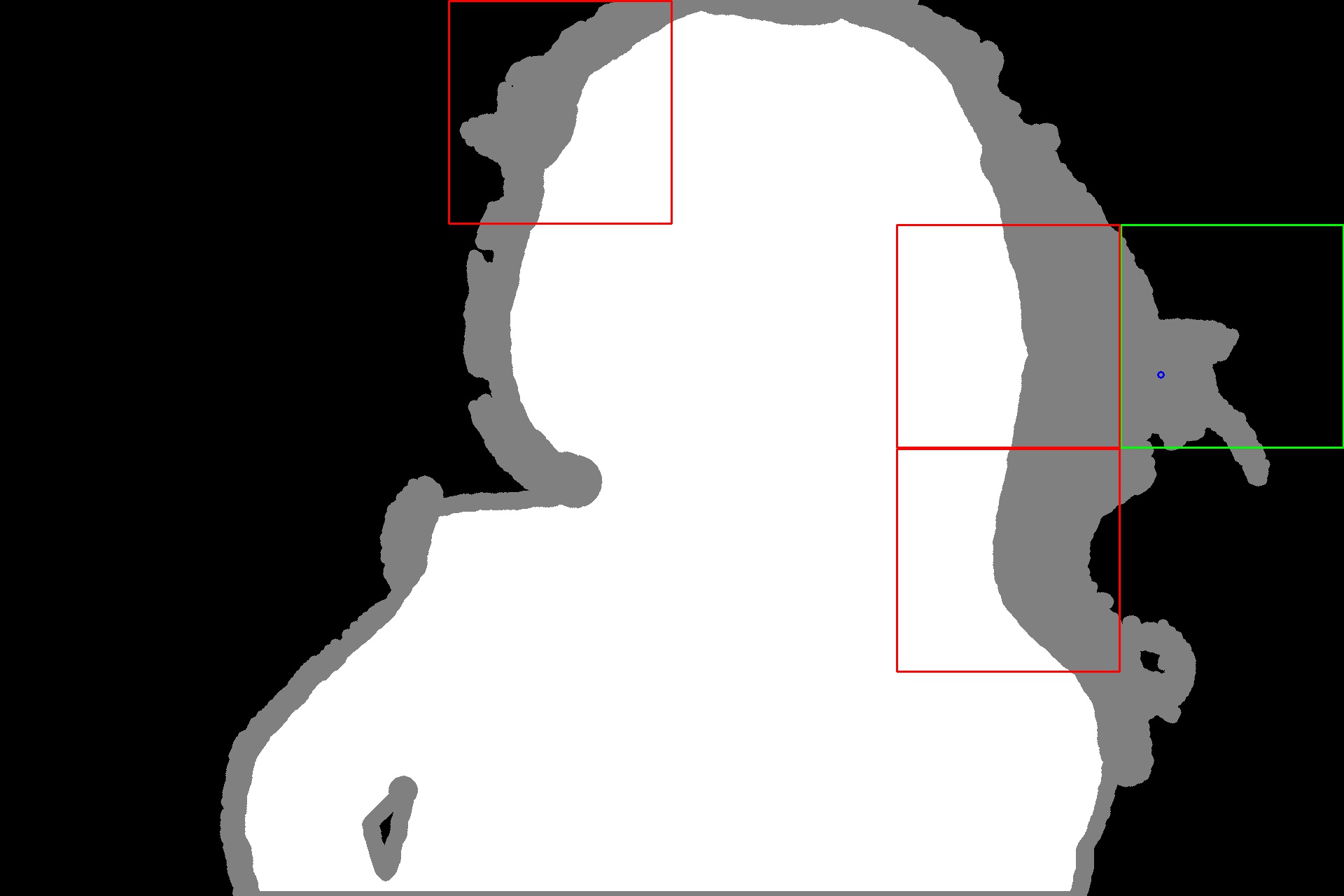}
    \caption{CPC attention visualization. Top row: whole image. Bottom row: whole trimap. Green box: query patch with the sampled pixel in blue circle. Red boxes: context patches. Zoom in for more details.}
    \label{fig:attention}
    \vspace{-1.2em}
\end{figure}

\subsection{Attention Visualization on Context Patches}
In Fig.~\ref{fig:attention}, we visualize the attention maps of the selected context patches for some given query patch. For each query patch in green box, we first select the top-3 context patches indicated by the red boxes. We then randomly sample a pixel in query patch marked by blue circle and show its attention/correlation maps on the context patches. Brighter color represents larger attention weights. It is worth noting that our method could select context patches which are far away from the query patch, which cannot be achieved with conventional CNNs with a fixed receptive field. Also, the attention weights indicate that our method can effectively leverage the information of similar pixels in the context patches.

\begin{table}[th]
    \begin{center}
        \caption{The ablation study on module, patch size and sampled patch number. Model A: without CPC. Model B: with CPC (Non-local). Model C: with CPC (TGNL). Each model name is followed by (sampled patch number $K$, patch size). MS: \{320, 480, 640\}.}
    \begin{tabular}{l|C{0.6cm}C{1.8cm}C{0.6cm}C{0.6cm}}
        \hline
        Models & SAD & MSE ($10^{-3}$) & Grad & Conn \\
        \hline
        Model A (3, 320) & 41.1 & 10.1 & 19.1 & 38.7 \\
        Model B (3, 320) & 35.6 & 8.0 & 17.9 & 33.1 \\
        Model C (3, 320) & 33.5 & 7.3 & 14.5 & 29.9 \\
        \hline
        Model C (3, 480) & 33.0 & 7.0 & 14.2 & 29.3 \\
        Model C (3, 640) & 33.1 & 7.0 & 14.4 & 29.6 \\
        Model C (3, MS) & 32.8 & 6.9 & 14.2 & 29.1 \\
        \hline
        Model C (1, 320) & 34.9 & 7.4 & 14.8 & 30.6 \\
        Model C (5, 320) & 33.3 & 7.2 & 14.3 & 29.7  \\
        Model C (7, 320)& 33.2 & 7.2 & 14.3 & 29.6 \\
        Model C (All, 320) & 32.2 & 7.2 & 14.2 & 29.5 \\
        \hline
    \end{tabular} \label{tab:abla}
    \end{center}
    \vspace{-1.9em}
\end{table}

\subsection{Ablation Study}
\subsubsection{Module Selection}
To investigate how each module contributes to the state-of-the-art performance of our method, we make an ablation study on the proposed modules in AIM dataset. The results are summarized in Tab.~\ref{tab:abla}. The baseline model is the encoder-decoder structure without CPC module (Model A), and it shares the same backbone with our proposed model. Compared with the baseline, CPC helps gain a large performance improvement. Substituting normal Non-local operation (Model B) with the proposed TGNL (Model C) inside the CPC further boosts the performance of the overall model.

\subsubsection{Patch Size}
In this section, we make an ablation study to show the dependency of CPC on patch size of query and context patches. We test our CPC (TGNL) model with patch sizes 320, 480, 640 and \{320, 480, 640\} as shown in the second section of Tab.~\ref{tab:abla}.
The maximum SAD difference among various patch size settings is only 0.7. This implies a nice property that CPC module is agnostic to patch size to some extent. This is because our model is already designed to capture long-range cross-patch context by CPC and larger patch size does not make the model capture much extra context. This property is useful in real applications since given limited computation resources, our method can run on smaller patches without sacrificing the performance.

\subsubsection{Context Patch Number}
During testing, to predict alpha matte for the query patch, we sample $K$ patches in the context pool. In this section, we explore how $K$ impacts test performance. We trained and tested the CPC (TGNL) model with $K=1,3,5,7$ and all the context patches.
As shown in Tab.~\ref{tab:abla}, even with a single context patch, our method already achieves significantly improvement over the baseline model (no CPC), showing the effectiveness of our CPC module.
When all patches in the original image are used as context patches, the model yields the best performance of SAD $32.2$. Since when $K\ge 3$ the model performs stably, we choose $K=3$ in our experiments considering the trade-off of computational cost and performance. 
\section{Conclusion}
\label{sec:con}
In this paper, we propose HDMatt, a first deep learning based model for HR image matting. Instead of taking the whole image as input for inference, we apply a patch-based training and inference strategy to overcome hardware limitations for HR inference. To maintain a high-quality alpha matte, we explicitly model the cross-patch long-range context dependency using a Cross-Patch Context module. Our method achieves new state-of-the-arts on AIM, AlphaMatting benchmarks and produce impressive visual results on real-world high-resolution images.

\bibliography{references.bib}

\clearpage
\appendix
\section*{Appendices}

\section{Patch Stitching Method}
\label{app:stitch}
As we mentioned in the paper, our patch-based method runs in a crop-and-switch manner. To obtain final alpha prediction, alpha matte patches are stitched together with overlap. Overlapping regions are merged using blending for a smooth transition.

Given a predicted alpha matte patch $\boldsymbol{\alpha^p}$, we assign a weight $w_i^p$ for each pixel $\alpha_i^p$ in $\boldsymbol{\alpha^p}$. In $\boldsymbol{\alpha^p}$, there is an overlapping region (OR) with the neighboring patches on its boundaries and non-overlapping region (NOR) on its central part. In NOR, $w_i^p$ is always set to $1$. In OR, $w_i^p$ is defined as
\begin{equation}
    w_i^p = \frac{\mathrm{boundary\_dist}(\alpha_i^p)}{margin},
\end{equation}
where $\mathrm{boundary\_dist}(\cdot)$ is the shortest distance between $\alpha_i^p$ and patch boundaries and $margin$ is the width of the overlapping region. Thus, $w_i^p \in [0, 1]$ in OR. During training, $w_i^p$ is used as weights to compute the overall training loss as mentioned in our paper.
Finally, for the a pixel $\alpha$ in the whole alpha matte, it is the weighted sum as below
\begin{equation}
  \alpha = \frac{\sum_{p=1}^{P} \alpha_{i_p}^p w_{i_p}^p}{\sum_{i=1}^{P} w_{i_p}^p}, \quad P\in\{1, 2, 4\},
\end{equation}
where $P$ is the number of overlapping patches for the pixel $\alpha$ in the whole alpha matte and $i_p$ is the pixel index in $p$-th patch that corresponds to the pixel $\alpha$.

\section{The AlphaMatting benchmark}
\label{app:alpha}
\begin{figure*}[h]
\centering
  \includegraphics[width=.158\linewidth]{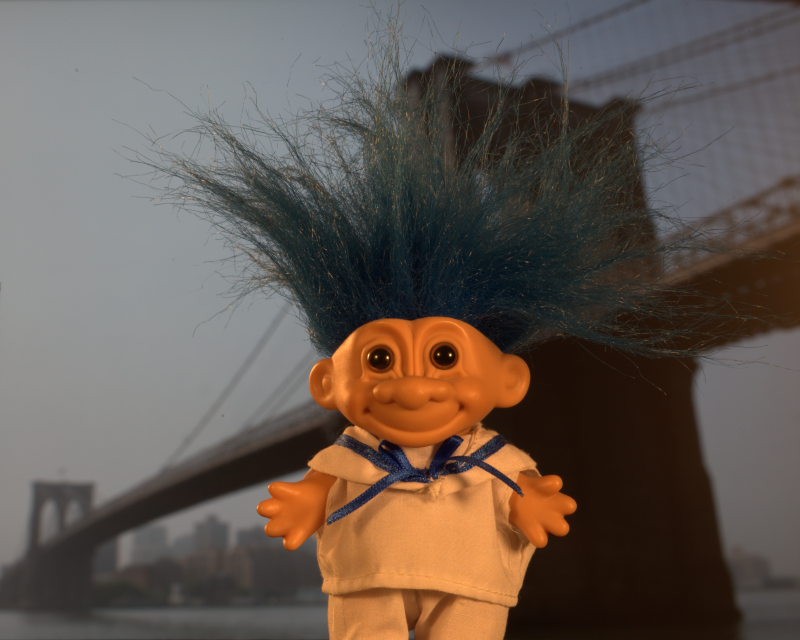}
  \includegraphics[width=.158\linewidth]{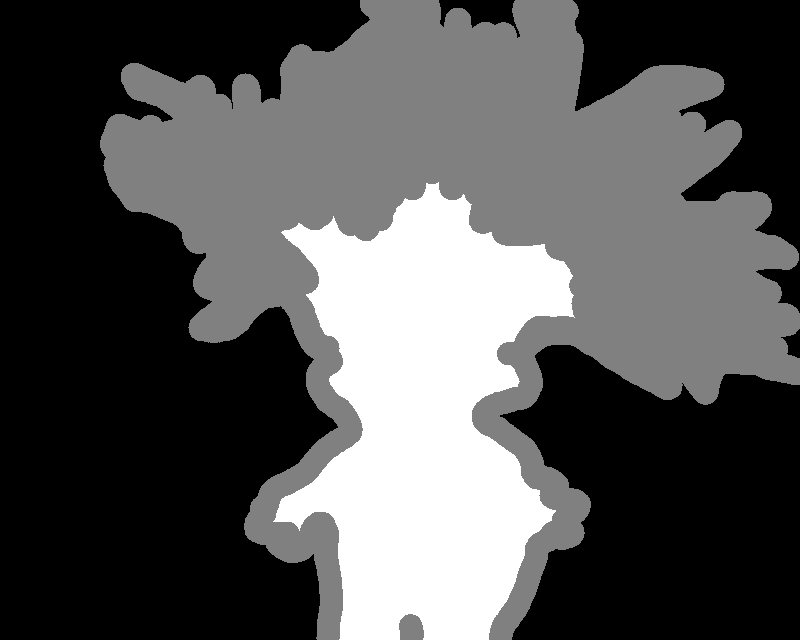}
  \includegraphics[width=.158\linewidth]{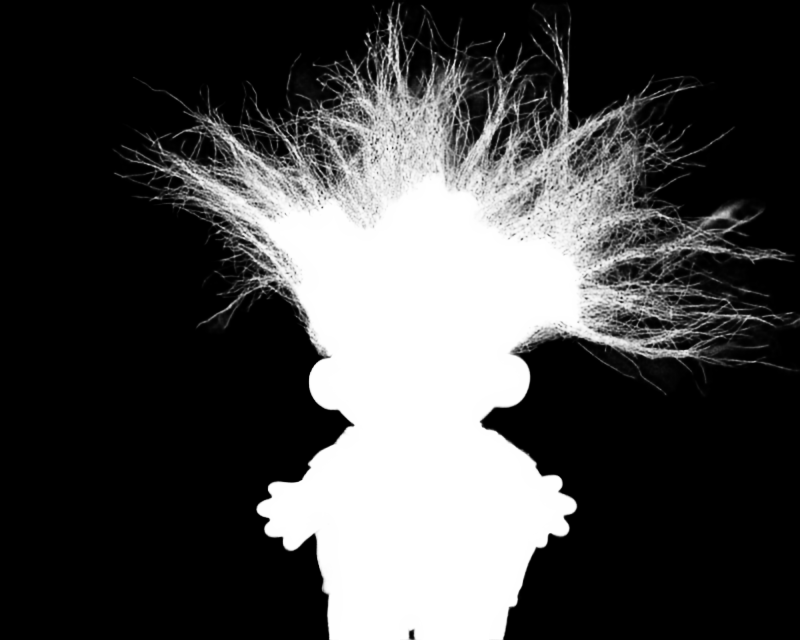}
  \includegraphics[width=.158\linewidth]{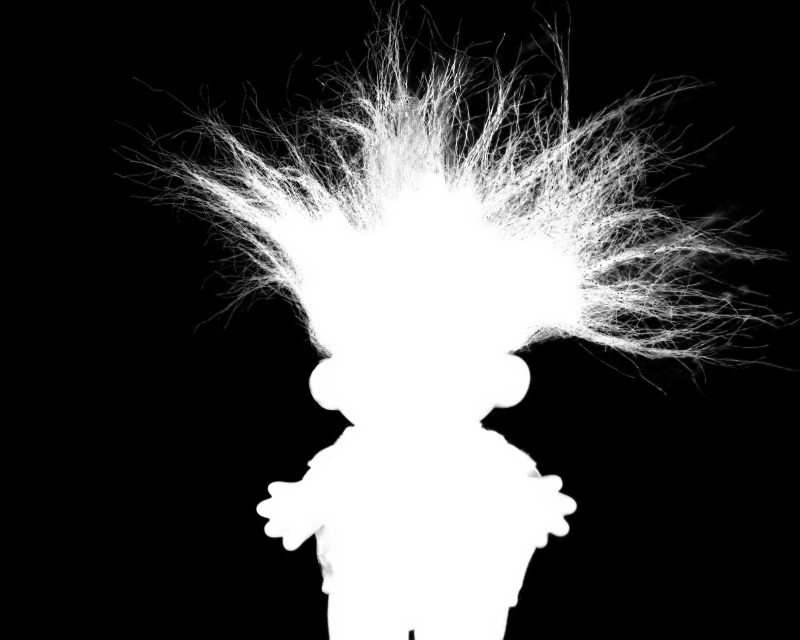}
  \includegraphics[width=.158\linewidth]{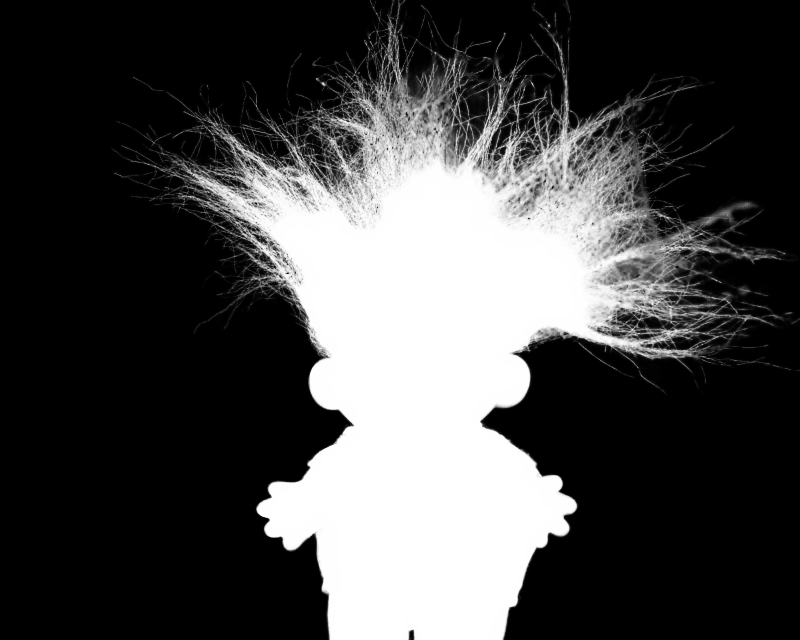}
  \includegraphics[width=.158\linewidth]{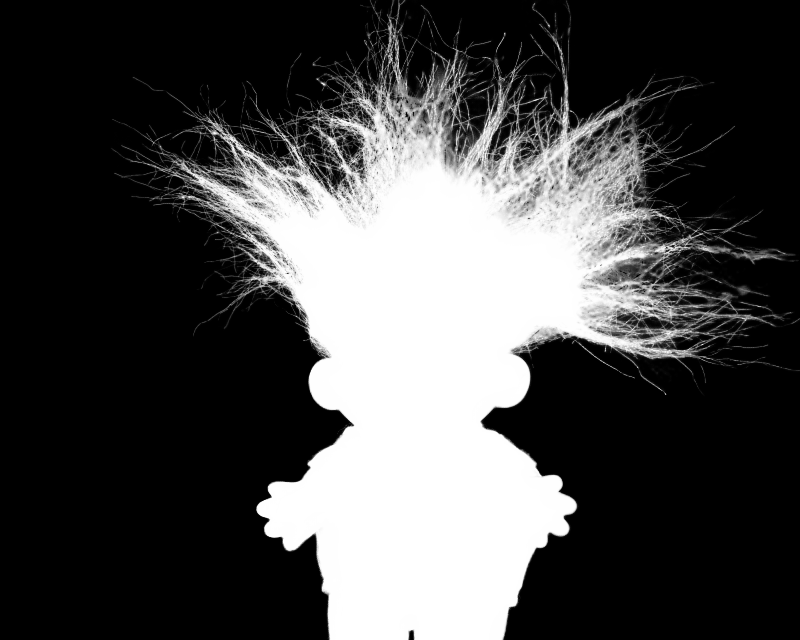}

  \includegraphics[width=.158\linewidth]{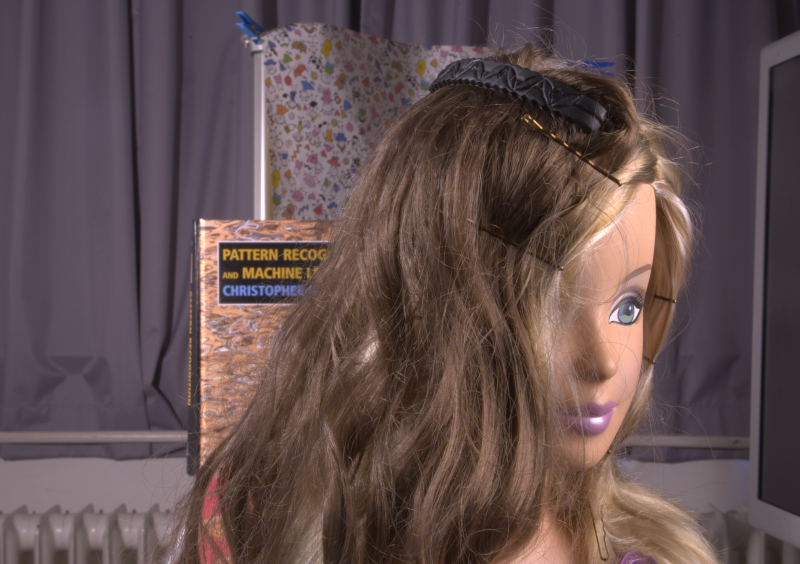}
  \includegraphics[width=.158\linewidth]{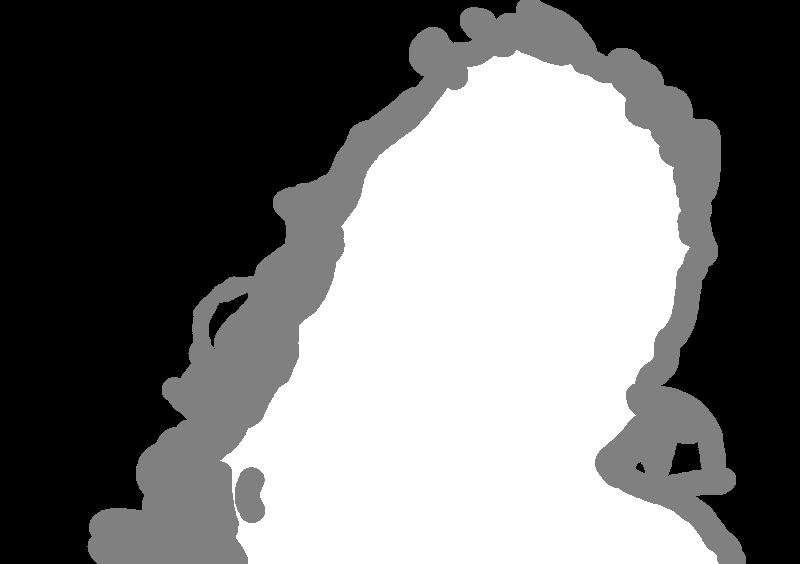}
  \includegraphics[width=.158\linewidth]{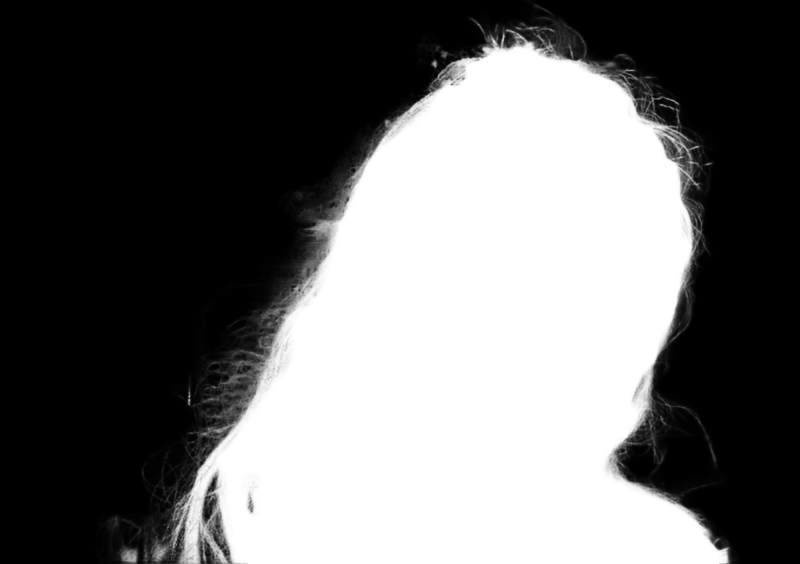}
  \includegraphics[width=.158\linewidth]{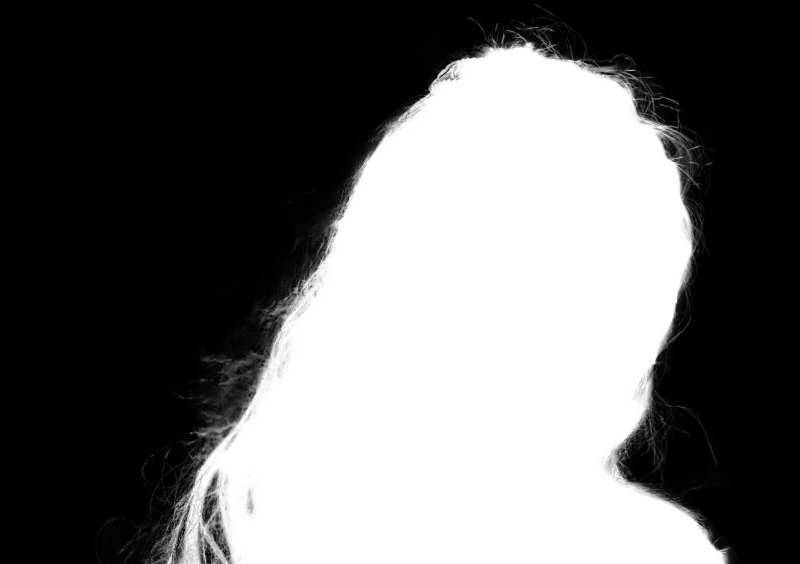}
  \includegraphics[width=.158\linewidth]{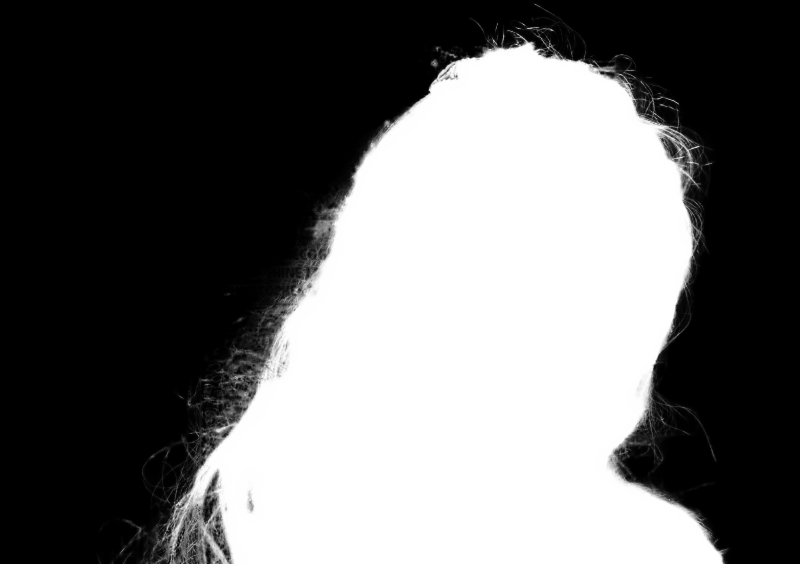}
  \includegraphics[width=.158\linewidth]{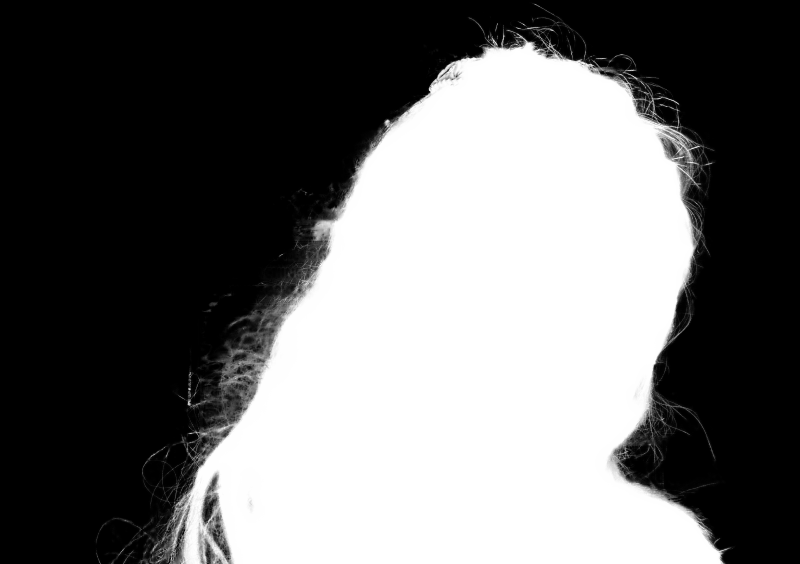}
  
  \includegraphics[width=.158\linewidth]{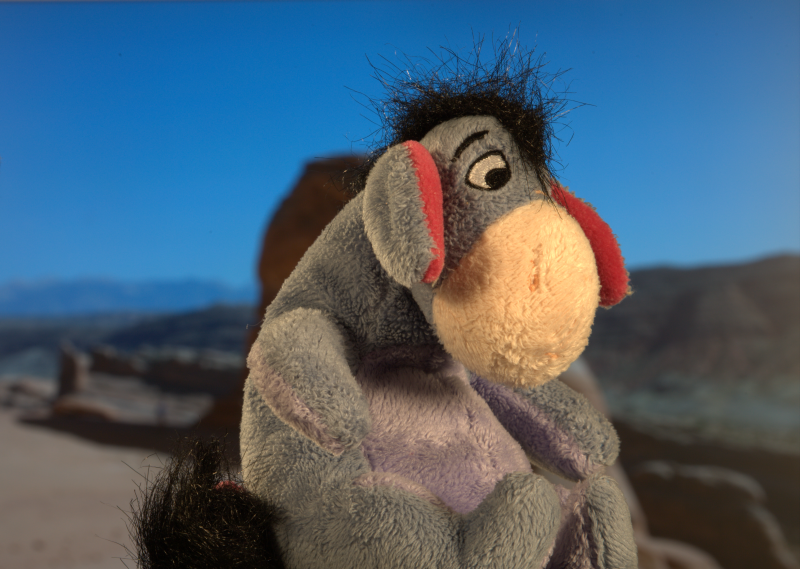}
  \includegraphics[width=.158\linewidth]{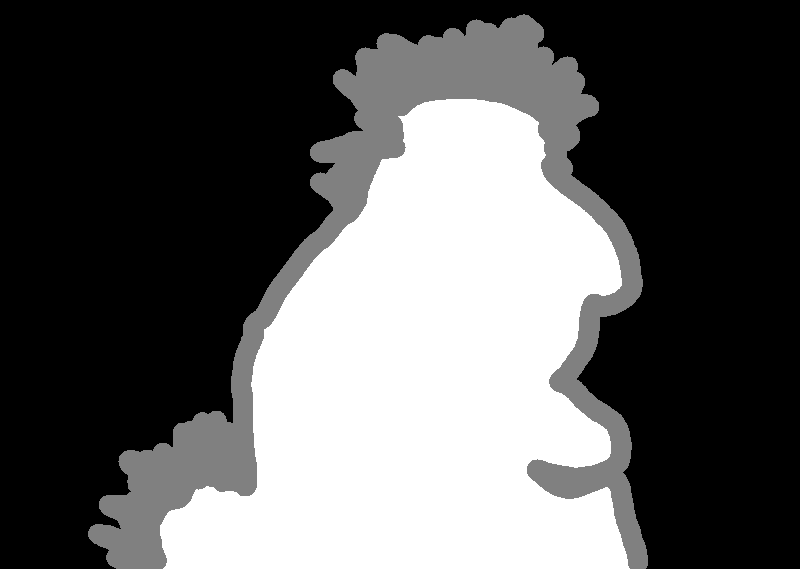}
  \includegraphics[width=.158\linewidth]{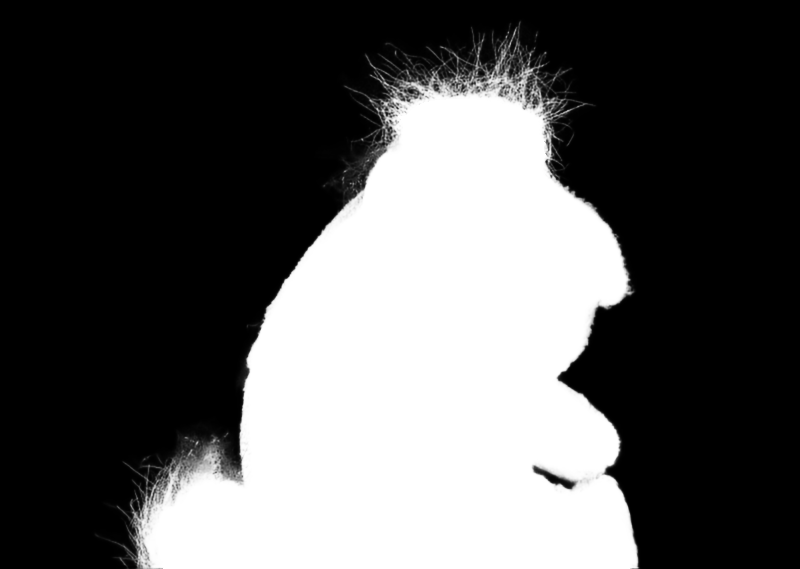}
  \includegraphics[width=.158\linewidth]{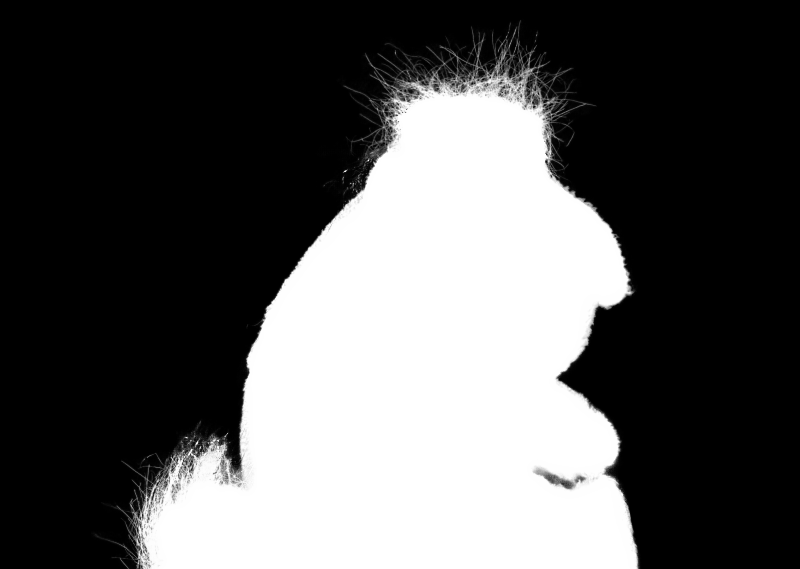}
  \includegraphics[width=.158\linewidth]{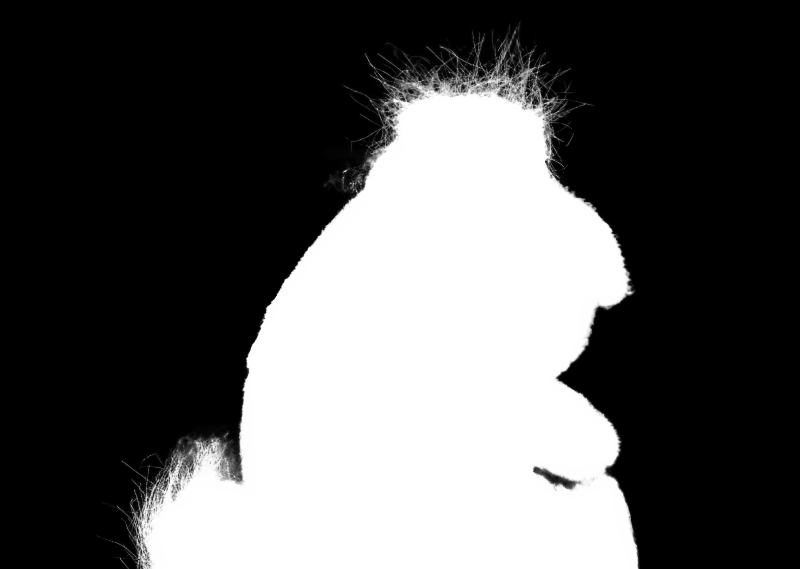}
  \includegraphics[width=.158\linewidth]{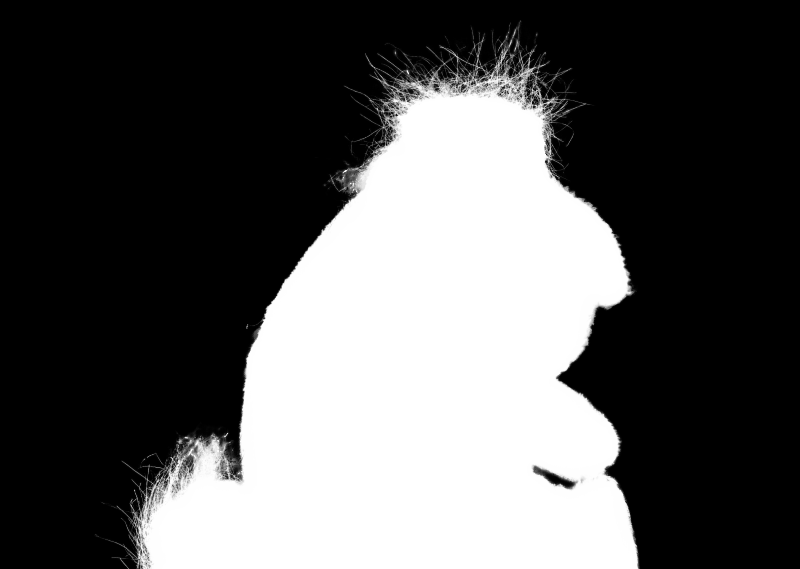}
  
  \includegraphics[width=.158\linewidth]{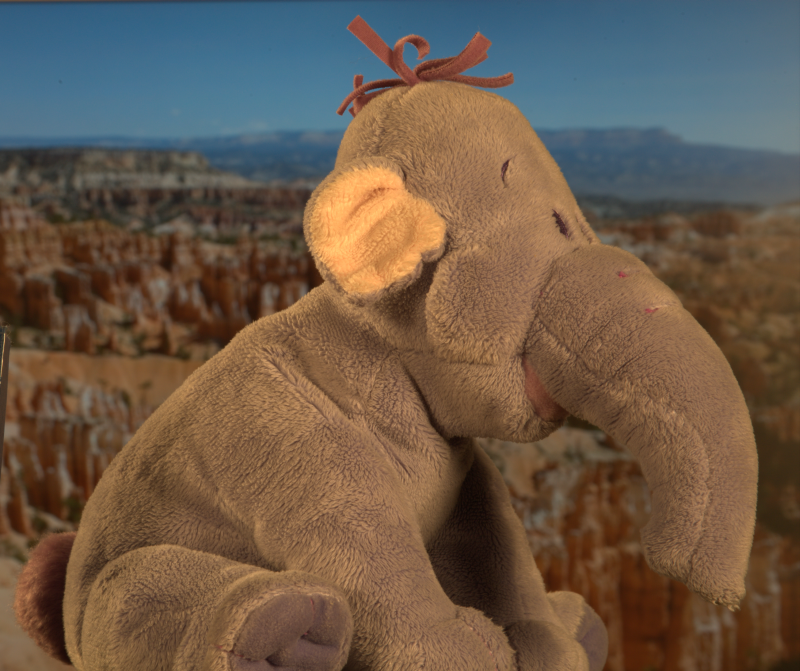}
  \includegraphics[width=.158\linewidth]{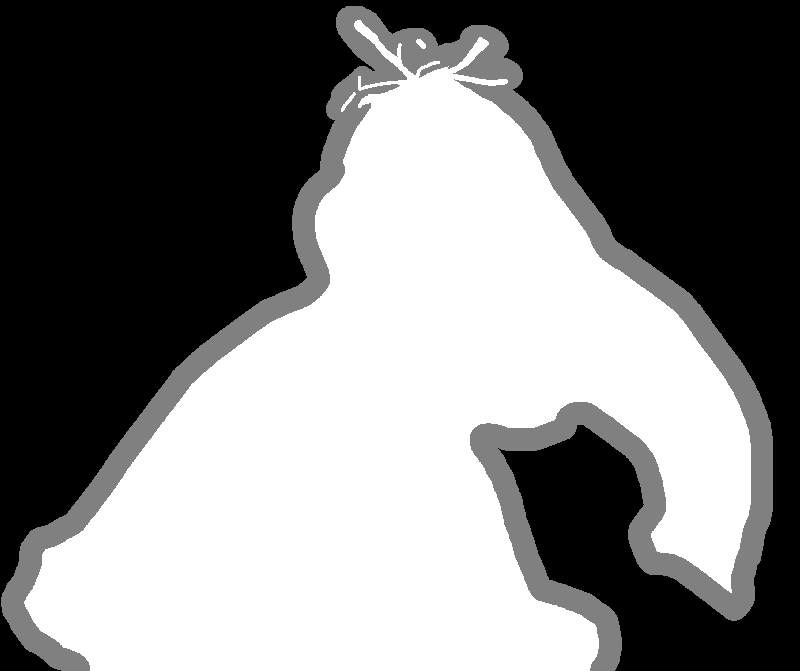}
  \includegraphics[width=.158\linewidth]{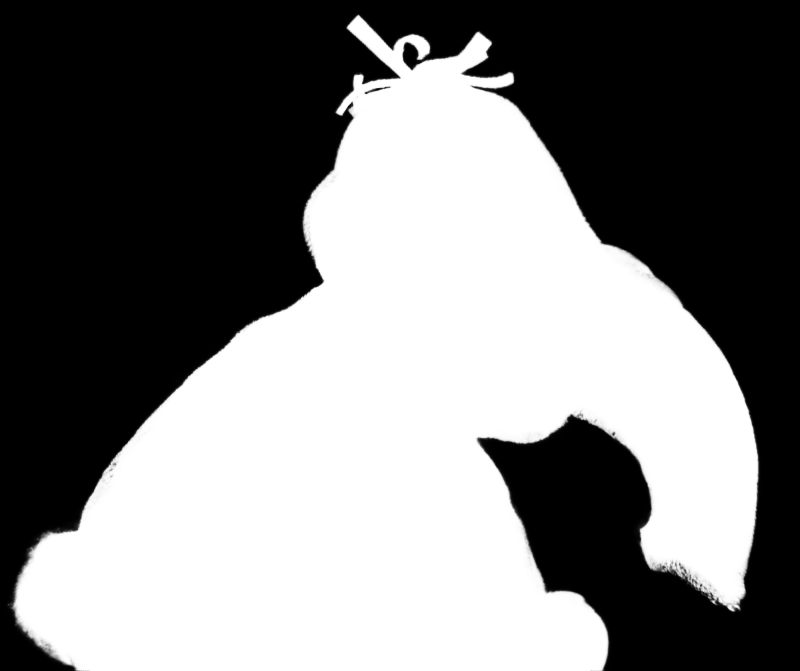}
  \includegraphics[width=.158\linewidth]{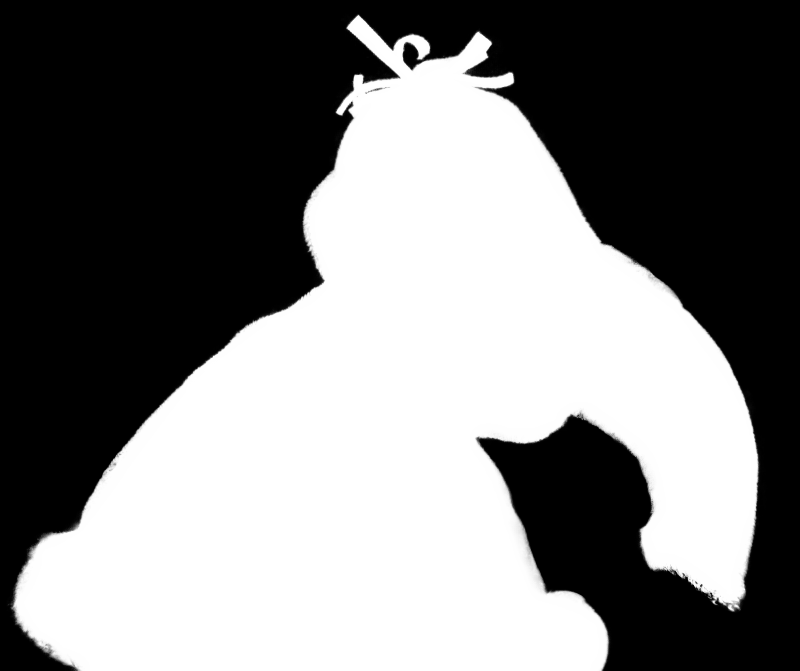}
  \includegraphics[width=.158\linewidth]{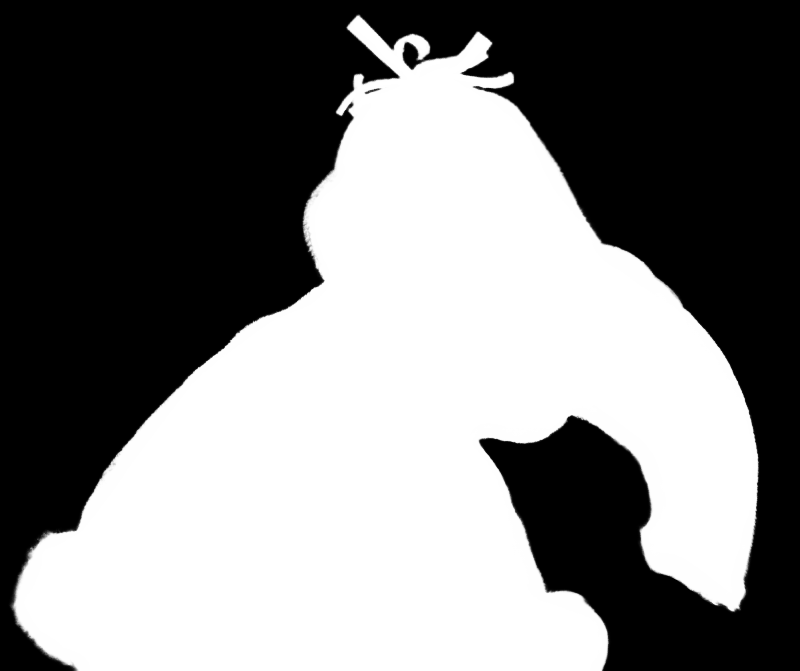}
  \includegraphics[width=.158\linewidth]{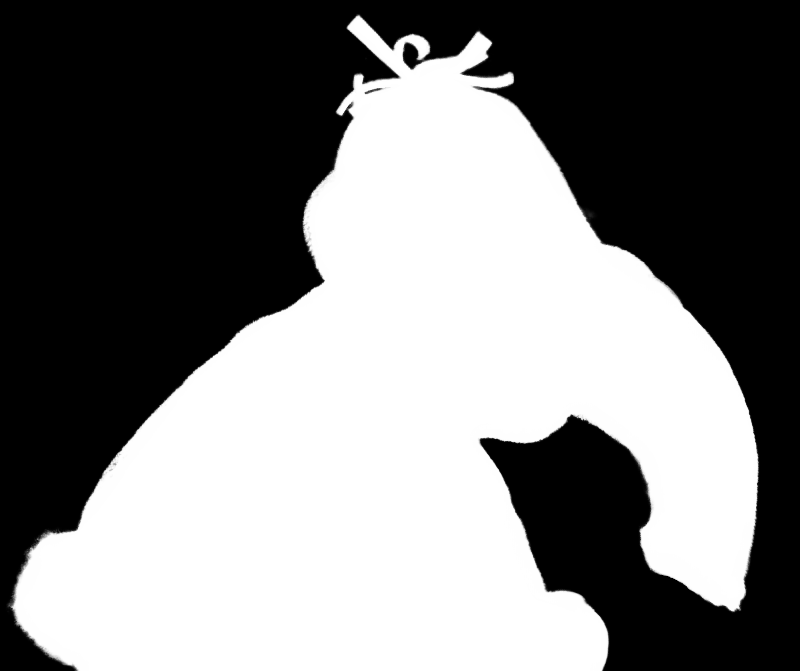}
  
  \includegraphics[width=.158\linewidth]{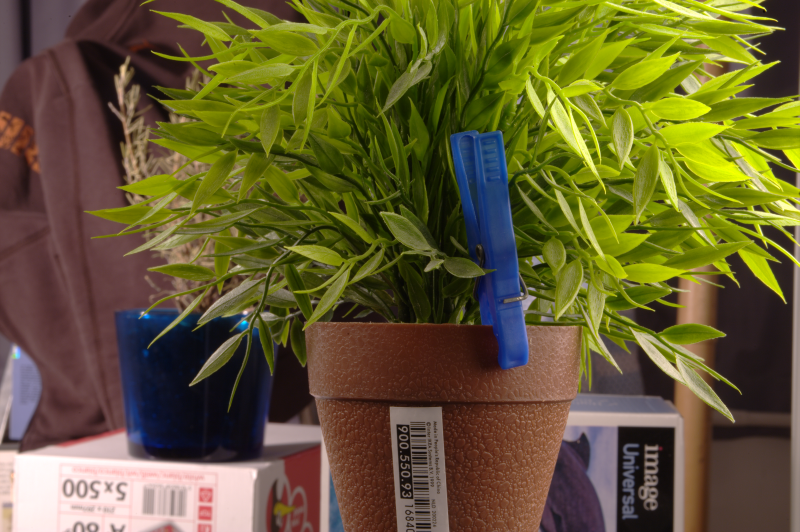}
  \includegraphics[width=.158\linewidth]{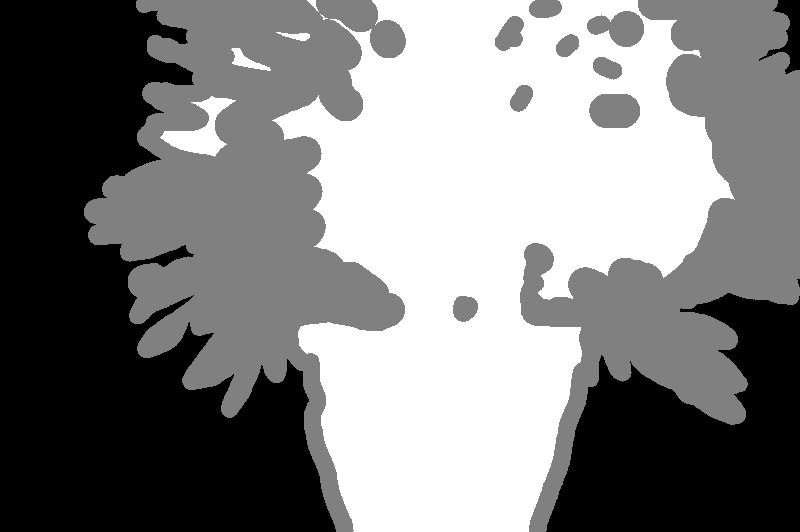}
  \includegraphics[width=.158\linewidth]{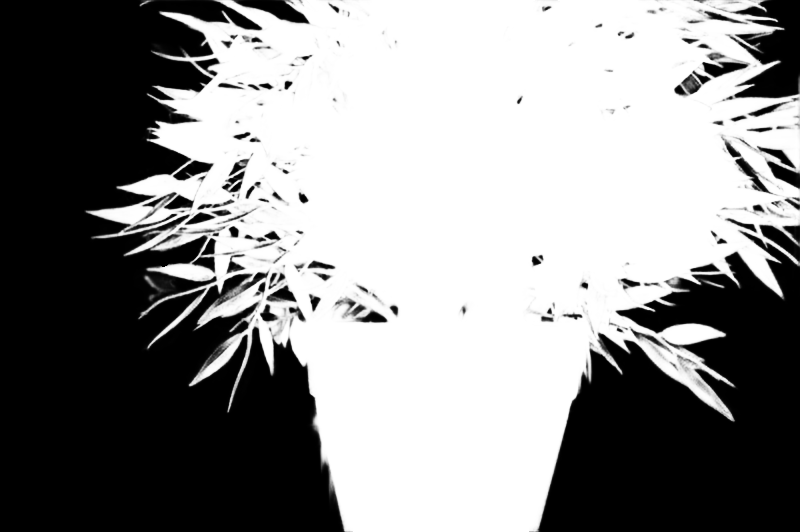}
  \includegraphics[width=.158\linewidth]{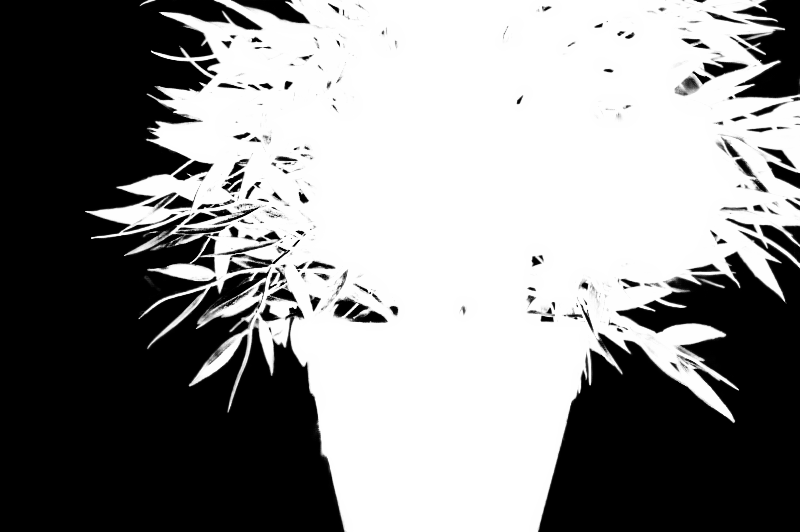}
  \includegraphics[width=.158\linewidth]{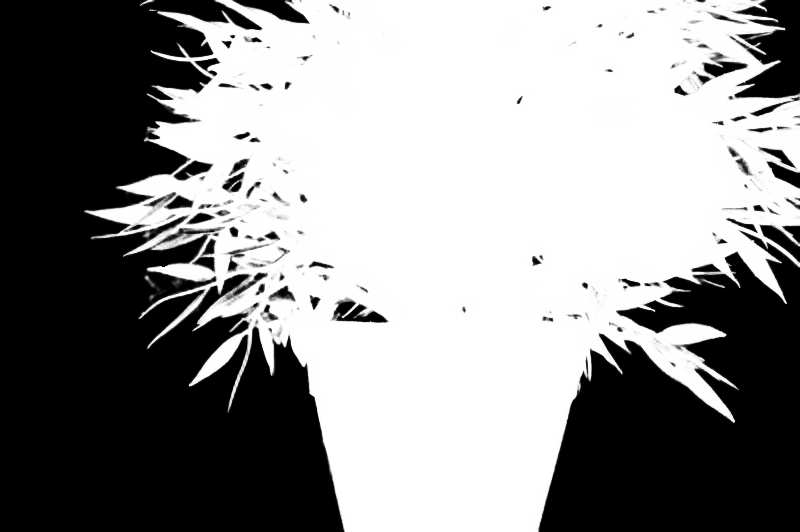}
  \includegraphics[width=.158\linewidth]{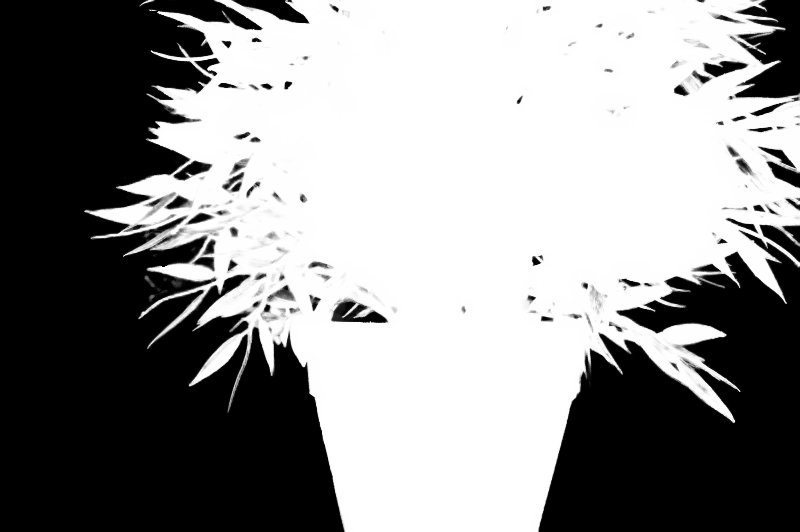}

  \includegraphics[width=.158\linewidth]{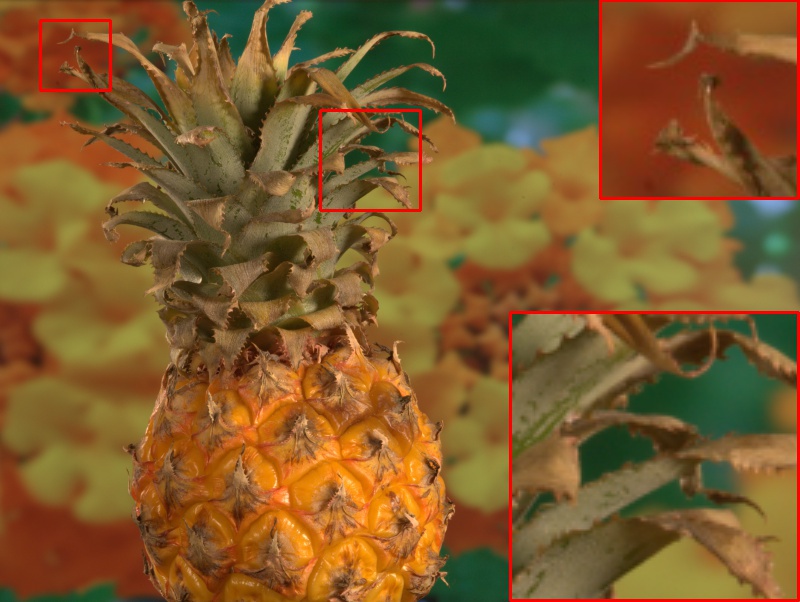}
  \includegraphics[width=.158\linewidth]{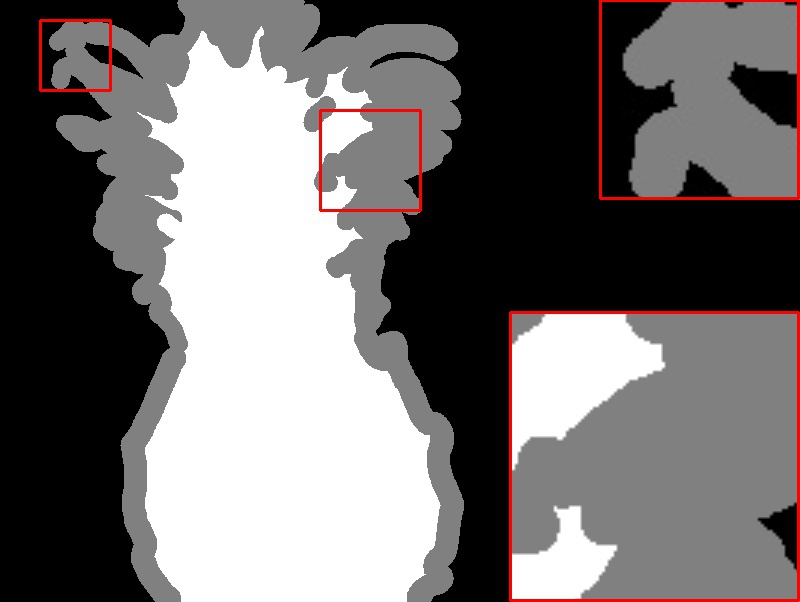}
  \includegraphics[width=.158\linewidth]{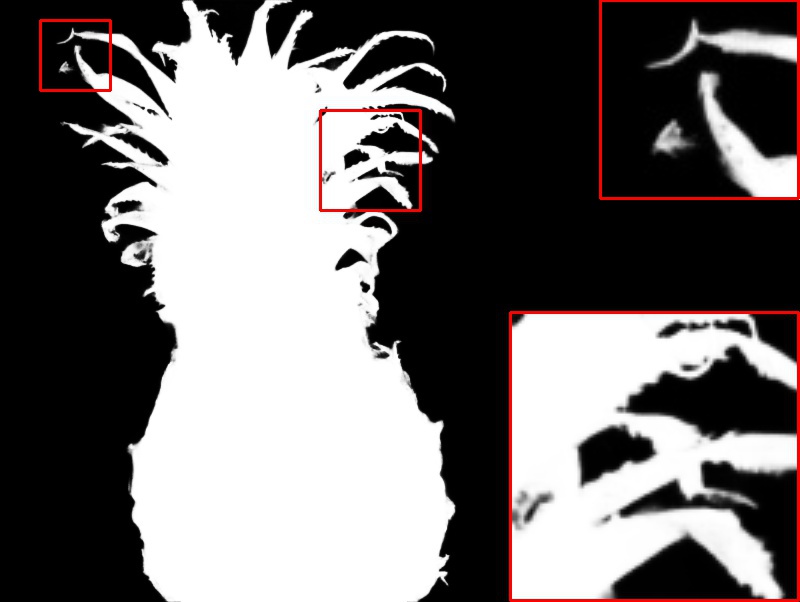}
  \includegraphics[width=.158\linewidth]{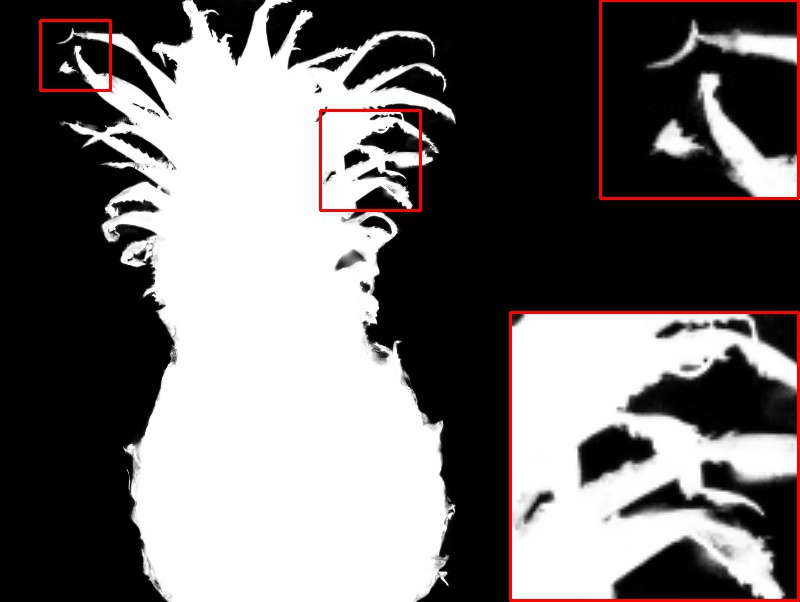}
  \includegraphics[width=.158\linewidth]{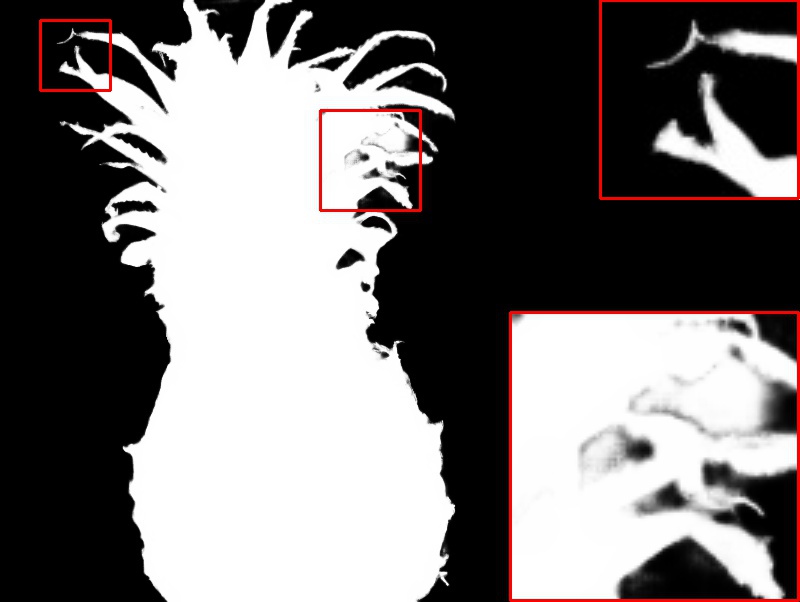}
  \includegraphics[width=.158\linewidth]{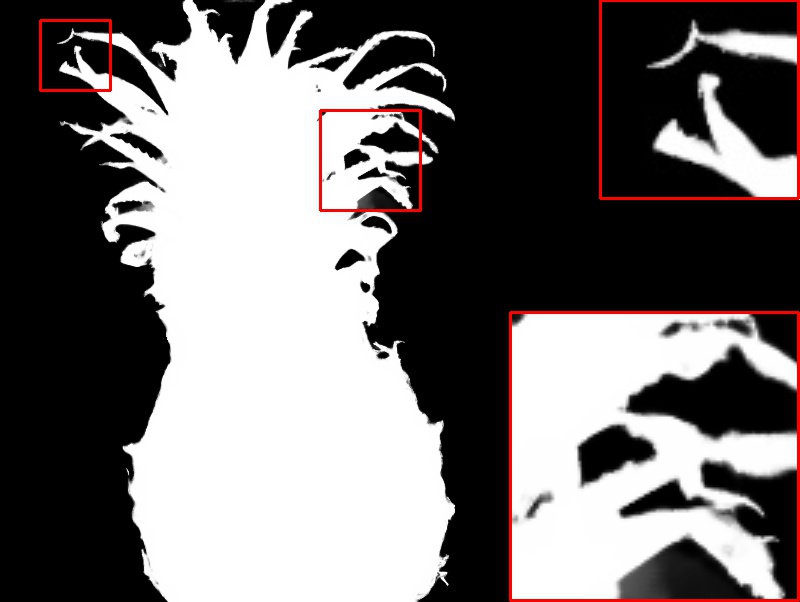}

  \includegraphics[width=.158\linewidth]{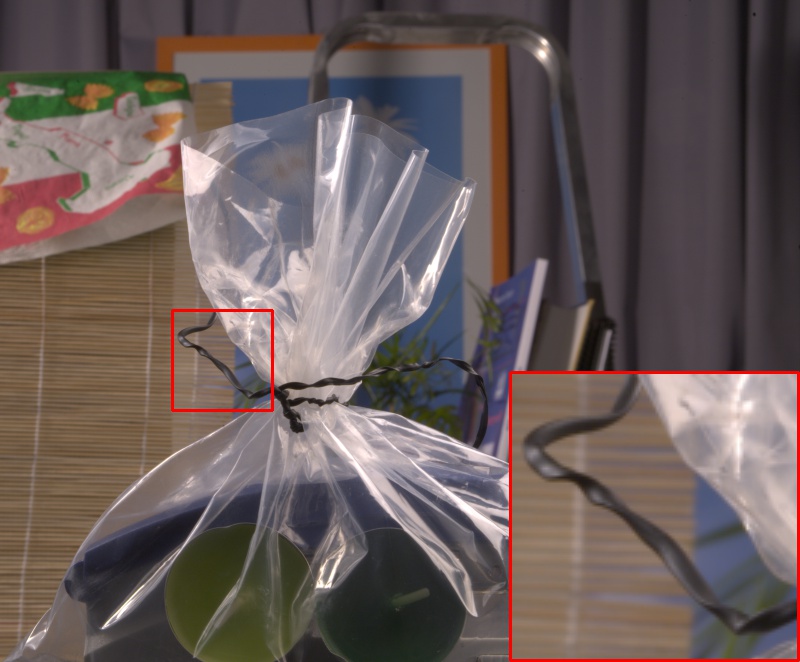}
  \includegraphics[width=.158\linewidth]{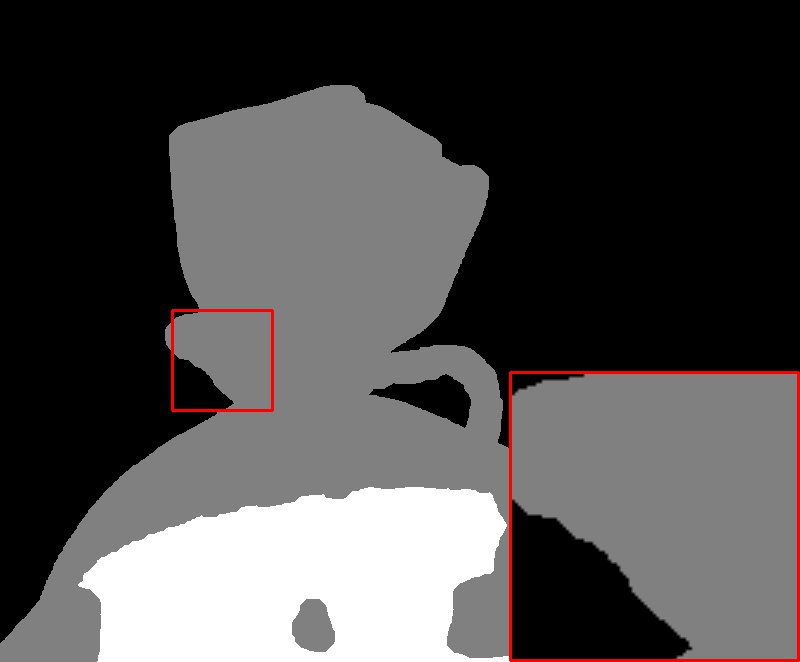}
  \includegraphics[width=.158\linewidth]{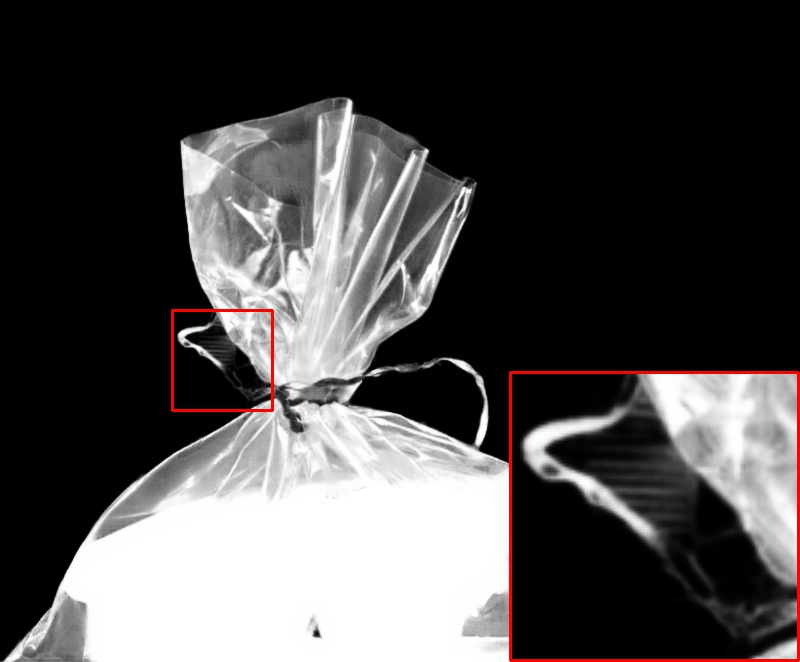}
  \includegraphics[width=.158\linewidth]{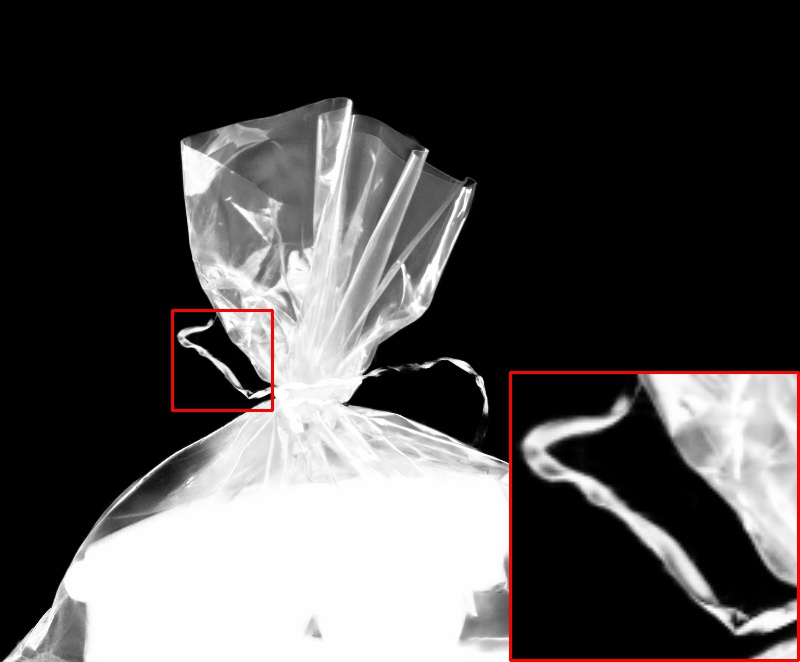}
  \includegraphics[width=.158\linewidth]{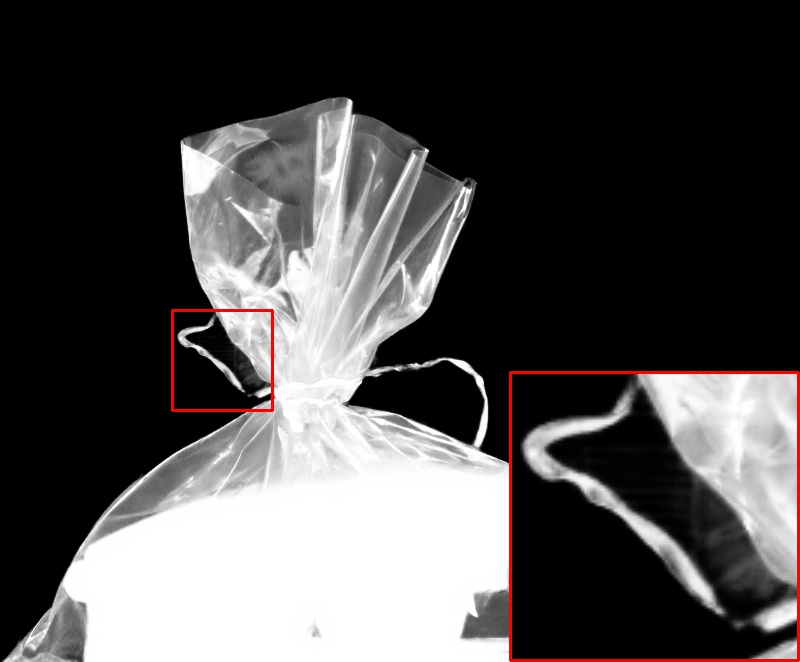}
  \includegraphics[width=.158\linewidth]{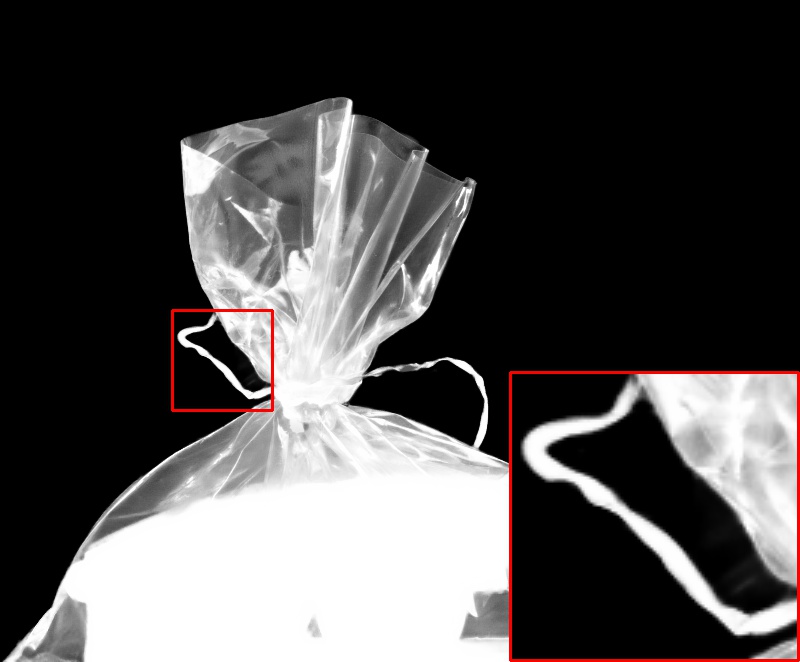}

  \includegraphics[width=.158\linewidth]{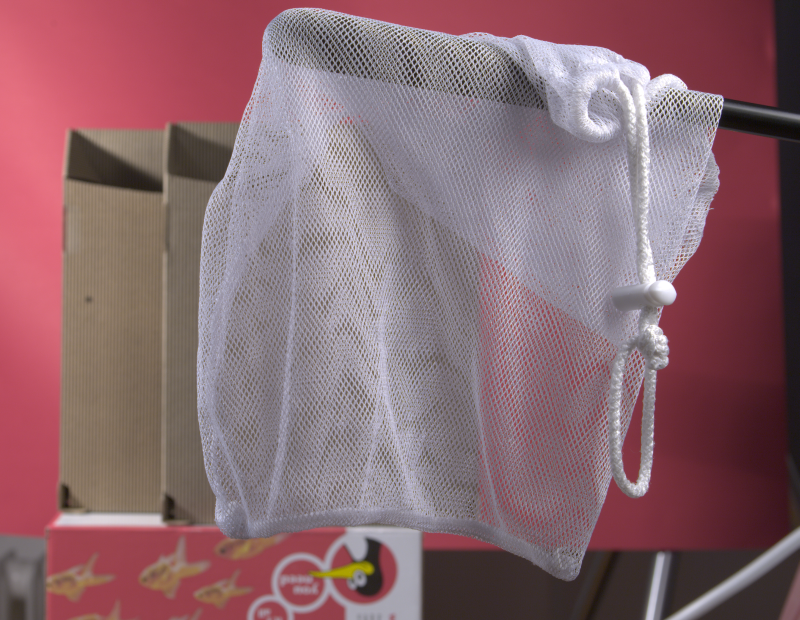}
  \includegraphics[width=.158\linewidth]{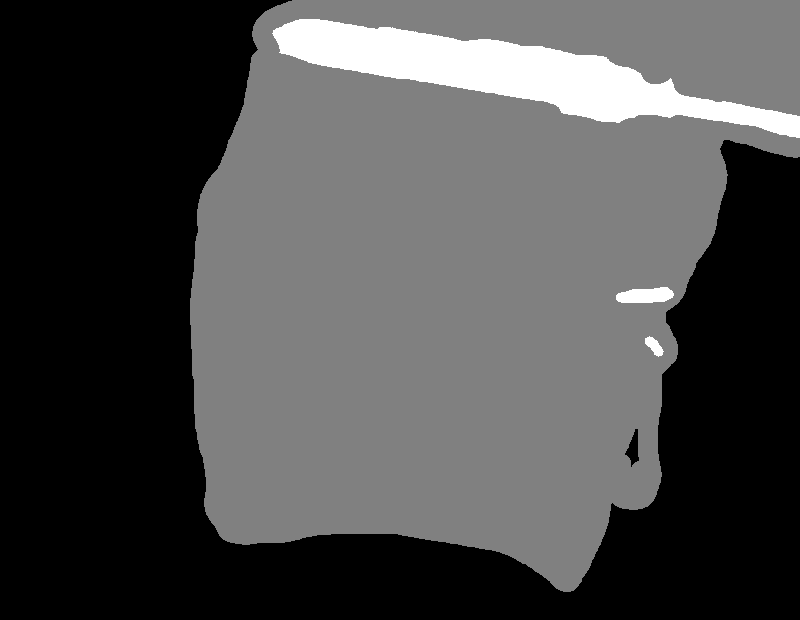}
  \includegraphics[width=.158\linewidth]{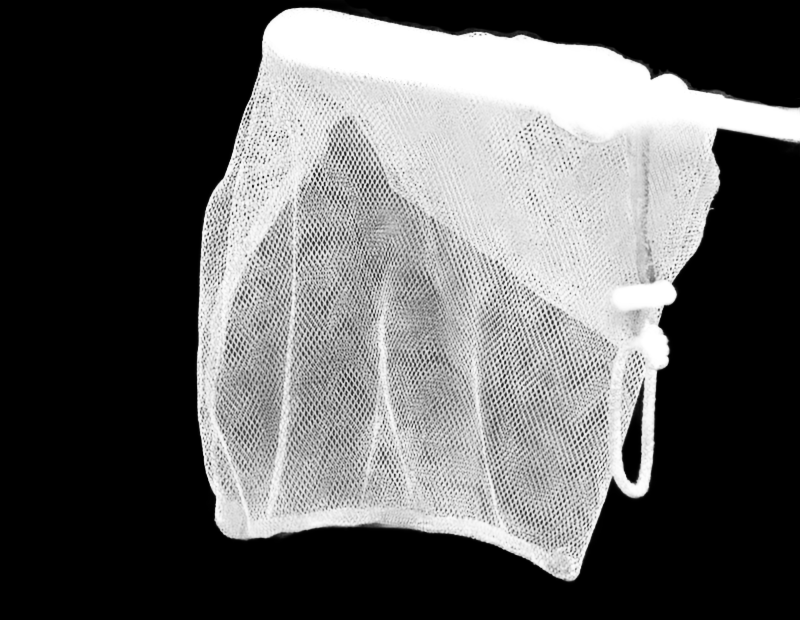}
  \includegraphics[width=.158\linewidth]{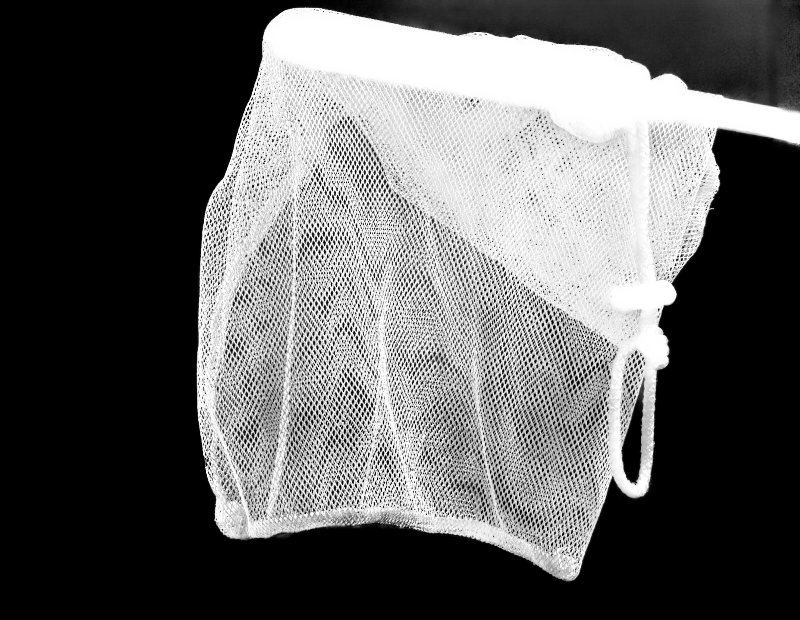}
  \includegraphics[width=.158\linewidth]{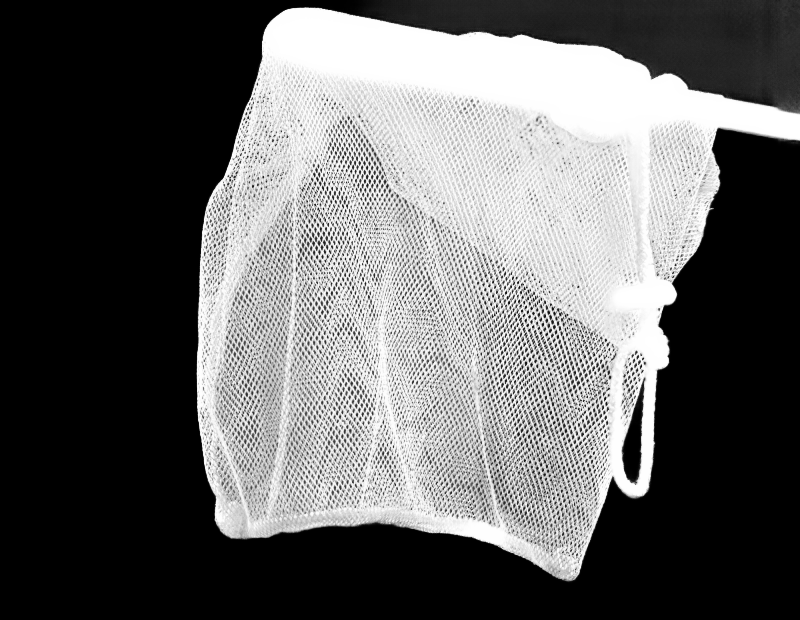}
  \includegraphics[width=.158\linewidth]{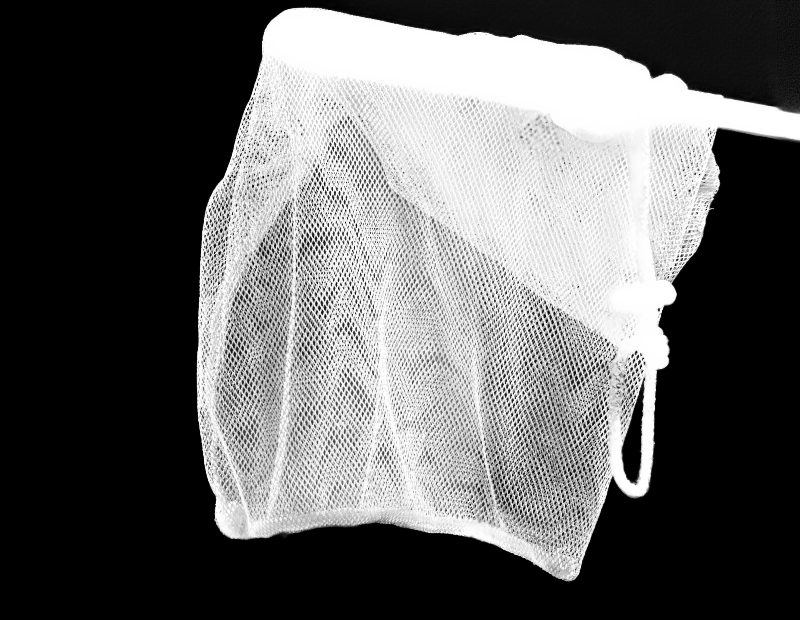}
 
\caption{Visual results on alphamatting.com testset. From left to right: Image, Trimap (U), AdaMatting~\cite{cai2019disentangled}, SampleNet~\cite{tang2019learning}, GCA Matting~\cite{li2020natural} and HDMatt (Ours).}
\label{fig:compare_alphamat}
\end{figure*}

In this section, we show visual results of the top-ranking methods on AlphaMatting benchmark in Fig.~\ref{fig:compare_alphamat}.

\end{document}